\documentclass[lettersize,journal]{IEEEtran}
\usepackage{amsmath,amsfonts}
\usepackage{algorithmic}
\usepackage{algorithm}
\usepackage{array}
\usepackage{textcomp}
\usepackage{stfloats}
\usepackage{url}
\usepackage{verbatim}
\usepackage{graphicx}
\usepackage{cite}
\usepackage{tabularx}
\usepackage{url}
\usepackage{float}
\usepackage{subcaption}
\usepackage{multirow}
\usepackage{booktabs}
\usepackage{siunitx}
\usepackage[frozencache=true,cachedir=minted-cache]{minted}
\usepackage[inline]{enumitem}
\usepackage{tabularx}
\usepackage{hyperref}

\DeclareMathOperator*{\argmin}{arg\,min}
\newcommand{\mat}[1]{\boldsymbol{#1}}
\renewcommand{\vec}[1]{\boldsymbol{#1}}

\usepackage{fancyhdr}
\fancypagestyle{mahmood}{%
  \fancyhf{} 
  
  \fancyhead[C]{\footnotesize \textcopyright 2024 IEEE.  Personal use of this material is permitted.  Permission from IEEE must be obtained for all other uses, in any current or future media, including reprinting/republishing this material for advertising or promotional purposes, creating new collective works, for resale or redistribution to servers or lists, or reuse of any copyrighted component of this work in other works.}
}%
\makeatletter
\let\ps@IEEEtitlepagestyle\ps@mahmood
\makeatother
\begin{document}
\title{NanoSLAM: Enabling Fully Onboard SLAM \\ for Tiny Robots}

\author{Vlad~Niculescu,
        Tommaso~Polonelli,    ~\IEEEmembership{Member,~IEEE,} Michele~Magno,~\IEEEmembership{Senior Member,~IEEE,}
        and~Luca~Benini,~\IEEEmembership{Fellow,~IEEE}%
        
\thanks{V. Niculescu and L. Benini are with the Integrated Systems Laboratory of ETH Z\"urich, ETZ, Gloriastrasse 35, 8092 Z\"urich, Switzerland (e-mail: vladn@iis.ee.ethz.ch, lbenini@iis.ee.ethz.ch).}%
\thanks{T. Polonelli and M. Magno are with the Center for Project-Based Learning of ETH Z\"urich, ETZ, Gloriastrasse 35, 8092 Z\"urich, Switzerland (e-mail: tommaso.polonelli@pbl.ee.ethz.ch, michele.magno@pbl.ee.ethz.ch).}%
\thanks{L. Benini is also with the Department of Electrical, Electronic and Information Engineering of University of Bologna, Viale del Risorgimento 2, 40136 Bologna, Italy.}
\thanks{Copyright (c) 20xx IEEE. Personal use of this material is permitted. However, permission to use this material for any other purposes must be obtained from the IEEE by sending a request to pubs-permissions@ieee.org.}
\thanks{Digital Object Identifier 10.1109/JIOT.2023.3339254}}

\maketitle

\begin{abstract}
Perceiving and mapping the surroundings are essential for autonomous navigation in any robotic platform. The algorithm class that enables accurate mapping while correcting the odometry errors present in most robotics systems is Simultaneous Localization and Mapping (SLAM). 
Today, fully onboard mapping is only achievable on robotic platforms that can host high-wattage processors, mainly due to the significant computational load and memory demands required for executing SLAM algorithms. 
For this reason, pocket-size hardware-constrained robots offload the execution of SLAM to external infrastructures.  
To address the challenge of enabling SLAM algorithms on resource-constrained processors, this paper proposes NanoSLAM, a lightweight and optimized end-to-end SLAM approach specifically designed to operate on centimeter-size robots at a power budget of only~\qty{87.9}{\milli\watt}.
We demonstrate the mapping capabilities in real-world scenarios and deploy NanoSLAM on a nano-drone weighing \qty{44}{\gram} and equipped with a novel commercial RISC-V low-power parallel processor called GAP9. 
The algorithm, designed to leverage the parallel capabilities of the RISC-V processing cores, enables mapping of a general environment with an accuracy of \qty{4.5}{\centi\meter} and an end-to-end execution time of less than \qty{250}{\milli\second}.
\end{abstract}


\section*{Supplementary Material}
Supplementary video at \url{https://youtu.be/XUSVLHJ87J0} \\ Project's code at \url{https://github.com/ETH-PBL/NanoSLAM}

\section{Introduction}
The field of autonomous pocket-size robotics systems and Unmanned Aerial Vehicles (UAVs) experienced rapid growth in the past years due to the advancement and miniaturization of capable embedded computing platforms creating new possibilities for IoT applications~\cite{wang2020deep, huang2022edge,yang2021survey, vermesan2020internet}.
Nano-robots, and especially palm-size UAVs, weigh only a few tens of grams and benefit from increased agility compared to their standard-size counterparts, enabling them to fly in narrow spaces reliably~\cite{duisterhof2021tiny, karam2022micro}.
Furthermore, their reduced dimensions make nano-UAVs perfect candidates for safely operating near humans, especially in cramped indoor environments~\cite{niculescu2021improving,liu2021cooperative}.
In most practical applications, the mission of the nano-UAV is to follow a path through the environment that is predefined or adjusted dynamically during the mission~\cite{niculescu2021improving,duisterhof2021tiny}.
For instance, finding the source of gas leaks~\cite{duisterhof2021sniffy} or localizing and reaching sensor nodes for data acquisition~\cite{niculescu2022energy} are only a few examples of such applications.

The environments where nano-UAVs typically fly are filled with walls and obstacles, and thus, optimal path planning requires good knowledge of the surroundings map~\cite{duisterhof2021tiny}.
Furthermore, in a wide range of applications, the map can change over time, so preprogramming the map into the nano-UAVs is not an ideal solution~\cite{muller2023fully}.
In smart buildings, for instance, where the layout of reconfigurable walls can change~\cite{kisseleff2021reconfigurable}, or simply in crowded offices where tables, chairs, and furniture are often moved. Moreover, the arrangement of pallets and shelves in warehouses can change from one day to another, and therefore, the UAVs used for inventory need a constantly updated map for reliable navigation~\cite{cristiani2020inventory, chen2020warehouse, moura2021graph}.

In the scenarios mentioned so far, the drone needs an accurate environmental knowledge and the ability to localize itself within the map~\cite{muller2023fully}.
The algorithm class that performs both tasks is called Simultaneous Localization and Mapping (SLAM).
Among the existing SLAM algorithms, graph-based SLAM~\cite{grisetti2010tutorial, chen2021cramer} is one of the most adopted variations of the algorithm due to its high accuracy and capability to refine the complete trajectory. 
Moreover, graph-based SLAM models each trajectory pose (i.e., position and heading) as a graph node and the odometry measurements as graph edges. 
Due to the odometry errors that typically characterize any robotic platform, the uncertainty in the poses grows as the drone moves~\cite{Chao10, huang2022edge}.
Hence, upon revisiting a location (i.e., loop closure), the pose error is higher than at the initial visit.
To mitigate this issue, the robot also acquires environmental observations (i.e., depth measurements) during the flight~\cite{zhou2022efficient}.
By comparing the observations associated with two different poses, an accurate rigid body transformation can be derived between the two, using an approach called scan-matching~\cite{zhou2022efficient}.

While the transformation provided by scan-matching allows correcting the current pose, graph-based SLAM propagates this information back to the previously added nodes in the graph (i.e., graph optimization) and corrects the whole trajectory~\cite{grisetti2010tutorial}.
In conclusion, the corrected trajectory depends on the accuracy of the scan-matching and, therefore, on the observations' precision~\cite{srinara2021performance}.
In most common applications, the observations consist of depth measurements, typically provided by LiDARs or stereo cameras~\cite{eyvazpour2022hardware, shen2022pgo}.
Although SLAM paired with LiDARs is widely used in applications with standard-size UAVs, these solutions require large amounts of computational resources and memory, which are not available on nano-UAVs~\cite{zhou2022efficient, karam2022micro}.
Furthermore, even the most compact LiDARs used with standard-size UAVs are about one order of magnitude heavier\footnote{The Craziflye 2.1 weights \qty{27}{\gram} and supports a maximum payload of \qty{15}{\gram}, while a lightweight LiDAR such as UST-10/20LX from Hokuyo weighs \qty{130}{\gram}.} than the maximum payload of nano-UAVs~\cite{chen2023self}.

The recent release of lightweight (i.e., \SI{42}{\milli\gram}), low-resolution, and energy-efficient depth sensors based on Time of Flight (ToF) technology has changed the status quo in the feasibility of SLAM for nano-UAVs~\cite{niculescu2022towards}. With the aid of such sensors, recent works demonstrated SLAM on nano-UAVs, but only under the assumption that the complex SLAM could be offloaded to an external base station~\cite{zhou2022efficient}.
This approach reduces the flight time due to the significant power consumption introduced by the radio communication with the base station~\cite{niculescu2021improving}.
Even more serious issues are the latency associated with the wireless communication protocol and the limited radio link range, which typically constrains the operating area within a few tens of meters in indoor environments~\cite{niculescu2021improving}.
Furthermore, because of the limited measurement capabilities of the early ToF sensors (i.e., a single distance value per sensor and a narrow FoV)~\cite{duisterhof2021tiny}, the existing systems that enable SLAM with nano-UAVs can only map simple-geometry environments such as long flat corridors.
In contrast to the existing works, we exploit the VL53L5CX ToF sensor, which features an 8$\times$8 resolution and provides a 64-pixel depth map with a Field of View (FoV) of \ang{45}.
By mounting four such sensors on the nano-UAV (i.e., front, rear, left, right), we achieve a cumulative FoV of \ang{180}.
Furthermore, spinning the drone by \ang{45} results in a full angular coverage (i.e., \ang{360}), providing superior loop-closure performance compared to the previous ToF-based solutions and achieving centimeter-precision scan-matching accuracy, similar to the LiDAR-based approaches.

Despite the sparse information provided by the ToF sensors, scan-matching remains a computationally intense and memory-hungry problem.
Furthermore, the computational requirements are further exacerbated by the graph optimization performed by the graph-based SLAM, which is independent of the depth observations.
Standard-size UAV systems used for SLAM typically employ powerful embedded computers such as Qualcomm Snapdragon, Nvidia Jetson TX2, or Xavier~\cite{shen2022pgo}, which have a power consumption of a few tens of watts, about two orders of magnitude higher than the power budget nano-UAVs typically have for computation.
Recent trends in microcontroller design emphasize parallel processing, hardware accelerators, and energy efficiency. 
The GAP9 System on Chip (SoC) from GreenWaves Technologies\footnote{\url{https://greenwaves-technologies.com/}} exemplifies these trends, being suited for specialized applications with nano-UAVs. 
With advanced parallel capabilities provided by the RISC-V cores, power optimization, and sensor integration, GAP9 empowers nano-UAVs with real-time edge computing, extended flight times, and enhanced data processing.
GAP9 is based on the Parallel Ultra-Low-Power (PULP) computing paradigm~\cite{rossi2021vega}, has a small form factor, and a power consumption below~\qty{180}{\milli\watt}.

This paper proposes NanoSLAM, the first fully deployable framework that enables SLAM onboard nano-UAVs, performing the whole computation and environmental perception without relying on any external infrastructure or computation offload. Furthermore, by exploiting novel and low-power depth sensors in combination with the parallel capabilities of GAP9 SoC, our system achieves accurate indoor mapping comparable with SoA results from bigger and more computationally capable drones, commonly referred to as Micro-Aerial Vehicles (MAVs) or standard-size UAVs~\cite{wang2020deep}. Exploiting the parallel capability and energy efficiency of GAP9, we executed scan-matching and SLAM in real-time onboard the nano-UAV, which was not performed by any previous work.
The contribution of this paper can be summarized as follows: \begin{enumerate*}[label=(\roman*),,font=\itshape]
\item An optimized parallel implementation of the graph-based SLAM algorithm that runs in the GAP9 SoC in real-time in less than \qty{250}{\milli\second}. We comprehensively examine the various stages of the SLAM algorithm, providing an in-depth analysis of the optimizations made to each stage. Additionally, we evaluate the algorithm's execution time and memory requirements.
\item A parallel implementation and evaluation of the Iterative Closest Point (ICP) algorithm, an SoA in scan-matching, running onboard in \qty{55}{\milli\second}. 
\item A custom plug-in companion board for the commercial Crazyflie 2.1 nano-UAV that extends the sensing capabilities of the drone with 4 ToF matrix sensors, allowing it to perform scan-matching and autonomous navigation.
\item A communication protocol that orchestrates the integration and data exchange between the drone's stock MCU and the GAP9 SoC, dictating how to store that graph, exchange graph poses, add edges, and perform graph optimization.
\item An extensive in-field experimental evaluation that proves our system's closed-loop mapping functionality, which exploits NanoSLAM to achieve a trajectory error reduction by up to 67\% and a mapping accuracy of \qty{4.5}{\centi\meter}.
\end{enumerate*}
\section{Related Work} \label{sec:related}
In the field of robotics, several essential components are indispensable for facilitating autonomous navigation on diverse unmanned vehicles. These components encompass real-time environment perception~\cite{muller2022robust}, onboard computational capabilities for prompt mission inference~\cite{muller2023fully,cerutti2020sound}, and, pertinent to the focus of this paper, the competence to map and explore unknown environments~\cite{dilshad2022locateuav}.
Mapping an environment is generally done by employing different combinations of sensors~\cite{huang2022edge}, such as LiDARs, stereo cameras, laser scanners, or radars. Subsequently, environmental and spatial information collected from these sensors is paired with estimation methods, including particle filters~\cite{muller2023fully}, Extended Kalman Filters (EKFs), covariance intersection that enables position estimation, and, finally, SLAM~\cite{li2023crowdfusion} that combines the position information with environmental observations to generate a layout of the environment. As discussed in the literature~\cite{latif2019slam,placed2023survey}, SLAM consists of two components: the front-end processing represented mainly by feature extraction and loop closure, which is largely dependent on the sensors used, and the sensor-agnostic trajectory and map optimization, in charge of the back-end processing~\cite{suzuki2020time}.

Based on the back-end optimization type, SLAM algorithms are classified as filtering-based and graph-based~\cite{placed2023survey}.
Filtering-based SLAM typically employs a Particle Filter (PF) that incorporates odometry and depth measurements (e.g., from LiDAR) as they come and updates the robot's pose and map in an online fashion~\cite{montemerlo2002fastslam}.
The PF iterates between a prediction and a correction step and estimates the probability distribution of the pose using a set of particles, each particle representing a pose belief.
The prediction state updates the state probability distribution by propagating the set of particles through the motion model of the robot.
The correction step relies on the measurement model of the sensor -- a mathematical formula indicating the relation between the sensor measurement and the robot pose -- and updates the likelihood associated with each particle.
Despite their widespread usage, particle filters remain susceptible to tuning issues and they only update the pose estimate based on the previously acquired measurements.
On the other hand, the graph-based SLAM solutions model the robot poses as graph nodes and the sensor measurements as edges and optimize the graph to determine the optimal node values~\cite{grisetti2010tutorial}.
The optimization implies minimizing a cost function through an iterative process, employing the first-order Taylor expansion, and reducing the optimization problem to a linear equation system.
In graph-based SLAM, the most challenging task lies in solving the equation system, as conventional methods, like matrix inversion, impose significant resource demands, exhibiting a time complexity of $O(N^3)$ and a space complexity of $O(N^2)$~\cite{saad2003iterative}.
Graph-based SLAM solutions are the most accurate, as they estimate the full trajectory of the robot from the complete set of
measurements~\cite{grisetti2010tutorial}.

As the name suggests, visual SLAM (vSLAM) uses images to extract depth information~\cite{cao2022edge}. 
It can use simple monocular cameras (e.g., wide angle, fish-eye, and spherical cameras), compound eye cameras (e.g., stereo and multi cameras), and RGB-D cameras such as depth or ToF sensors~\cite{cao2022edge}. 
While SLAM can be enabled at a low cost with relatively inexpensive and limited cameras, the process involves large data volumes and is often marred with limited mapping accuracy~\cite{kasper2019benchmark}. 
On the other side, LiDARs are more precise for depth estimation and are commonly used for applications involving high-speed moving vehicles such as self-driving cars and drones~\cite{causa2022uav}. 
The SLAM front-end also uses LiDARs or other distance sensors to detect loop closures and correct the robot's pose through scan-matching -- categorized into scan-to-map and scan-to-scan~\cite{zou2021comparative}. 
In the former, the laser scans are aligned with a global map to mitigate the robot's trajectory drift, while the latter computes relative pose transformations directly by overlapping two laser scans. 
The computation time of the scan-matching approaches increases with the size of the laser scans, which could pose challenges in LiDAR-based applications due to their dense output.
Although LiDARs yield accurate mapping results when paired with SLAM, they are generally expensive and heavy, weighing a few hundred grams~\cite{causa2022uav}.

Today, SLAM is useful in many applications~\cite{placed2023survey} such as navigating a fleet of mobile robots to arrange shelves in a warehouse~\cite{cristiani2020inventory}, parking self-driving cars in empty spots, autonomous race competitions~\cite{Ghignone2023}, or delivering packages by navigating drones in unknown environments~\cite{karam2022micro}. 
Many available tools already provide plug-and-play SLAM solutions that could be paired with other tasks such as sensor fusion, object tracking, path planning, and path following~\cite{Ghignone2023}. 
Although the mapping task seems to be a solved research problem in the literature, it relies on strong assumptions, such as memory availability of several gigabytes and powerful processor, e.g., the Intel i7 family~\cite{causa2022uav,chang2022lamp}. 
Moreover, carrying heavy and power-hungry 3D scanners, such as stereo cameras and LiDARs, is not considered a limitation for conventional robotic applications~\cite{causa2022uav,chang2022lamp,Ghignone2023}. 
However, these assumptions do not hold for miniaturized and low-power robotic platforms, where the hardware cost is a concern, the payload is limited to a few tens of grams, and the computation power budget is limited to hundreds of \qty{}{\milli\watt}~\cite{niculescu2021improving,muller2023fully,mayer2020smart}. Hence, enabling onboard mapping on this tiny class of devices is still an open problem. 
Furthermore, onboard computation has demonstrated a remarkable resilience against malicious threats, effectively mitigating the vulnerabilities and privacy concerns associated with communication-based attacks~\cite{mekdad2023survey}. 
Since UAVs are sensor-driven devices, they present susceptibility to malicious attacks, such as Global Navigation Satellite System (GNSS) wireless link data jamming or injection of false sensor data into cameras and Inertial Measurement Units (IMUs)~\cite{mekdad2023survey}.
Thus, to improve security, it is paramount for miniaturized robots to not rely on any external infrastructure for perception and computation.

This paper focuses on nano-UAVs as a specific application scenario to empirically validate the efficacy of our lightweight NanoSLAM approach. However, the challenges discussed in enabling mapping on nano-UAVs can be extended to the broader domain of micro-robotics and, more generally, to low-cost and resource-constrained devices~\cite{alghamdi2021architecture}.
Standard-size UAVs distinguish themselves from MAVs and nano-UAVs in their physical dimensions, weight, total power consumption, and onboard processing capabilities~\cite{alghamdi2021architecture}. 
For the latter two, the sensing and processing power budget represents about $\frac{1}{10}$ of the power consumed by the motors~\cite{niculescu2021improving}. 
Presently, the majority of cutting-edge advancements in robotic perception and mapping have been showcased on standard-size UAVs and MAVs, which possess a power budget ranging from \SI{50}{\watt} to \SI{100}{\watt} and a total mass of $\geq$ \SI{1}{\kilo\gram}~\cite{zhou2023racer}. 
Consequently, these vehicles can be equipped with high-performance onboard computing platforms, such as GPUs featuring gigabytes of memory \cite{zhou2023racer}. 
Conversely, nano-UAVs, typically based on low power MCUs, weigh less than \qty{50}{\gram} with a power budget in the range of \qty{5}{\watt} -- \qty{10}{\watt}, with only \qty{0.5}{\watt} -- \qty{1}{\watt} being allocated for powering the sensors, all the electronics,  and the computational units~\cite{niculescu2021improving,alghamdi2021architecture}. 
Low-power MCUs usually offer limited memory capacity, typically ranging from \qty{100}{\kilo\byte} to \qty{1}{\mega\byte}, posing a significant constraint for visual-based perception and mapping~\cite{niculescu2021improving,muller2023fully}.
\begin{table*}[t]
    \centering
    \begin{tabular}{ >{\centering\arraybackslash}m{1.5cm} >{\centering\arraybackslash}m{2.6cm} >{\centering\arraybackslash}m{2.5cm} >{\centering\arraybackslash}m{2cm} >{\centering\arraybackslash}m{1.3cm} >{\centering\arraybackslash}m{1.4cm} >{\centering\arraybackslash}m{1.6cm} >{\centering\arraybackslash}m{1.0cm}} 
        \hline
        \hline
        Work & Onboard processing & Sensor & Latency & Map accuracy & Field test & Power consumption & System weight  \\
        \hline
        \multicolumn{8}{c}{Nano-UAV and MAV}\\
        
        \hline
        \textbf{This work} & Yes (GAP9) & $4\times$ ToF 64-pixel VL53L5CX & \qty{247}{\milli\second} & 4-\qty{8}{\centi\meter} &  Yes & \qty{350}{\milli\watt} & \qty{44}{\gram} \\
        
        \cite{karam2022micro} & No & $4\times$ ToF VL53L1x & Post-processing & 10-\qty{20}{\centi\meter} & Yes & - & \qty{27}{\gram}\\ 
        \cite{duisterhof2021tiny} & Yes (Cortex-M4) & $4\times$ ToF VL53L1x & $<$\qty{10}{\milli\second} & No map & Yes & \qty{240}{\milli\watt} & \qty{31.7}{\gram}\\
        \cite{zhou2022efficient} & No (Intel i7 station) & $4\times$ ToF VL53L1x & \qty{214}{\milli\second} & 5-\qty{15}{\centi\meter} & No & - & \qty{31.7}{\gram} \\
        \cite{karam2022microdrone} & No & $4\times$ ToF VL53L1x & Post-processing & \qty{4.7}{\centi\meter} & No & - & \qty{401}{\gram}\\
        \hline
        \multicolumn{8}{c}{Standard-size UAV}\\
        
        \hline
        \cite{causa2022uav} & No & LiDAR & - & 5-\qty{20}{\centi\meter}  & Yes & - & \qty{3.6}{\kilo\gram}\\
        
        \cite{shen2022pgo} & Yes (Xavier) & VLP-16 LiDAR & \qty{49}{\milli\second} & \qty{2.14}{\meter} & No & \qty{30}{\watt} & $>$\qty{2}{\kilo\gram} \\
        
        \cite{huang2022edge} & Yes (Jetson TX2 ) & RP-LiDAR & $\sim$\qty{1}{\second} & - & Yes & $>$\qty{10}{\watt} & $>$\qty{2}{\kilo\gram} \\
        \cite{chang2022lamp} & No (Intel i7 station) & LiDAR & Post-processing & 15-\qty{20}{\centi\meter} & Yes & - & -\\
        \cite{zhou2023racer} & Yes (Jetson TX2) & Intel RealSense D435 & $\sim$\qty{120}{\milli\second} & - & Yes & \qty{7.5}{\watt} & \qty{1.3}{\kilo\gram} \\
        \hline
        \hline
    \end{tabular}
    \caption{System and performance comparison between this paper and the State-of-the-Art (SoA) works present in the literature. Onboard processing, sensing elements,  mapping accuracy, and system setups are compared.
    \label{tab:related}}
\end{table*}
Previous studies conducted on MAVs and UAVs have commonly utilized miniature, conventional \ang{360} LiDAR sensors~\cite{jeong2022parsing} or depth stereo cameras \cite{zhou2023racer} to perform mapping. 
For instance, Kumar \textit{et al.}~\cite{kumar2017lidar} integrated single-layer LiDAR sensors with inertial measurement units for indoor mapping tasks using a DJI Phantom 3 drone. 
This setup required an additional desktop-class Intel i5 processor onboard. 
The LiDAR sensor employed measures \qtyproduct{62 x 62 x 87.5}{\milli\meter}, weighs \qty{210}{\gram}, and consumes approximately \qty{8.4}{\watt}. 
Similarly, Gao \textit{et al.}~\cite{gao20190flying} integrated a multi-layer LiDAR sensor with a desktop-class Intel i7 processor to enable 3D mapping of indoor environments. 
The LiDAR sensor they use consumes \qty{8}{\W} and measures \qtyproduct{103 x 103 x 72}{\mm} with a weight of \qty{509}{\g}. 
Another approach by Fang \textit{et al.}~\cite{fang2017robust} uses an RGBD camera combined with a particle filter to navigate through obstructed shipboard environments. 
Their platform is \qtyproduct{58 x 58 x 32}{\cm} in size, carries over \qty{500}{\g} of instrumentation, and is operated by a high-performance octa-core ARM processor. 
Table~\ref{tab:related} provides an overview of SoA mapping strategies in the UAV field, encompassing sensor types, mapping accuracy, and power consumption of the computing platforms. 
For example, Causa \textit{et al.}~\cite{causa2022uav} proposed a scalable mapping strategy based on LiDAR and GNSS, utilizing a standard-size UAV weighing \qty{3.6}{\kilo\gram} with off-board processing. 
Shen \textit{et al.}~\cite{shen2022pgo} focused on onboard intelligence, utilizing a power-hungry Nvidia Xavier (\qty{30}{\watt}) and a VLP-16 LiDAR. 
Huang \textit{et al.}~\cite{huang2022edge} entrusted the mapping algorithm and onboard processing to a Jetson TX2, equipped with a multi-core CPU and a GPU. 
Additionally, Chang \textit{et al.}~\cite{chang2022lamp} proposed a robust multi-robot SLAM system designed to support swarms, but the results were validated offline using an Intel i7-8750H processor. 
Although these approaches demonstrated good mapping capabilities in the range of 5 to \qty{20}{\centi\metre}, they involve large and heavy sensors that require power-intensive processing.
Implementing SLAM on nano-UAVs or any miniaturized and low-power hardware~\cite{han2022tinyml} is non-trivial due to the large memory and computation requirements typically associated with scan-matching or graph optimization.
Moreover, alternatives such as offloading heavy computation tasks to an external computer is often an unpractical solution. 
In~\cite{huang2022edge}, authors show how the communication latency of a cloud-based multi-robot SLAM solution can reach up to \qty{5}{\second}, an unacceptable value in most nano-UAV uses cases. 
The severe constraint imposed by continuous remote communication poses limits to the mapping speed and the overall system reliability~\cite{huang2022edge}, which further demonstrates the need for having a fully onboard SLAM even on resource-constrained nano-UAVs.  

One approach to address the computational challenge involves parallelizing different processes on ultra-low power parallel SoCs~\cite{han2022tinyml, rossi2021vega}.
Utilizing embedded accelerators or multicore MCUs for processing, leveraging Single Instruction Multiple Data (SIMD) calculations, can enhance performance in certain scenarios~\cite{tabanelli2023dnn,rossi2021vega}. 
To this end, novel PULP SoCs have emerged in recent years, offering clusters of cores within \qty{100}{\milli\watt} of power consumption. 
Rossi \textit{et al.}~\cite{rossi2021vega} present the basis of the commercial SoC GAP family from Greenwaves, which has already demonstrated its capabilities in the field of nano-UAVs for accurate localization~\cite{muller2023fully} and autonomous navigation~\cite{niculescu2021improving}. 
In particular, GAP9 is selected for the scope of this paper to carry the intensive computation.

To attain an optimal solution, the sensor selection needs to consider an optimal trade-off between power consumption, accuracy, and weight.
In \cite{kuhne2022parallelizing}, the authors explore the possibility to use visual-based perception to enable obstacle avoidance and mapping on nano-UAVs. 
However, today, this direction does not seem to be promising due to the low performances of miniaturized RGB cameras and the large amounts of data they generate -- which needs to be processed by resource-constrained processors~\cite{tijmons2017obstacle,niculescu2021improving}. 
In their work \cite{tijmons2017obstacle}, Tijmons \textit{et al.} propose a stereo vision-based obstacle avoidance system for a flapping wing UAV, which demonstrates promising results with an onboard processing frequency of \SI{15}{\hertz}. 
This approach aligns with common methodologies employed in standard-size UAVs. 
However, their implementation needs an additional microcontroller (i.e., STM32F405) exclusively dedicated to image processing and the sensor board alone requires an energy consumption of \SI{484}{\milli\watt}. 
It is worth noting that while the authors of \cite{tijmons2017obstacle} tested their system in real environments, they do not report any statistical analysis of the success rate. 
Furthermore, the authors acknowledge the limited robustness of their system in non-ideal flight conditions, such as the presence of small obstacles. 
Another practical example is provided by~\cite{niculescu2021improving}, where the authors introduce a grayscale camera-based navigation solution that is deployed onboard a nano-UAV to facilitate autonomous navigation and obstacle avoidance. 
A CNN is used for perception and exhibits reliable performance in detecting obstacles, allowing the drone to adjust its forward velocity or heading. 
However, in unfamiliar environments, particularly when executing \ang{90} turns, the CNN's performance drops drastically, resulting in a high probability of collision when the drone exceeds \qty{0.6}{\meter/\second}~\cite{niculescu2021improving}. 
Additionally, their solution often struggles to avoid collisions with unknown obstacles placed in narrow environments such as corridors. 
Thus, vision-based approaches are not optimal solutions to enable onboard depth estimation with nano-UAVs, which is why we employ sensors that directly measure the depth. 
Since the commercially available LiDAR exceeds the power and weight constraints of pocket-size UAVs, alternatives have been investigated. 
Recent studies have shown potential in enabling autonomous navigation with depth sensors based on the ToF technology. 
In~\cite{niculescu2022towards}, the authors investigate the possibility of using a commercial multi-zone ToF sensor that exhibits good measurement accuracy when measuring distances smaller than \qty{2}{\meter}. 
Moreover, \cite{muller2022robust} used a lightweight 64-pixel ToF sensor for robust obstacle avoidance in indoor and outdoor scenarios, with a maximum speed of \qty{1.5}{\meter/\second}. 
At the time of writing, two commercially available depth sensors stand out: the VL53L5CX from ST Microelectronics and the ToF IRS2381C REAL3 from Infineon. The latter boasts an impressive resolution of 38,000 pixels and a maximum range of 4 meters. However, it requires an external illuminator, consumes up to \qty{680}{\milli\watt} for the entire circuitry, and has a weight exceeding \qty{10}{\gram}. 
On the other hand, the VL53L5CX offers a lower resolution of 64 pixels but is significantly lighter, weighing only \qty{42}{\milli\gram}. 
Additionally, its prior utilization in the nano-UAV field~\cite{muller2023fully, niculescu2022towards} serves as a compelling motivation for selecting it for this paper.
As depicted in Table~\ref{tab:related}, the existing literature offers only a limited number of studies proposing mapping solutions that use UAVs and have been successfully evaluated in field~\cite{huang2022edge}. 
Notably, \cite{chang2022lamp, zhou2023racer} achieve their objectives without relying on external infrastructure. 
However, within the nano-UAV domain, even fewer works tackle the mapping challenge~\cite {karam2022micro, zhou2022efficient, karam2022microdrone}, and they offload the computation to an external base station. 
Furthermore, the existing works performing mapping with nano-UAVs are not able to reach the same level of accuracy as standard-size UAVs within the literature. 

To the best of our knowledge, this paper introduces the first system that enables entirely onboard SLAM execution to enable accurate mapping of general environments, providing a comprehensive methodology, implementation, and field results. 
Our study demonstrates the system's functionality even with low-power miniaturized sensors that weigh only \qty{44}{\gram}. 
The achieved accuracy aligns with the SoA for MAVs and standard-size UAVs, with a mapping error down to \qty{4.5}{\centi\meter}. 
The proposed system facilitates advanced autonomous capabilities in nano-UAVs, paving the way for enabling additional features such as optimal path planning and multi-agent collaboration.

\section{Algorithms} \label{sec:algorithms}
This section presents a lightweight localization and mapping methodology that leverages the scan-matching and graph-based SLAM algorithms, targetting pocket-size robotic platforms and emerging low-power processors.
Our solutions can enable any robotic platform of similar size or bigger to perform low latency SLAM in real-time, given depth measurement capabilities enabled by sensors such as $8 \times 8$ STMicroelectronics VL53L8CX described in Section~\ref{sec:related}.

\subsection{Scan Frames and Scans} \label{sec:algo-scans}
\begin{figure} [t]
\begin{centering}
\includegraphics[width=\columnwidth]{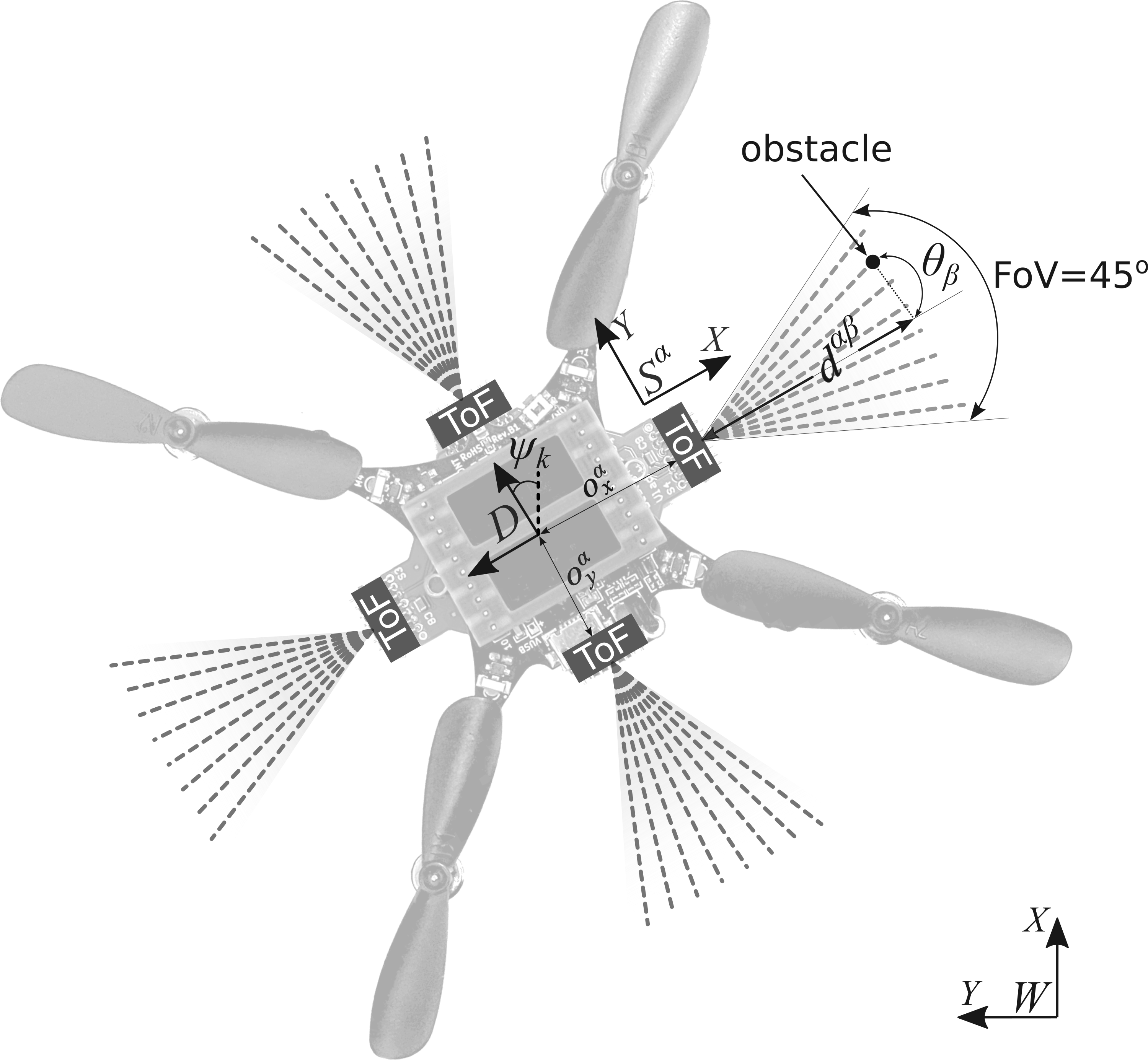}
\par\end{centering}
\centering{}
\caption{Illustration of the four ToF sensors onboard the drone and the coordinate frames of the world, drone, and sensors.}
\label{fig:drone}
\end{figure}

Our objective is to conduct 2D localization and mapping utilizing depth sensors. Therefore, we assume a system equipped with $n_{s}$ depth sensors (e.g., ToF) that provide measurements in the 2D plane with a resolution of $n_{z}$ pixels (i.e., zones) per sensor.
Figure~\ref{fig:drone} shows an example of such a system with $n_{s}=4$ and $n_{z}=8$, illustrating the drone, the ToF depth sensors, and how the distance measurements can be projected in the world frame. The world frame, body frame, and sensor frame are represented with $W$, $D$, and $S$, respectively.

Let $\vec{x}_k=(x_k,y_k,\psi_k)$ be the state of the drone (i.e., \textit{pose}) expressed in the world coordinate frame at the discrete timestamp $k$. 
Furthermore, we use $\alpha\in \{1,2,\ldots n_{s}\}$ to index among ToF sensors and $\beta\in \{1,2, \ldots n_{z}\}$ to index among the zones of each sensor. 
The distance provided by sensor $\alpha$ for the zone $\beta$ at instant $k$ is marked as $d_k^{\alpha\beta}$.
Equation~\ref{eq:scan-proj} shows the distance measurement projection $d_k^{\alpha\beta}$ acquired at pose $\vec{x}_k$ into the world coordinate frame $W$.
Indeed, the distance $d_k^{\alpha\beta}$ provided by the sensor is not the absolute distance to the object but the projection of the absolute distance on the $OX$ axis of the sensor frame $S^{\alpha}$. 
Thus, $tan(\theta_{\beta}) \cdot d_k^{\alpha\beta}$ represents the $y$-coordinate of the obstacle in the same sensor frame, where $\theta_{\beta}$ is the angle of each sensor zone. 
Translating the obstacle's coordinates to the origin of the drone's body frame $B$ and rotating it to the world frame $W$ leads to the second term of Equation~\ref{eq:scan-proj}.
The translation is performed by adding the offset $(o_x^{\alpha}, o_y^{\alpha})$ to the obstacle's position -- note that the offset is expressed in $D$, and it is different for each sensor.
$\mat{R}$ represents the 2D rotation matrix and the sum $\psi_k + \gamma_{\alpha}$ represents the angle between $S^{\alpha}$ and $W$, where $\psi_k$ is the heading angle between $D$ and $W$; $\gamma_{\alpha}$ represents the rotation of the sensor frame w.r.t $D$ and for the example in Figure~\ref{fig:drone},  $\gamma_{\alpha} \in \{ \SI{0}{\degree},  \SI{90}{\degree}, \SI{180}{\degree}, \SI{270}{\degree}\}$.
Lastly, we use the coordinates of the pose $(x_k, y_k)$ to perform another translation and obtain the coordinates of the obstacle expressed in the world frame $W$.
At every timestamp $k$, the ToF sensors provide at most $n_s n_z$ distance measurements -- $n_s$ sensors $\times$ $n_z$ zones -- as some distance measurements might be invalid and therefore not considered.
Projecting the $n_s n_z$ points using Equation~\ref{eq:scan-proj} leads to the collection $\{(p_{x}^{\alpha\beta}, p_{y}^{\alpha\beta}) \mid \alpha \leq n_{s}; \beta \leq n_{z} \}$ that we call a \textit{scan frame}.
\begin{equation} \label{eq:scan-proj}
\begin{pmatrix}
p_{x}^{\alpha\beta}(k)\\
p_{y}^{\alpha\beta}(k)
\end{pmatrix}
= 
\begin{pmatrix}
x_{k}\\
y_{k}
\end{pmatrix}
+
\mat{R}_{(\psi_k + \gamma_{\alpha})}
\begin{pmatrix}
d_k^{\alpha\beta} + o_{x}^{\alpha} \\
tan(\theta_{\beta}) \cdot d_k^{\alpha\beta} + o_{y}^{\alpha}
\end{pmatrix}
\end{equation}
The 2D point collection in the scan frame could be further used as input for the scan-matching algorithm.
However, the cardinality of a scan frame (i.e., the number of 2D points) is still too small to enable accurate scan-matching.
We overcome this issue by stacking $n_{sf}$ consecutive scan frames in a set that we call a \textit{scan} and define as $\mat{S_k}=\{(p_{x}^{\alpha\beta}(\tilde{k}), p_{y}^{\alpha\beta}(\tilde{k}))  \mid \alpha \leq n_{s}; \beta \leq n_{z};  k \leq \tilde k < k+n_{sf} \}$.
When the acquisition of a new scan is triggered, the robot starts appending new scan frames until it reaches the count of $n_{sf}$.
The resulting cardinality of a scan is $n_{sf} n_s n_z$ points (minus the invalid pixels), which we call \textit{the scan size}. 
Moreover, every scan $\mat{S}_k$ has an associated \textit{scan pose} $\vec{x}_k$, which is the drone's pose when the scan acquisition starts.

Let $f$ be the Field of View (FoV) of one ToF depth sensor, which leads to a cumulative FoV of $n_{s}f$, generally smaller than \ang{360}.
To virtually increase the FoV and achieve full coverage, the drone also spins by $\frac{360-n_s f}{n_s}$ degrees in place around the $z$-axis while acquiring the scan.
For example, given the scenario in Figure~\ref{fig:drone} and assuming an FoV of \ang{45} for each sensor, the drone should spin other \ang{45} during the scan to cover the surroundings completely.
With this mechanism, scan-matching can determine the transformation between two scans $\mat{S}_p$ and $\mat{S}_q$, which also applies to their associated scan poses $\vec{x}_p$ and $\vec{x}_q$.
The scan size is a balance between scan-matching accuracy and memory usage, determined by the limitations of the system.

\subsection{Scan-matching} \label{sec:algorithms-scan-matching}
Scan-matching is the process of determining the optimal rigid-body transformation between two scans.
This transformation consists of rotation and translation, and with an ideal noise-free scenario, it should result in perfect overlapping with the other scan.
Since scans and poses are strictly correlated, the transformation resulting from scan-matching also applies to the poses.
When the drone is near a previously visited position, scan-matching can derive an accurate transformation w.r.t. a previously acquired pose in that location.
In this way, scan-matching is used to correct the accumulated odometry errors.
In this work, we implement and use ICP, an SoA algorithm in scan-matching~\cite{gross2011point}.

We define two scans \(\mat{S_p} = \{ \vec{p}_1, \vec{p}_2, \ldots \}\) and \(\mat{S_q} = \{ \vec{q}_1, \vec{q}_2 \ldots \}\) where $\vec{p}_i$ and $\vec{q}_i$ are 2D points in the scans -- we changed the initial indexing to enhance readability. 
Determining the optimal overlap between $\mat{S}_p$ and $\mat{S}_q$ can be formulated as a least squares problem, as shown in Equation~\ref{eq:icp-obj}~\cite{gross2011point}.
Note that Equation~\ref{eq:icp-obj} requires to know what element $\vec{q}_i$ in scan $\mat{S}_q$ corresponds to the element $\vec{p}_i$ in scan $\mat{S}_p$.
If the correspondences are known, a direct and optimal solution can be obtained by solving the optimization problem in Equation~\ref{eq:icp-obj}.
This is typically done by offsetting each scan by its center of mass and then applying a rotational alignment based on the singular value decomposition method~\cite{gross2011point}.
\begin{equation} \label{eq:icp-obj} 
    \mat{R}^*,\vec{t}^* = \argmin_{\mat{R},\vec{t}} \sum{\lVert \vec{q}_i - (\mat{R} \vec{p}_i + \vec{t}) \rVert^2}~.
\end{equation}

However, the correspondences are unknown in our case and in most of real-world scan-matching applications.
A common heuristic for determining the correspondences is to use the Euclidean distance -- i.e., pairing each point $\vec{p}_i$ in $\mat{S}_p$ with the closest point $\vec{q}_j$ in $\mat{S}_q$~\cite{gross2011point}.
This implies solving the problem $\argmin_{j} \lVert \vec{p}_i - \vec{q}_j \rVert$ for every point $\vec{p}_i$, using an exhaustive search over all elements in $\mat{S}_q$.
Once these approximate correspondences are established, Equation~\ref{eq:icp-obj} determines the transformation between the two scans, which is then applied to $\mat{S}_p$. 
Repeating this process until the two scans overlap represents the ICP algorithm, which we summarize in Listing~\ref{lst:icp}.

%
\begin{listing}[t]
\AtBeginEnvironment{minted}{\linespread{1.4}}
\begin{minted}[mathescape=true,
             escapeinside=||,
             numbersep=5pt,
             gobble=2,
             fontsize=\small,
             framesep=2mm]{python}
             
for k in range(|$N_{iter}^{ICP}$|):
   # Compute correspondences
   for i in len(|$\mat{S}_p$|):
      correspondence[i] |$\leftarrow \argmin_{j} \lVert \vec{p}_i - \vec{q}_j \rVert$|
   # Calculate the transformation
   |$\mat{R}^*,\vec{t}^* \leftarrow$| Equation |$\ref{eq:icp-obj}$|
   # Apply transformation to scan $\mat{S}_p$
   |$\mat{S}_p \leftarrow  \mat{R}^* \mat{S}_p + \vec{t}^*$|
        
\end{minted}
\caption{The stages of the ICP algorithm. $N_{iter}^{ICP}$ represents the number of iterations and $\mat{R}^*,\vec{t}^*$ the final solution after the algorithm executes.} 
\label{lst:icp}
\end{listing}

\subsection{Graph-based SLAM Algorithm} \label{sec:algorithms-slam}
In most GPS-denied environments, such as indoor scenarios, the drone's internal state estimator computes the position and heading by integrating velocity and angular velocity measurements.
However, the measurements are affected by sensor noise, and integrating noisy data over time results in drift.
Equation~\ref{eq:scan-proj} shows that projecting distance measurements in the world frame to obtain a scan or the map requires trajectory knowledge.
Since the trajectory error impacts the accuracy of the map, we use SLAM to first correct the trajectory and then compute the map w.r.t. the corrected path.
For this purpose, we implement the graph-based SLAM introduced in \cite{grisetti2010tutorial}, which can use scan-matching information to correct the trajectory.
The graph-based SLAM represents the trajectory as a pose graph, where each pose (i.e., 2D position and heading) is modeled as a graph node, and the edges are relative constraints between the nodes.
We distinguish two types of graph edges: \textit{(i)} \textit{the odometry edges} incorporating motion information between any two consecutive poses, and \textit{(ii)} \textit{the loop closure (LC) edges} which embody relative measurements derived by ICP.

Let $N$ be the number of poses and $n$ the number of LC edges. 
Moreover, let $\mat{X} = \{ \boldsymbol{x}_0, \ldots, \boldsymbol{x}_{N-1}\}$ be the graph nodes expressed in $W$, and $\boldsymbol{z}_{ij}=(z_x,z_y,z_{\psi})$ the graph edge measurements, the latter being expressed in the coordinate frame of pose $\boldsymbol{x}_{i}$.
We note as $\vec{\hat z}_{ij}$ the prediction of an edge measurement, or in other words, the edge measurement computed given two poses $\vec{x}_{j}$ and $\vec{x}_{i}$.
Pose Graph Optimization (PGO) involves the estimation of optimal pose values that ensure consistency between the edge measurements $\vec{z}{ij}$ and the predicted measurements $\vec{\hat z}{ij}$.
As shown in \cite{grisetti2010tutorial}, this is done by minimizing the sum of the squared differences $\vec{e}_{ij}=\vec{z}_{ij} - \vec{\hat z}_{ij}$, where Equation~\ref{eq:slam-opt} gives the maximum likelihood solution that requires the initial pose $\boldsymbol{x}_{0}$ and the edges $\vec{z}_{ij}$ to compute the optimal poses.
The number of terms in the sum is equal to the number of edges in the graph, and $\Omega$ is the diagonal information matrix, which weighs the importance of each edge.
Since ICP typically provides accurate results, the LC edge measurements are more precise than the odometry edge measurements. 
\begin{align} \label{eq:slam-opt}
\vec{e}_{ij} &= \vec{z}_{ij} - \vec{\hat z}_{ij}(\vec{x}_{i}, \vec{x}_{j}) \nonumber~, \\
\mat{X}^* &= \argmin_{\mat{X}} \sum_{i,j} \vec{e}_{ij}^T \Omega \vec{e}_{ij}~.
\end{align}
\begin{listing}[t]
\AtBeginEnvironment{minted}{\linespread{1.4}}
\begin{minted}[mathescape=true,
             escapeinside=||,
             numbersep=5pt,
             gobble=2,
             fontsize=\small,
             framesep=2mm]{python}

# 1. Compute the odometry edge measurements
for i in range(|$N-1$|):
   |$\vec{z}_{i,i+1} \leftarrow $| Equations |$\ref{eq:meas_from_poses0} - \ref{eq:meas_from_poses1}$|
# 2. Compute the LC edge measurements
for k in range(n):
   |$\vec{z}_{ij} \leftarrow $| ICP(|$\vec{x}_i, \vec{x}_j$)|
# 3. Graph optimization
for k in range(|$N_{iter}^{SLAM}$|):
   # 3a. Compute $\mat{H}$ and $\vec{b}$
   |$\mat{H} \leftarrow \vec{0}$, $\vec{b} \leftarrow \vec{0}$, $\mat{H}_{11} \leftarrow \mat{I}_3$|
   for edge in edges:
      # Compute the Jacobians
      |$\mat{A}_{ij} \leftarrow \frac{\partial e_{ij}(\vec{x})}{\partial x_i}\bigg\vert _{\vec{x}=\vec{x}^*}$ $\mat{B}_{ij} \leftarrow \frac{\partial e_{ij}(\vec{x})}{\partial x_j}\bigg\vert _{\vec{x}=\vec{x}^*}$|
      # Construct the linear system matrix
      |$\mat{H}_{ii} += \mat{A}_{ij}^T \Omega \mat{A}_{ij}$ $\mat{H}_{jj} += \mat{A}_{ij}^T \Omega \mat{B}_{ij}$|
      |$\mat{H}_{ij} += \mat{A}_{ij}^T \Omega \mat{B}_{ij}$ $\mat{H}_{ji} += \mat{B}_{ij}^T \Omega \mat{B}_{ij}$|
      # Construct the linear system vector
      |$\vec{ b}_i += \mat{A}_{ij}^T \Omega \vec{e}_{ij}$ $\vec{b}_j += \mat{B}_{ij}^T \Omega \vec{e}_{ij}$|
   # 3b. Solve the linear system $\mat{H} \Delta\vec{x}=-\vec{b}$
   |Permutation $\mat{H}_P = \mat{P} \mat{H} \mat{P}^T$, $\vec{b}_P=\mat{P}\vec{b}$, $\Delta\vec{x}_P = \mat{P} \Delta\vec{x}$|
   # Solve $\mat{H}_P \Delta\vec{x}_P=-\vec{b}_P$
   |Cholesky decomposition $\mat{H}_P = \mat{L}_P\mat{L}_P^T$|
   |Forward substitution $\mat{L}_P \vec{y} = -\vec{b}_P$|
   |Backward substitution $\mat{L}_P^T \Delta\vec{x}_P = \vec{y}$|
   # Retrieve the solution $\Delta\vec{x}$
   |Inverse permutation $\Delta\vec{x} = \mat{P}^{-1} \Delta\vec{x}_P$|
   # 3c. Update the solution
   |$\vec{x}^* \leftarrow \vec{x}^* + \Delta\vec{x}$|
        
\end{minted}
\caption{The graph-based SLAM algorithm that performs PGO. The outer \textit{for} loop runs for $N_{iter}^{SLAM}$ iterations.} 
\label{lst:graph-based-slam}
\end{listing}

Running SLAM onboard a resource-constrained device in real-time requires solving the optimization problem in Equation~\ref{eq:slam-opt} efficiently.
Since this is a non-linear problem, there is no closed-form solution, but iterative methods such as Gauss-Newton have been proven effective if a good initial guess is known.
In every iteration, the error function $\vec{e}_{ij}(\vec{x}_{i}, \vec{x}_{j})$ is approximated with its first-order Taylor expansion, reducing the problem to a linear equation system.
This paper provides an efficient implementation of the graph-based SLAM algorithm derived from~\cite{grisetti2010tutorial}, which is in charge of PGO. 
We summarize the algorithm in Listing~\ref{lst:graph-based-slam} and discuss it in detail in Section~\ref{sec:implementation}.
\begin{align} \label{eq:meas_from_poses0}
\begin{pmatrix} z_x \\ z_y \end{pmatrix} &=
\mat{R}_{-\psi_i}
\begin{pmatrix} x_{i+1} - x_i \\ y_{i+1} - y_i \end{pmatrix}~, \\
z_{\psi} &= \psi_{i+1} - \psi_i ~.
\label{eq:meas_from_poses1}
\end{align}

The graph-based SLAM algorithm requires the initial pose $\vec{x}_0$, the edge measurements $\vec{z}_{ij}$, and an initial guess of the pose values.
The initial pose $\vec{x}_0$ is the drone's pose right after take-off, and without loss of generality, it is always considered $(0,0,0)^T$.
Since there is no additional information about the poses, the best initial guess is computed by forward integrating the odometry measurements w.r.t. $\vec{x}_0$.
Consequently, the poses' initial guess encompasses the same information as the odometry edge measurements, and therefore, it suffices to store only the poses.
This mechanism is convenient because, in many robotics applications, the robot's state estimator directly integrates the odometry measurements and provides the pose values.
In this way, the odometry edge measurements are calculated right before the optimization is performed, as shown in the first \textit{for} loop in Listing~\ref{lst:graph-based-slam}.
Each measurement $\vec{z}_{i,i+1}$ is expressed in a coordinate frame rotated by $\psi_i$ and computed using Equations~\ref{eq:meas_from_poses0} -- \ref{eq:meas_from_poses1}.
The second \textit{for} loop calculates the LC edges, using the ICP algorithm introduced in Section~\ref{sec:algorithms-scan-matching}.

Once all the edge measurements are calculated, the actual graph optimization can start, performed in the double \textit{for} loop.
Minimizing Equation~\ref{eq:slam-opt} when $e_{ij}(\vec{x}_i, \vec{x}_j)$ is linearized around the current pose guess is equivalent to solving the linear equation system $\mat{H} \Delta\vec{x}=-\vec{b}$~\cite{grisetti2010tutorial}.
$\mat{H}$ and $\vec{b}$ are computed in the inner \textit{for} loop.
$\mat{A}_{ij}$ and $\mat{B}_{ij}$ are $3\times3$ matrices and represent the Jacobians obtained after linearization. 
Similarly, the $3\times3$ blocks $\mat{H}_{ii}$, $\mat{H}_{ij}$, $\mat{H}_{jj}$, and $\mat{H}_{ji}$ represent the contribution on the $\mat{H}$ matrix of each graph edge from node $i$ to node $j$.
The dimension of the blocks $\vec{b}_i$ and $\vec{b}_j$ is $3\times1$, and they construct the system vector $\vec{b}$.
Given the constituent elements of matrix $\mat{H}$ and vector $\vec{b}$, their dimension is $3N\times3N$ and $3N\times1$, respectively.
The $3N\times1$ vector $\vec{x}^*$ serves as the ongoing estimate of the poses (stacked together), continuously refined during the iterative graph optimization process. 
Before the optimization starts, the initial guess is loaded into $\vec{x}^*$.

The next step is to solve the linear system $\mat{H} \Delta\vec{x}=-\vec{b}$.
Inverting matrix $\mat{H}$ would demand significant memory and computational resources, inefficient for resource-constrained devices~\cite{saad2003iterative}. 
Nonetheless, more efficient alternatives have been suggested in the literature, which leverage the Cholesky decomposition~\cite{touchette2016efficient, quintana2001note}.
Since $\mat{H}$ is symmetric positive-definite, the decomposition calculates the lower triangular matrix $\mat{L}$, such that $\mat{H} = \mat{L} \mat{L}^T$.
The equation system becomes $\mat{L} \mat{L}^T \Delta\vec{x}=-\vec{b}$.
In addition, we make the notation $\vec{y} = \mat{L}^T \Delta\vec{x}$.
Since $\mat{L}$ is triangular, solving $\mat{L} \vec{y}=-\vec{b}$ is trivial using the forward substitution method.
Having $\vec{y}$, the solution $\Delta\vec{x}$ is easily calculated by solving $\vec{y} = \mat{L}^T \Delta\vec{x}$ using backward substitutions.
Lastly, the solution $\Delta\vec{x}$ is added to the current estimate of poses.
Typically, the outer loop iterates until $\Delta\vec{x}$ reaches a sufficiently small value or becomes zero.

The ordering of the rows and columns of matrix $\mat{H}$ influences the non-zero count and computation time of matrix $\mat{L}$~\cite{saad2003iterative}.
Permuting both the rows and columns of $\mat{H}$ is done by the multiplication $\mat{P} \mat{H} \mat{P}^T$, where $\mat{P}$ is the permutation matrix -- i.e., an identity matrix with reordered rows~\cite{saad2003iterative}.
Exploiting the property $\mat{P}^{-1}=\mat{P}^T$, the linear system is rewritten as $\mat{H} \mat{P}^T \mat{P}\Delta\vec{x}=-\vec{b}$. 
Multiplying both sides by $\mat{P}$ on the left and making the substitutions $\mat{H}_P = \mat{P} \mat{H} \mat{P}^T$, $\vec{b}_P=\mat{P}\vec{b}$, and $\Delta\vec{x}_P = \mat{P} \Delta\vec{x}$ leads to $\mat{H}_P \Delta\vec{x}_P=-\vec{b}_P$.
Lastly, $\Delta\vec{x}$ is retrieved from $\Delta\vec{x}_P$, which implies a negligible overhead.
Therefore, applying the permutation leads to the same mathematical problem, requiring the additional step of retrieving $\Delta\vec{x}$ from $\Delta\vec{x}_P$, which comes with negligible overhead. 
The process of solving $\mat{H} \Delta\vec{x}=-\vec{b}$ leveraging the Cholesky decomposition and the permutation mechanism is described in step 3b of Listing~\ref{lst:graph-based-slam}.
In Section~\ref{sec:implementation}, we discuss how the permutation matrix $\mat{P}$ is obtained.

\subsection{SLAM in Real-world Scenarios} \label{sec:algorithms-slam-ex}
\begin{figure} [t]
\begin{centering}
\includegraphics[width=\columnwidth]{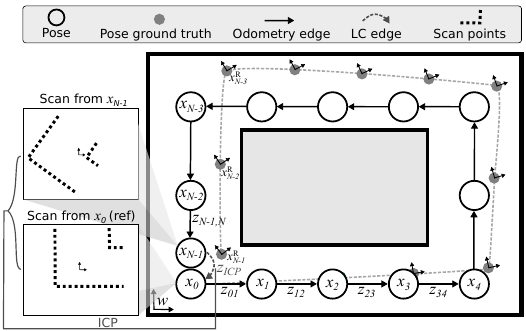}
\par\end{centering}
\centering{}
\caption{The figure shows how a robot trajectory is discretized into a pose graph when passing through a square loop corridor. The drone keeps adding poses to the graph using information from the internal state estimator. The poses are affected by errors due to odometry drift.}
\label{fig:slam_icp}
\end{figure}

Figure~\ref{fig:slam_icp} shows how a robot trajectory can be discretized into a pose graph.
In this example, the drone flies along a square loop corridor, following the outer wall until it reaches the start point again.
As the drone advances, it keeps adding new poses to the graph at fixed intervals using the information provided by the internal state estimator.
The pose $\vec{x}_0$ is the starting point and, therefore, error-free, but since the following poses are obtained based on integration w.r.t. $\vec{x}_0$, they are affected by errors due to odometry drift.
We note the poses as $\vec{x}_0, \vec{x}_1, \dots \vec{x}_{N-1}$ and represent them with empty circles in Figure~\ref{fig:slam_icp}, while the odometry constrains $\vec{z}_{01}, \vec{z}_{12}, \dots$ are the edges connecting the circles.
Performing the graph optimization at this point would lead to no change in the poses because any pose $\vec{x}_{i+1}$ is obtained by integrating the measurement $\vec{z}_{i,i+1}$ w.r.t. $\vec{x}_{i}$ and therefore the poses and edge measurements are already in agreement -- i.e., the sum from Equation~\ref{eq:slam-opt} is already zero.

In Figure~\ref{fig:slam_icp}, the filled grey circles denoted as $\vec{x}_0^R, \vec{x}_1^R, \dots$, represent the actual (i.e., the ground truth) position and heading of the poses, which are not known to the drone.
At the end of the mission, even if the drone estimates that it crosses the starting point again (i.e., $\vec{x}_0 = \vec{x}_{N-1}$), its actual pose is $\vec{x}_{N-1}^R$.
To mitigate the odometry errors, the drone acquires observations (i.e., a scan) in $\vec{x}_{N-1}$, which it compares with the scan acquired in $\vec{x}_0$, as shown in Figure~\ref{fig:slam_icp}.
We call as \textit{reference scan} the scan acquired when a place is visited for the first time -- e.g., the scan acquired in $\vec{x}_0$.
Furthermore, we define as \textit{LC scan} the scan acquired when a place is revisited -- e.g., the scan acquired in $\vec{x}_{N-1}$.
An LC scan is always paired with a reference scan or another LC scan, and ICP is used to derive a transformation between the two.
In the example from Figure~\ref{fig:slam_icp}, ICP is used to derive a transformation between $\vec{x}_0$ and $\vec{x}_{N-1}$, and therefore add a new LC edge to the graph -- from node $N-1$ to node $0$.
Once there is at least one new LC edge in the graph, graph-based SLAM can run to correct the existing poses.
After the optimization completes, the LC edges are typically kept in the graph.
In the context of our approach, we assume an unchanging environment. 
Yet, should alterations occur within the environment that lead to scans that do not overlap, we identify these situations and discard the LC edge.
The procedure for quantifying the degree of overlap between two scans is elaborated upon in Section~\ref{sec:results}.

\subsection{Optimizing Large Graphs} \label{sec:algorithms-hierarchical}
\begin{figure} [t]
\begin{centering}
\includegraphics[width=\columnwidth]{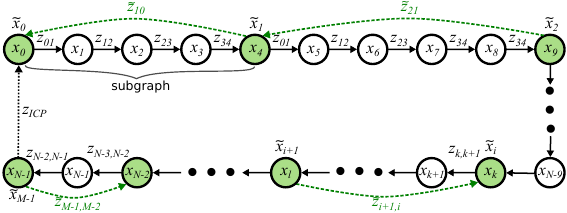}
\par\end{centering}
\centering{}
\caption{Representation of the sparse graph (green) as a subset of the complete graph. The black arrows represent the odometry edges, the dashed black arrow represents the LC edge, and the dashed green arrows represent the additional constraints derived for optimizing the subgraphs.}
\label{fig:multi-graph}
\end{figure}

\begin{figure} [t]
\begin{centering}
\includegraphics[width=\columnwidth]{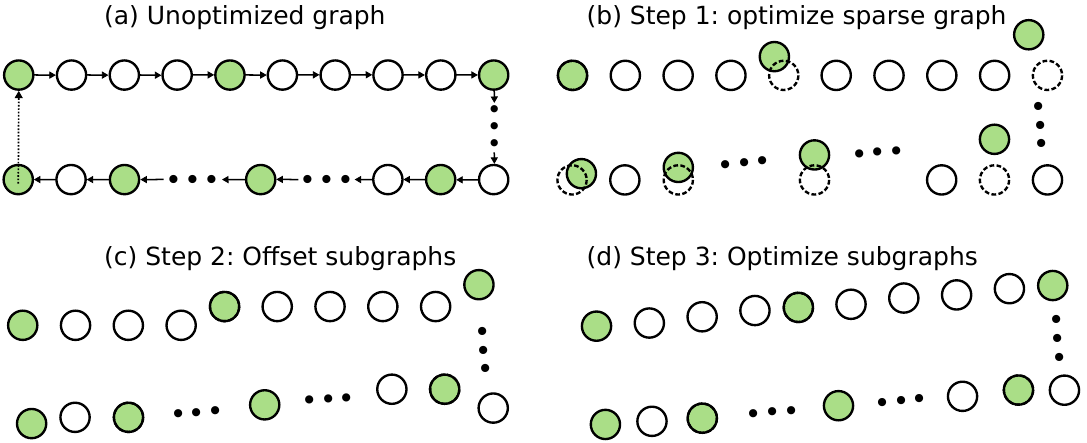}
\par\end{centering}
\centering{}
\caption{A breakdown of the hierarchical graph-based SLAM.}
\label{fig:multi-graph-steps}
\end{figure}

The elements of graph-based SLAM were presented in a simple example in Figure~\ref{fig:slam_icp}, but they are representative of any graph and any number of poses or constraints.
However, optimizing graphs larger than a few hundred poses with this method might be challenging because embedded platforms are typically constrained to a few hundred of \qty{}{\kilo\byte} of RAM. 
To address this problem, we implement a solution based on the hierarchical optimization approach introduced in~\cite{grisetti2010hierarchical}.
The idea is to divide the graph into multiple subgraphs and apply the graph-based SLAM algorithm from Listing~\ref{lst:graph-based-slam} on each subgraph -- we refer to this approach as hierarchical graph-based SLAM.
For this purpose, a \textit{sparse graph} $\mat{\tilde{X}} = \{ \vec{\tilde{x}}_0, \ldots, \vec{\tilde{x}}_{M-1}\}$ is created first, whose poses are a subset (but still representative) of the complete graph $\mat{X}$.
We mention that the poses marked with a tilde are just an alternative notation for the poses already present in $\mat{X}$ to emphasize that we are referring to the sparse graph.
We provide a graphical representation of such a hierarchical optimization problem in Figure~\ref{fig:multi-graph}, where the poses of the sparse graph  are represented in green.
Furthermore, Figure~\ref{fig:multi-graph-steps} shows a four-step breakdown of the hierarchical optimization.
Using the scan-matching constraints (e.g., $z_{ICP}$ in Figure~\ref{fig:slam_icp}), the sparse graph is optimized, resulting in the new set of poses $\{ \vec{\tilde{x}}_1^{opt}, \ldots, \vec{\tilde{x}}_M^{opt}\}$ -- as shown in Figure~\ref{fig:multi-graph-steps}-(b).
The idea now is to use the optimized poses of the sparse graph as constraints to correct the entire graph $\mat{X}$.

For each pair of consecutive poses in the sparse graph $(\vec{\tilde{x}}_{i}^{opt}, \boldsymbol{\tilde{x}}_{i+1}^{opt})$ we build a subgraph consisted of these poses and the in-between poses of the complete graph -- e.g., $\{ \vec{x}_0, \vec{x}_1, \ldots, \vec{x}_4\}$ or $\{ \vec{x}_4, \vec{x}_5, \ldots, \vec{x}_9\}$.
To be more general, we consider the subgraph $\{ \vec{x}_{k}, \vec{x}_{k+1}, \ldots, \vec{x}_{l} \}$.
We recall that the $\vec{x}_{k} = \vec{x}_{i}$ and $ \vec{x}_{l} = \vec{x}_{i+1}$.
Since we have already corrected the extremes (i.e., $\vec{\tilde{x}}_i^{opt}$ and $\vec{\tilde{x}}_{i+1}^{opt}$), this information can be further used to derive the constraint that allows optimizing the whole subgraph.
In this scope, we firstly offset every pose in the subgraph as shown in Figure~\ref{fig:multi-graph-steps}-(c), so that $\vec{x}_{k} / \tilde{\vec{x}}_i$ matches $\vec{\tilde{x}}_i^{opt}$ -- necessary because PGO never corrects the first pose.
Then Equations~\ref{eq:meas_from_poses0} -- \ref{eq:meas_from_poses1} are used to derive a constraint $\tilde{\vec{z}}_{i+1,i}$ between poses $\vec{\tilde{x}}_{i+1}^{opt}$ and $\vec{\tilde{x}}_{i}^{opt}$, which is added to the subgraph as an LC edge from node $l$ to $k$ -- this only simulates the effect of loop closure, as the LC edge is not provided by ICP directly.
After these operations are performed on the $M-1$ subgraphs as shown in Figure~\ref{fig:multi-graph-steps}-(d), the optimization of $\mat{X}$ is complete.
This section, therefore, introduces two manners of performing PGO: directly applying graph-based SLAM on the existing pose graph or dividing the graph into multiple smaller subgraphs and optimizing every subgraph individually.
The advantages of every approach are discussed in Section~\ref{sec:implementation}.

Sampling the poses of the sparse graph from the complete graph $\mat{X}$ is based on a threshold on the robot movement.
The elements comprising the complete graph are chronologically traversed, and a new node is exclusively incorporated into the sparse graph if the Euclidean distance from the most recently added node exceeds a threshold value $d_{min}$ or if the difference in heading surpasses a threshold value $\Delta \psi_{min}$.
The sparse graph must also include all scan poses as a mandatory requirement, in addition to the threshold-based added poses. 
This is because the LC edges resulting from ICP only play a role in optimizing the sparse graph and not also the subgraphs.
However, the number of scan poses is usually negligible compared to the sparse graph size.

\section{Nano-UAV System Setup} \label{sec:background}
Our mapping system is designed to be flexible and cover a large set of robotic platforms. 
The only prerequisite concerns the sensor, which has to be a depth camera. 
Thus the algorithm and the implementation can be adapted to support a different hardware setting, e.g., various processors or sensing elements. 
In this paper, we selected the Commercial-Off-The-Shelf (COTS) nano-UAV Crazyflie 2.1 from Bitcraze to demonstrate the effectiveness of our solution in ultra-constrained platforms. 
In this way, our results can be easily replicated using commercially available hardware.

\begin{figure}[t]
    \centering
    \begin{subfigure}{\columnwidth}
    \centering
    \includegraphics[width=1\linewidth]{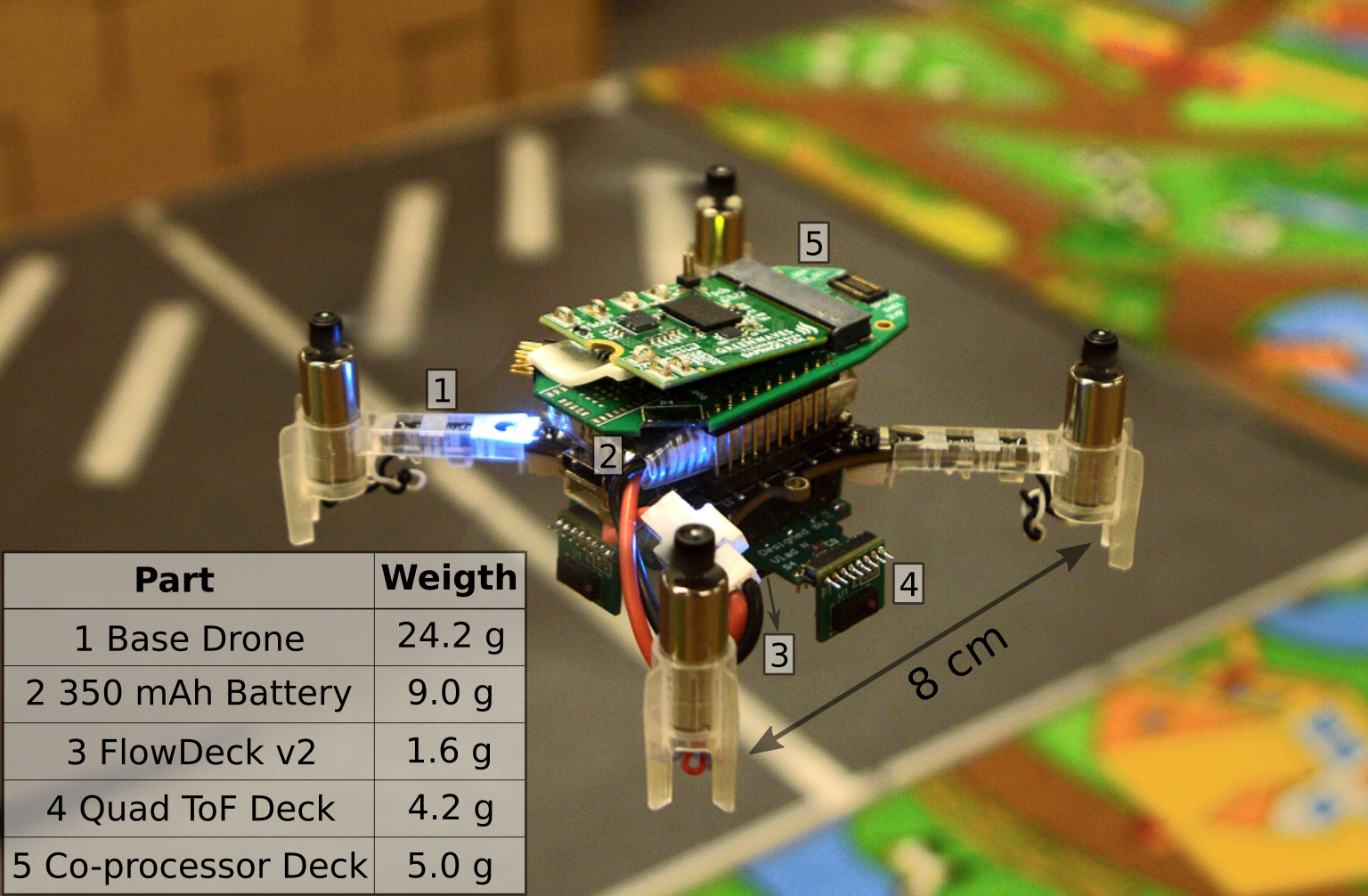}
    \caption{The fully integrated system.
    \label{fig:drone-full}}
    \end{subfigure}
    \begin{subfigure}{0.49\columnwidth}
    \centering
    \includegraphics[width=1\linewidth]{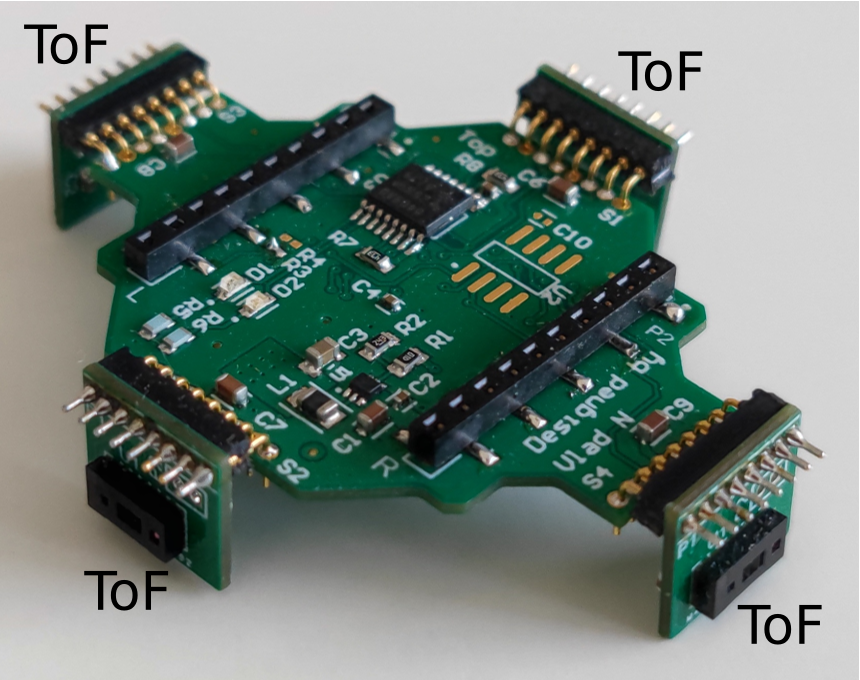}
    \caption{The custom quad ToF deck.
    \label{fig:tof-deck}}
   \end{subfigure}
   \begin{subfigure}{0.49\columnwidth}
   \centering
   \includegraphics[width=1\linewidth]{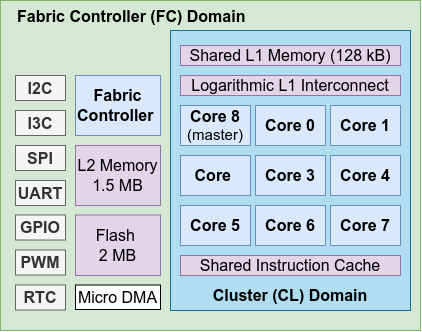}
   \caption{Architecture of the GAP9.
   \label{fig:gap9-diag}}
   \end{subfigure}
\caption{(a) Our prototype based on Crazyflie 2.1 extended with the ToF Deck and the Co-processor Deck. (b) The custom quad ToF deck featuring four ToF multi-zone sensors. (c) A simplified diagram showing the blocks of the GAP9 that are most relevant for this work. 
\label{fig:hardware}}
\end{figure}

The open-source firmware of Crazyflie 2.1 provides capabilities for flight control, state estimation, radio communication, and setpoint commander. 
The drone's main PCB also acts as a frame, comprising the electronics such as an IMU, a radio transceiver (Nordic nRF51822), and an STM32F405 processor. The latter features a maximum clock frequency of \qty{168}{\mega\hertz} and \qty{192}{\kilo\byte} of RAM, but over 70\% of the resources are already used by the firmware to perform the control and estimation.
Furthermore, the drone features extension headers that can be used to add additional decks (i.e., plug-in boards). We, therefore, also included the commercial Flow deck v2, which exploits a downward-facing optical flow camera and single-zone ToF ranging sensor to enable velocity and height measurements fused by the onboard EKF to perform position and heading estimation.
In addition to the Flow deck, we equip the drone with two custom-designed boards: one containing four lateral depth ToF sensors to enhance the drone's capabilities to sense the surroundings, and the second deck~\cite{muller2023fully} contains the GAP9 SoC, used as a co-processor to extend the Crazyflie 2.1 computation capabilities. 
In this configuration, the total weight at take-off is \qty{44}{\g}, including all the hardware used for the scope of this paper.
The fully integrated system featuring our custom hardware is shown in Figure~\ref{fig:drone-full}.

\subsection{Custom Quad ToF Deck}
The VL53L5CX is a lightweight multi-zone 64-pixel ToF sensor, weighing only \qty{42}{\milli\gram}. 
Its suitability for nano-UAV applications was evaluated in a study by Niculescu \textit{et al.}~\cite{niculescu2022towards}. 
This sensor offers a maximum ranging frequency of \qty{15}{\Hz} for an 8$\times$8 pixel resolution, with a FoV of \ang{45}. 
Additionally, the VL53L5CX provides a pixel validity matrix alongside the 64-pixel measurement matrix, automatically identifying and flagging noisy or out-of-range measurements. 
To accommodate the use of multi-zone ranging sensors on the Crazyflie 2.1 platform, a custom deck was developed specifically for the VL53L5CX ToF sensors, as shown in Figure~\ref{fig:tof-deck}. 
This deck can be used in conjunction with the Flow deck v2 and incorporates four VL53L5CX sensors positioned to face the front, back, left, and right directions, enabling obstacle detection from a cumulative FoV of \ang{180}. 
As a result, the final design of the custom deck weighs a mere \qty{4.2}{\g}.

\subsection{Co-processor Deck - GAP9 SoC}
The second custom deck included in the system setup weighs \qty{5}{\gram} and features the GAP9 SoC, the commercial embodiment of the PULP platform~\cite{rossi2021vega}, produced by Greenwaves Technologies. 
Figure~\ref{fig:gap9-diag} shows the main elements of the GAP9 architecture.
The GAP9 SoC features 10 RISC-V-based cores, which are grouped into two power and frequency domains. The first domain is the Fabric Controller (FC), which features a single core operating at up to \qty{400}{\mega\hertz} coupled with \qty{1.5}{\mega\byte} of SRAM (L2 memory). 
The FC acts as the supervisor of the SoC, managing the communication with the peripherals and orchestrating the on-chip memory operations. 
The second domain is the CLuster (CL) consisting of nine RISC-V cores that can operate up to \qty{400}{\mega\hertz}, specifically designed to handle highly parallelizable and computationally intensive workloads.
Among the nine cores of the cluster, one acts as a ``master core'', receiving a job from the FC and delegating it to the other eight cores in the cluster, which carry the computation. 
The CL cores share the same instruction cache, allowing them to efficiently execute the same code on different data.
The CL is coupled with \qty{128}{\kilo\byte} of L1 memory (shared by the CL cores), and the transfers between L2 and L1 are performed via the Direct Memory Access (DMA) peripheral, requiring no involvement from the FC or CL during the transfers.
In addition, the CL also features a neural engine that provides hardware-level acceleration for performing operations such as convolutions, batch normalization or ReLU activations, supporting the execution of quantized deep learning models.
To achieve an optimal execution time of a CL task, the data associated with the task should be transferred to L1 before the task is started.
When the CL task completes, the result can be transferred back to L2 and further used by the FC.
The GAP9 is interfaced with the STM32 via SPI and carries all the intensive computation required by PGO and scan-matching.
\section{Implementation} \label{sec:implementation}

Our system features two computational units: the STM32 MCU, part of the commercial Crazyflie 2.1 platform, and the more powerful GAP9 SoC, which extends the computational capabilities of the former.
We extend the base firmware of the STM32 with our application -- implemented through the Bitcraze application layer -- containing only lightweight functionalities such as the ToF sensor data acquisition and the flight strategy, which have a negligible impact on the MCU load. 
Instead, we delegate the memory and computationally demanding tasks to the GAP9, which continuously communicates with the STM32 during the mission.
Thus, computationally intensive solutions such as ICP, the graph-based SLAM, scan computation, or map generation run entirely on the GAP9.
In the following, we provide the implementation details of NanoSLAM, which is based on the algorithms introduced in Section~\ref{sec:algorithms}.

\subsection{Sensor Processing}
As mentioned before, our system performs mapping in 2D.
However, since each of the four ToF sensors provides an 8$\times$8 distance matrix, we must process this information and reduce it to one plane (i.e., one row).
For this reason, we discard the first two rows from the bottom and the top, leaving only the middle four rows that better represent the drone's plane.
In the following, we select the median of the four remaining pixels for each column, obtaining a row vector of size eight for each ToF sensor.
In case there are no valid pixels in a particular column (e.g., no obstacle within \qty{4}{\meter}), the entire column is discarded.
This approach ensures more robustness to outliers than simply selecting one of the middle rows from each matrix.

\subsection{Scan-matching Implementation}
Before detailing the actual scan-matching implementation, we provide the values of the scan parameters introduced in Section~\ref{sec:algo-scans}.
Indeed, our setup matches the configuration shown in Figure~\ref{fig:drone}, featuring four ToF sensors of eight zones each -- i.e., $n_s=4$ and $n_z=8$.
During a scan, the drone undergoes a \qty{45}{\degree} rotation while adding new scan frames to the scan with a frequency of \qty{7.5}{\hertz}.
We empirically choose $n_{sf}=20$ as a trade-off between scan-matching accuracy and memory footprint, resulting in a scan duration of about \qty{2.7}{\second}.
Given these settings, the scan size is at most $n_{scan}=n_{sf} n_s n_z=640$ points.

We recall that the ICP algorithm introduced in section~\ref{sec:algorithms-scan-matching} has two stages: determining the correspondences and calculating the transformation given the correspondence pairs.
The latter exhibits a time complexity of $O(n_{scan})$ and is typically very fast.
The correspondences calculation, represented by the inner \textit{for} loop in Listing~\ref{lst:icp}, takes more than 95\% of execution time, operating with $O(n_{scan}^2)$ complexity.
Furthermore, since the correspondences are calculated independently of each other, we leverage the parallel capabilities of GAP9, distributing the inner \textit{for} loop from Listing~\ref{lst:icp} to eight cores of the CL in GAP9.
In our implementation, we choose a fixed number of iterations $N_{iter}^{ICP}$ to ensure a deterministic execution time. 
We empirically determined with in-field experiments that ICP always converges within $N_{iter}^{ICP}=25$ iterations, and after that, the solution $(\mat{R}^*,\vec{t}^*)$ does not change anymore.

\subsection{Graph-based SLAM Implementation} \label{sec:implementation-slam}
In the following, we provide the implementation details of the graph-based SLAM algorithm, which is presented in Listing~\ref{lst:graph-based-slam}.
Having introduced how ICP is implemented, we now focus on step 3 from Listing~\ref{lst:graph-based-slam}, the heart of graph-based SLAM.
Once all the odometry constraints are computed, each iteration of the algorithm consists of two main phases: \textit{(i)} calculating $\mat{H}$ and $\vec{b}$, and \textit{(ii)} solving the equation system $\mat{H} \Delta \vec{x} = -\vec{b}$.
The main challenge is to enable the onboard execution, given the limited available amount of RAM.
Storing all entries of the $3N \times 3N$ $\mat{H}$ matrix would result in about \qty{1.44}{\mega\byte} for a realistic pose number of 200 and a 4-byte float representation of the matrix entries. 
This requirement is infeasible for resource-constrained platforms -- even for our capable target platform, GAP9, which would rapidly run out of memory storing such a matrix.
\begin{figure}[t]
    \centering
    \begin{subfigure}{0.50\columnwidth}
    \centering
    \includegraphics[width=1\linewidth]{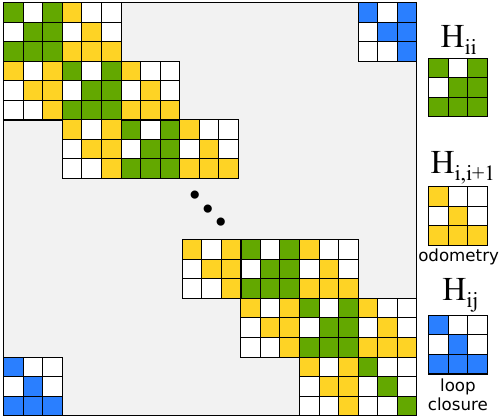}
    \caption{The filling pattern of $\mat{H}$.
    \label{fig:h-mat-struct}}
   \end{subfigure}
    \begin{subfigure}{0.45\columnwidth}
    \centering
    \includegraphics[width=1\linewidth]{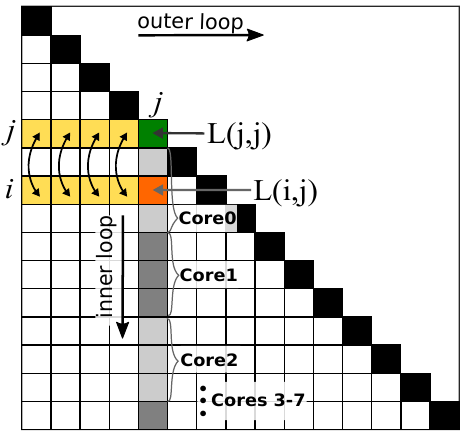}
    \caption{The $\mat{L}$ matrix.
    \label{fig:cholesky-filling}}
    \end{subfigure}
\caption{(a) The figure shows how the odometry and LC edges differently impact the sparsity of the $\mat{H}$ matrix. (b) The computation of the Cholesky decomposition and how each element's calculation is distributed among the CL cores. 
\label{fig:matrix-structure}}
\end{figure}

However, as Listing~\ref{lst:graph-based-slam} shows, constructing matrix $\mat{H}$ implies looping through all edges and modifying the blocks $\mat{H}_{ii}$, $\mat{H}_{ij}$, $\mat{H}_{ji}$, and $\mat{H}_{jj}$, for each graph edge from $i$ to $j$.
Due to the highly accurate results offered by ICP, we experimentally selected an information matrix $\Omega=20\mat{I}$ for the LC edges and $\Omega=\mat{I}$ for the odometry edges. 
This deliberate choice assigns greater significance to the LC edges during the optimization process.
The number of odometry edges (i.e., $N-1$) is typically much larger than the number of LC edges, and for most of the blocks $\mat{H}_{ij}$, it holds that $j=i+1$.
Thus, most non-zero elements of $\mat{H}$ are concentrated around the main diagonal.
Figure~\ref{fig:h-mat-struct}, provides a graphical representation of the $\mat{H}$ matrix, where the contribution of blocks $\mat{H}_{ii}$ and $\mat{H}_{i+1,i+1}$ is represented in green, and the contribution of $\mat{H}_{i,i+1}$ and $\mat{H}_{i+1,i}$ in yellow.
Blocks in blue correspond to the LC edges, and their placement in the matrix does not follow a pattern.

Furthermore, by calculating the individual elements of each block with the equations from Listing~\ref{lst:graph-based-slam}, one could notice that some elements are always zero and represented in white in Figure~\ref{fig:h-mat-struct}.
This fact increases, even more, the sparsity of $\mat{H}$, resulting in $17N-13+10n$ non-zero elements according to the filling pattern of Figure~\ref{fig:h-mat-struct}.
Moreover, due to the fact that $\mat{H}$ is symmetric, it is sufficient only to store the elements below and including the main diagonal, implying $10N-5+5n$ non-zero elements.
As a numerical example, for the realistic values of $N=200$ poses and $n=10$ LC constraints, the ratio between non-zero elements and the total number of elements $3N \times 3N$ is about  0.56\%, which proves that it is extremely memory inefficient to store a matrix in a dense form.

To exploit the high sparsity level, we propose storing $\mat{H}$ in a CSR sparse matrix representation~\cite{saad2003iterative}; we note the non-zero element count as $nz$.
This representation uses three arrays: \textit{(i)} \textit{the values}: it has size $nz$ and stores the non-zero elements; \textit{(ii)} \textit{the column index}: it has size $nz$ and stores the column index associated with each value in the values array; \textit{(iii)} \textit{the row pointer}: it has size $3N+1$, and its elements mark where in the values array a new row starts.
Inserting a new element in the CSR matrix requires modifying the three arrays accordingly.
Our software implementation solely utilizes static memory allocation to prevent memory leaks and overflows. Consequently, inserting elements in the sparse matrix must occur row-wise, in the ascending order of column indices -- in this way, the arrays of the sparse matrix are never modified, only extended. 
Otherwise, a random insertion order would imply memory moves within the sparse matrix, slowing the execution. 
Since the filling pattern of $\mat{H}$ is deterministic given the graph, an ordered element insertion is possible.
\begin{listing}[t]
\AtBeginEnvironment{minted}{\linespread{1.4}}
\begin{minted}[mathescape=true,
             escapeinside=||,
             numbersep=5pt,
             gobble=2,
             fontsize=\small,
             framesep=2mm]{python}
# Outer loop: iterating over columns              
for j in range(N):
   |$sum0 = \sum_{k=0}^{j-1} \mat{L}^2(j,k)$|
   |$\mat{L}(j,j) = \sqrt{\mat{H}(j,j) - sum0}$|
   # Inner loop: iterating over rows  
   for i in range(j+1, N):
      |$sum1 = \sum_{k=0}^{j-1} (\mat{L}(i,k) \cdot \mat{L}(j,k))$|
      |$\mat{L}(i,j) = (\mat{H}(i,j) - sum1) / \mat{L}(j,j)$|
        
\end{minted}
\caption{The Cholesky–Crout algorithm that performs the Cholesky decomposition column-by-column.} 
\label{lst:cholesky}
\end{listing}

The stages of graph-based SLAM exhibit a computational complexity not exceeding $O(N)$, except for the Cholesky decomposition. 
To leverage the parallel capabilities of the system, we employ the Cholesky-Crout scheme~\cite{saad2003iterative}. This scheme efficiently computes the matrix $\mat{L}$ column by column, as outlined in Listing~\ref{lst:cholesky}. To better illustrate the distribution of computation across eight cores of the GAP9 CL, we complement Listing~\ref{lst:cholesky} with Figure~\ref{fig:cholesky-filling}.
In each iteration for column $j$, the algorithm initially calculates the variable $sum0$, which represents the sum of squared elements from line $j$, excluding the diagonal. 
In Figure~\ref{fig:cholesky-filling}, these elements are visually depicted by the upper yellow line, with each CL core responsible for computing the sum of $j/8$ elements. 
The value of $\mat{L}(j,j)$ (highlighted in green) is subsequently determined based on $sum0$, and afterward, all column elements are computed within the inner loop. 
To offload the computation of the inner loop, we employ the CL, where each core performs the calculation for a predetermined number of $\mat{L}(i,j)$ entries. 
Each element $\mat{L}(i,j)$ (in orange) is derived from $sum1$, which is computed as the dot product between row $i$ and row $j$ -- considering only the elements to the left of column $j$.

It is imperative to note the direct dependence of $\mat{L}(i,j)$ on $\mat{H}(i,j)$, signifying that any non-zero element below the main diagonal in matrix $\mat{H}$ will correspondingly yield a non-zero element in matrix $\mat{L}$ with identical indices. 
Furthermore, each element $\mat{L}(i,j)$ is contingent upon the elements located to its left within the same row. 
Consequently, the existence of a non-zero element $\mat{H}(i,j)$ implies the existence of a non-zero element $\mat{L}(i,j)$, which, in turn, can influence the non-zero status of all subsequent elements in the same row of $\mat{L}$.
\begin{figure} [b]
\begin{centering}
\includegraphics[width=\columnwidth]{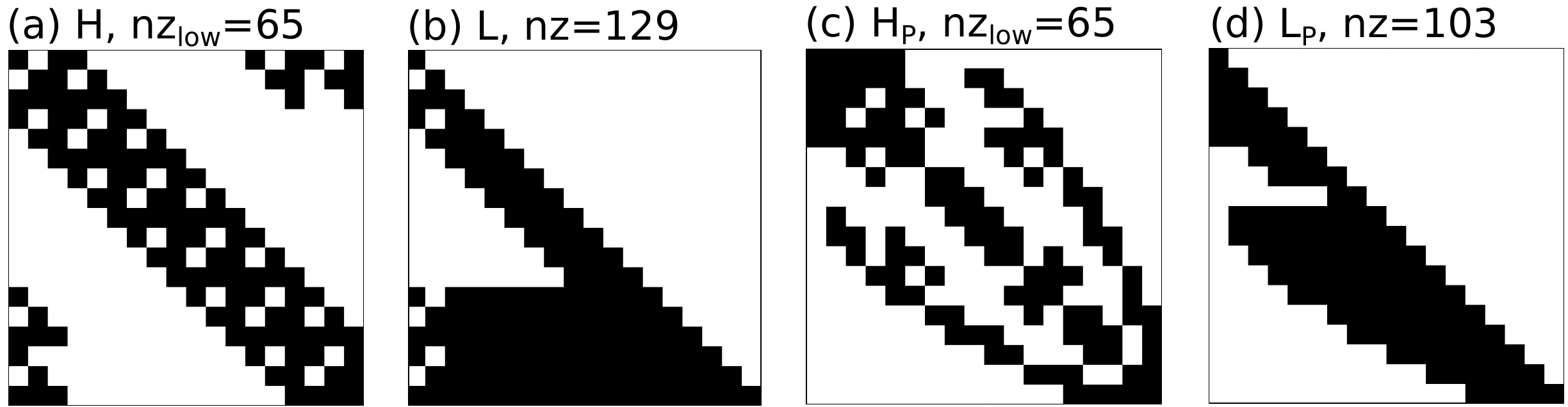}
\par\end{centering}
\centering{}
\caption{An example of the $\mat{H}$ and $\mat{L}$ matrices for a graph with six poses and two LC edges (left) and how these matrices change when the RCM permutation is applied to $\mat{H}$ (right).}
\label{fig:rcm}
\end{figure}
\begin{table} [t]
\centering
\caption{The non-zero count of $\mat{H}$, $\mat{L}$ and $\mat{L}_P$. For the symmetric $\mat{H}$ matrix, only the non-zero elements from the main diagonal or lower are counted.}
\label{tab:cholesky-nz}
\renewcommand{\arraystretch}{1.4} 
\begin{tabular}{ >{\arraybackslash}m{0.2cm} >{\centering\arraybackslash}m{0.44cm} >{\centering\arraybackslash}m{0.44cm} >{\centering\arraybackslash}m{0.44cm} > {\centering\arraybackslash}m{0.44cm} >{\centering\arraybackslash}m{0.44cm} >{\centering\arraybackslash}m{0.44cm} >{\centering\arraybackslash}m{0.44cm} >{\centering\arraybackslash}m{0.55cm} >{\centering\arraybackslash}m{0.55cm}} 
\hline\hline
                             &             & \textbf{20} & \textbf{80} & \textbf{140} & \textbf{200} & \textbf{260} & \textbf{320} & \textbf{380} & \textbf{440}  \\ 
\hline
\multirow{3}{*}{\rotatebox[origin=c]{90}{non-zero}} & \textbf{H}  & 205         & 805         & 1405         & 2005         & 2605         & 3205         & 3805         & 4405          \\
                             & \textbf{L}  & 549        & 2349        & 4149         & 5949         & 7749         & 9549         & 11349         & 13149          \\
                             & \textbf{Lp} & 336         & 1479        & 2355         & 3405         & 4391         & 4985         & 6458         & 7475          \\
\hline\hline
\end{tabular}
\end{table}

To analyze a concrete example, Figure~\ref{fig:rcm}-(a) illustrates the $\mat{H}$ matrix of a graph with six poses and two LC edges, where the black entries are the non-zero elements.
For the symmetric $\mat{H}$ matrix, the figure provides $nz_{low}$, which is the non-zero count of the elements below or on the main diagonal.
Figure~\ref{fig:rcm}-(b) illustrates the $\mat{L}$ matrix obtained from the Cholesky decomposition of $\mat{H}$.
The non-zero entries originating from the LC edges in the corner of matrix $\mat{H}$ resulted in corresponding non-zero elements in matrix $\mat{L}$, subsequently leading to non-zero elements throughout the entire row to the right.
Although one might argue that the LC edges have a negligible impact on the non-zero count of matrix $\mat{H}$, this is obviously not the case for matrix $\mat{L}$.
The further the non-zero elements of $\mat{H}$ are from the main diagonal, the more non-zero entries they will yield in $\mat{L}$.

To mitigate this problem, we employ the permutation solution introduced in Section~\ref{sec:algorithms-slam}, which brings the elements of $\mat{H}$ closer to the main diagonal. 
The Reverse Cuthill–McKee (RCM) algorithm~\cite{azad2017reverse} computes the permutation vector $\vec{\pi}$ that defines the rearrangement of the rows of the identity matrix to obtain the permutation matrix $\mat{P}$.
The $\mat{P}$ obtained through RCM minimizes the bandwidth of a given matrix -- i.e., how spread apart the elements are from the main diagonal.
Note that accessing any element $\mat{H}_P(i,j)$ is equivalent to accessing $\mat{H}(\vec{\pi}(i), \vec{\pi}(j))$ and therefore it is not necessary to store and compute matrix $\mat{H}_P$.
Similarly, $\vec{b}_P(i)=\vec{b}(\vec{\pi}(i))$.

Applying this algorithm to the example $\mat{H}$ matrix from Figure~\ref{fig:rcm}-(a) leads to the permuted $\mat{H}_P$ shown in 
Figure~\ref{fig:rcm}-(c), whose bandwidth is visibly reduced.
Applying the Cholesky decomposition to $\mat{H}_P$ leads to the matrix $\mat{L}_P$ from Figure~\ref{fig:rcm}-(d) with 103 non-zero entries, about 20\% less than $\mat{L}$.
Furthermore, Table~\ref{tab:cholesky-nz} provides the resulting non-zero count of $\mat{L}$ and $\mat{L}_P$ for a graph with two LC edges, varying the number of poses in the range 20 -- 440.
The lower bound (20) is the smallest number of poses for which the sparse Cholesky decomposition is faster than the dense implementation, and 440 represents the maximum number of poses that can be optimized at once, as we explain in Section~\ref{sec:performance}.
Also, in this case, the reduction in the non-zero count of $\mat{L}$ after permutation is at least 37\%.
Our system uses the RCM algorithm to determine the permutation matrix $\mat{P}$ and then solve the linear system as described in Listing~\ref{lst:graph-based-slam}.
All steps involved in calculating $\Delta x$ constitute a graph-based SLAM iteration.
We empirically determined that the entries of $|\Delta\vec{x}|$ are always smaller than $10^{-4}$ after three iterations, so we set $N_{iter}^{SLAM} = 3$.

\subsection{Hierarchical SLAM Implementation}
In Section~\ref{sec:algorithms-hierarchical}, the hierarchical graph-based SLAM method was introduced as an alternative approach to performing PGO, allowing to optimize graphs that exceed 440 poses. 
This approach utilizes the parameters $d_{min}$ and $\Delta \psi_{min}$, which determine the inclusion of new poses in the sparse graph based on the robot's movement and rotation. 
Typically, in exploration scenarios, significant variations in the heading are rare, as the robot primarily rotates when encountering walls or obstacles. 
As a result, the parameter $d_{min}$ has the greatest impact on the size of the sparse graph.

We investigate the influence of the $d_{min}$ parameter on the accuracy of the optimized graph. 
Increasing the value of $d_{min}$ reduces the size of the sparse graph, allowing for the mapping of larger environments. 
However, this also leads to a loss in capturing fine drone movements, resulting in decreased accuracy. 
To assess this impact, we conducted an experiment using a square loop corridor, creating an associated graph with 2000 poses, which significantly exceeds the limits of the graph-based SLAM algorithm when executed onboard.
We varied the $d_{min}$ parameter within the range of \qty{0.1}{\meter} to \qty{6.4}{\meter}, as detailed in Table~\ref{tab:hierarchical}, and measured the accuracy of the resulting optimized poses for each case. 
To evaluate the accuracy, we computed the Pearson correlation coefficient between the optimized poses obtained using the hierarchical approach with varying $d_{min}$ values and the poses derived from directly applying graph-based SLAM (performed on an external base station).
As an additional metric, we also compare the Root-Mean-Squared-Error (RMSE) w.r.t. the directly optimized graph, considering only the $x$ and $y$ components of each pose.
Table~\ref{tab:hierarchical} shows that values of $d_{min}\leq 0.8$ provide almost the same accuracy as optimizing the graph directly with graph-based SLAM, leading to a correlation coefficient larger than 99\% and an RMSE smaller than \qty{1}{\centi\meter}.
Consequently, any $d_{min}$ smaller than \qty{0.8}{\meter} is appropriate for creating the sparse graph.
\begin{table} [t]
\centering
\caption{Hierarchical graph-based SLAM. \label{tab:hierarchical}}

\begin{tabular}{lccccccc} 
\hline\hline
$\vec{d_{min} (m)}$ & \textbf{0.1} & \textbf{0.2} & \textbf{0.4} & \textbf{0.8} & \textbf{1.6} & \textbf{3.2} & \textbf{6.4} \\ 
\hline
Correlation (\%)    & 99.9   & 99.9   & 99.9   & 99.7   & 98.9   & 96.7   & 93.4 \\
RMSE (\qty{}{\centi\meter})    & 0.04   & 0.12   & 0.27   & 0.59   & 24.3   & 46.7   & 84.0 \\

\hline\hline
\end{tabular}
\end{table}

\subsection{The Exploration Strategy and Corner Detection}
\label{sec:implementation-exploration}
In the following, we explain how the drone explores the environment and decides which areas are appropriate for acquiring scans.
We mention that our mapping solution is completely independent of the type of trajectory and applicable to any environment.
Thus, to demonstrate the capabilities of our system, we use a simple \textit{exploration strategy} that drives a drone through the environment, always following the wall on the right.
In case of no walls around, the drone moves forward.
If a frontal wall or obstacle is detected, the drone changes direction to the left or right, depending on which direction is free -- if both are free, the drone chooses left.
Conversely, if a dead-end is detected, the drone lands, and the mission ends.
This exploration strategy is configured with the target velocity of \qty{0.5}{\meter/\second} -- correlated with the size of the rooms we explore.

Since our system should work autonomously in any environment, it must possess the ability to determine when a new scan should be acquired.
Regions that exhibit rich textures, such as corners, are highly suitable for acquiring scans that facilitate precise scan-matching. 
To achieve this objective, we implemented a corner detector that takes a scan frame as input and utilizes the Hough transform~\cite{mukhopadhyay2015survey} to identify all the straight lines defined by the points within the scan frame.
The presence of any pair of lines that create an angle of at least \qty{30}{\degree} indicates that the scan frame represents a corner.
The corner detector runs in the STM32 in less than \qty{1}{\milli\second}.
\subsection{The STM32 Application}
The STM32 MCU is the manager of all the processes running onboard the nano-UAV.
Even if it does not carry any heavy computation, it is responsible for off-loading it to the GAP9 via SPI communication.
The application we developed on the STM32 is structured in three Free-RTOS tasks: \textit{the mission task}, \textit{the flight task}, and \textit{the sensor task}.
Overall, these tasks require less than \qty{4}{\kilo\byte} and only 2\% of additional CPU load in total.
The sensor task communicates with the ToF sensors via I2C.
It configures each sensor before the mission starts, fetches data from the ToF matrix sensors, and passes it to the other tasks.
The flight task runs the wall-following exploration strategy introduced in Section~\ref{sec:implementation-exploration}.
However, other tasks can notify the flight process via a Free-RTOS queue to perform other maneuvers, such as stopping and spinning the drone to acquire a scan.
The loops of the flight task and the sensor task run with a frequency of \qty{15}{\hertz}.
\begin{figure} [t]
\begin{centering}
\includegraphics[width=\columnwidth]{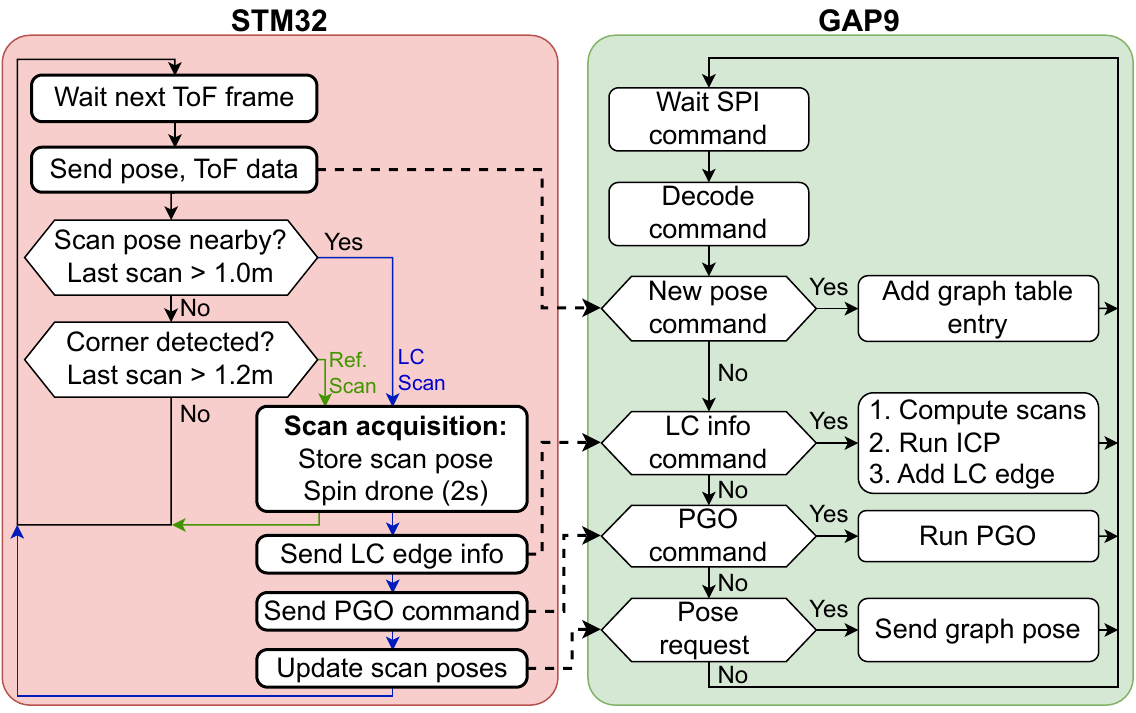}
\par\end{centering}
\centering{}
\caption{The flow diagram of our software and the interaction between the STM32 and the GAP9.}
\label{fig:app-flow}
\end{figure}

The mission task manages the scan acquisition and the communication with the GAP9 using SPI packets.
The flowchart from Figure~\ref{fig:app-flow}--left presents a detailed illustration of the mission flow. In every iteration, the task fetches the ToF data from the sensor task and the current pose from the internal state estimator and sends this information to the GAP9.
In the absence of any previous scans in the current location, if the current scan frame corresponds to a corner and the drone has traveled a minimum distance of \qty{1.2}{\meter} from the last scan, the drone captures a reference scan.
Then, the scan pose is stored in a structure called \textit{scan pose list}, which stores the locations of all acquired scans.
On the other hand, if the current location is actually revisited -- i.e., the drone is closer than \qty{0.6}{\meter} to one entry of the 
scan pose list -- an LC scan is acquired, given at least \qty{1}{\meter} from the last scan.
The STM32 informs the GAP9 about the LC and then sends a PGO command.
Lastly, the STM32 updates the scan pose list, fetching the updated scan pose values from the GAP9.
The loop of a mission task runs at \qty{7.5}{\hertz}, skipping every second ToF frame.

Note that the scan acquisition is identical for the reference and LC scans, and only the subsequent steps differ.
During a scan, the flight task is spinning the drone by \qty{45}{\degree}, while the mission task continues sending new poses to the GAP9 -- this was omitted in Figure~\ref{fig:app-flow} for the sake of readability.
Furthermore, it is important to impose a minimum distance between scans.
Conducting consecutive scans at the same location does not enhance the system's efficacy but leads to a substantial surge in memory utilization.
The distance thresholds were determined experimentally.
\begin{table} [t]
\centering
\caption{Structure of a graph table entry}
\label{tab:pose_table}
\begin{tabular}{ccc} 
\hline\hline
\textbf{Field} & \textbf{Representation} & \textbf{Size}  \\ 
\hline
Pose ID        & \textit{int32}          & \qty{4}{\byte}                   \\
Timestamp      & \textit{int32}          & \qty{4}{\byte}                   \\
Pose           & $3 \times \textit{float}$          & \qty{12}{\byte}                   \\
ToF Data       & $4 \times 8 \times \textit{int16}$          & \qty{64}{\byte}                  \\
\hline\hline
\end{tabular}
\end{table}

\subsection{The GAP9 Application}
The STM32 MCU performs several crucial functions, including managing sensor communication, controlling the drone, and determining when to acquire new scans. 
In contrast, the GAP9 processor assumes a subordinate role by handling computationally intensive tasks.
The STM32 sends SPI packets to the GAP9, and each packet consists of a command and a corresponding data field, with the interpretation of the data contingent upon the specific command type. 
The GAP9 continually awaits the arrival of a new SPI packet, upon which it proceeds to decode and execute the command.
Four possible SPI commands are defined in the system, distinguished by a command ID.
The first is the \textit{new pose} command, signaling that a new pose should be added to the graph, along with its associated ToF data.
The graph is stored in memory as a table (i.e., \textit{the graph table}), where each graph table entry contains the pose ID, the timestamp, the pose values, and the ToF data from the four sensors.
The structure of each graph table entry is presented in Table~\ref{tab:pose_table}, and the total size of an entry is \qty{84}{\byte}.
Note that one graph table entry carries all the necessary information to compute one scan frame.

The second command is the \textit{LC information} command.
Within this command, the STM32 communicates that an LC edge should be added to the graph, from an id $j$ to id $i$, communicated in the SPI packet.
The GAP9 application fetches the graph table entries from $i$ to $i+19$ and from $j$ to $j+19$ and then calculates their associated scans $\mat{S}_i$ and $\mat{S}_j$.
Having the two scans, it then computes the LC edge measurement using ICP and stores it into the \textit{LC edge list}.
The third command defined within our system is the \textit{PGO} command, which informs the GAP9 to optimize the existing poses in the graph, given the LC edge list.
Our system always uses the hierarchical graph-based SLAM for PGO, as it allows for mapping larger environments.
When PGO completes, the graph table is updated with the new pose values.
Optionally, the map is regenerated by combining the scan frames computed from every graph pose entry.
Lastly, the fourth command is the \textit{pose request}, which enables the STM32 to obtain the value of a particular pose in the graph table by communicating its ID in the SPI packet.
This is necessary because the STM32 application must update the scan pose list after every PGO.
Furthermore, it must also update the drone's state estimator with the updated value of the most recent pose. 
Figure~\ref{fig:app-flow} illustrates the behavior of each command and the interaction with the STM32.
NanoSLAM represents the whole logic running in the GAP9, which stores the graph, fetches scans, uses scan-matching to add new LC edges, and exploits hierarchical PGO to correct the poses.

\section{Performance Analysis} \label{sec:performance}
In this section, we provide a breakdown of the execution time of our algorithms.
We mainly evaluate the ICP, graph-based SLAM, and its hierarchical extension, emphasizing the benefits of the eight-core parallelization.
Lastly, we also provide a power breakdown for the individual stages of graph-based SLAM.
Power measurements are conducted upstream of the buck converter on the GAP9 co-processor deck, which receives a voltage supply of \qty{4}{\volt} and generates an output voltage of 
\qty{1.8}{\volt} intended for supplying the GAP9 SoC.
The GAP9 always operates at the maximum frequency -- i.e., \qty{400}{\mega\hertz} for both FC and CL.

\subsection{Execution Time of ICP}
Table~\ref{tab:icp-exec} shows the ICP execution time as a function of the scan size.
The first line of the table provides the total execution time when the algorithm is parallelized and computed with the aid of the CL. 
To highlight the benefit of parallelization, we also provide the execution time when ICP runs entirely in the FC -- i.e., second table line.
The speedup, defined as the ratio between the execution time on the FC and the CL, increases with the scan size.
This is due to the memory transfer overhead; for larger scan sizes, the computation time of the correspondences is significantly higher than the time necessary to transfer the scans from L2 to L1, which the CL can access. 
For a scan size larger than 640, the achieved speedup is above seven with eight cores.
Furthermore, for the scan size used for the scope of this paper (i.e., 640), the ICP executes in \qty{55}{\milli\second}.

\begin{table} [t]
\centering
\caption{Execution time in \qty{}{\milli\second} of the ICP algorithm w.r.t. the scan size. CL is the cluster with 8+1 cores, while the FC is the single-core fabric controller of the GAP9 SoC.}
\label{tab:icp-exec}
\begin{tabular}{lcccccccc} 
\hline\hline
\textbf{Scan size} & \textbf{128} & \textbf{256} & \textbf{384} & \textbf{512} & \textbf{640} & \textbf{768} & \textbf{896} & \textbf{1024}  \\ 
\hline
CL   & 3            & 10           & 21           & 36           & 55           & 79           & 107          & 138            \\
FC   & 16           & 63           & 141          & 249          & 386          & 557          & 758          & 990            \\
\textbf{Speedup}    & \textbf{5.33}           & \textbf{6.30}           & \textbf{6.71}          & \textbf{6.91}          & \textbf{7.01}          & \textbf{7.05}          & \textbf{7.08}          & \textbf{7.11}            \\
\hline\hline
\end{tabular}
\end{table}

\subsection{Execution Time of Graph-based SLAM}
\begin{table} [t]
\centering
\caption{The execution time in \qty{}{\milli\second} of the Cholesky decomposition. CL is the cluster with 8+1 cores, while the FC is the single-core fabric controller of the GAP9 SoC.}
\label{tab:cholesky-time}
\begin{tabular}{ >{\arraybackslash}m{2.0cm} >{\centering\arraybackslash}m{0.35cm} >{\centering\arraybackslash}m{0.35cm} >{\centering\arraybackslash}m{0.35cm} >{\centering\arraybackslash}m{0.35cm} >{\centering\arraybackslash}m{0.35cm} >{\centering\arraybackslash}m{0.4cm} >{\centering\arraybackslash}m{0.4cm} >{\centering\arraybackslash}m{0.4cm}} 
\hline\hline
\textbf{Nr. of poses}     & \textbf{20} & \textbf{80} & \textbf{140} & \textbf{200} & \textbf{260} & \textbf{320} & \textbf{380} & \textbf{440}  \\ 
\hline
CL (8 cores)        & 0.51         & 2.52         & 5.51          & 9.39         & 14.71          & 20.18          & 26.88          & 34.96           \\ 
FC (1 core)       & 0.86         & 8.76         & 24.88          & 49.72          & 82.74         & 124.1        & 173.9         & 232.0          \\
\textbf{Speedup}                     & \textbf{1.68}         & \textbf{3.47}       & \textbf{4.52}         & \textbf{5.29}        & \textbf{5.63}        & \textbf{6.15}        & \textbf{6.47}        & \textbf{6.64} \\  
\textbf{Speedup SLAM}  & \textbf{1.28}         & \textbf{2.24}       & \textbf{2.97}         & \textbf{3.55}        & \textbf{4.04}        & \textbf{4.43}        & \textbf{4.79}        & \textbf{5.08} \\       
\hline\hline
\end{tabular}
\end{table}
\begin{figure} [t]
\begin{centering}
\includegraphics[width=\columnwidth]{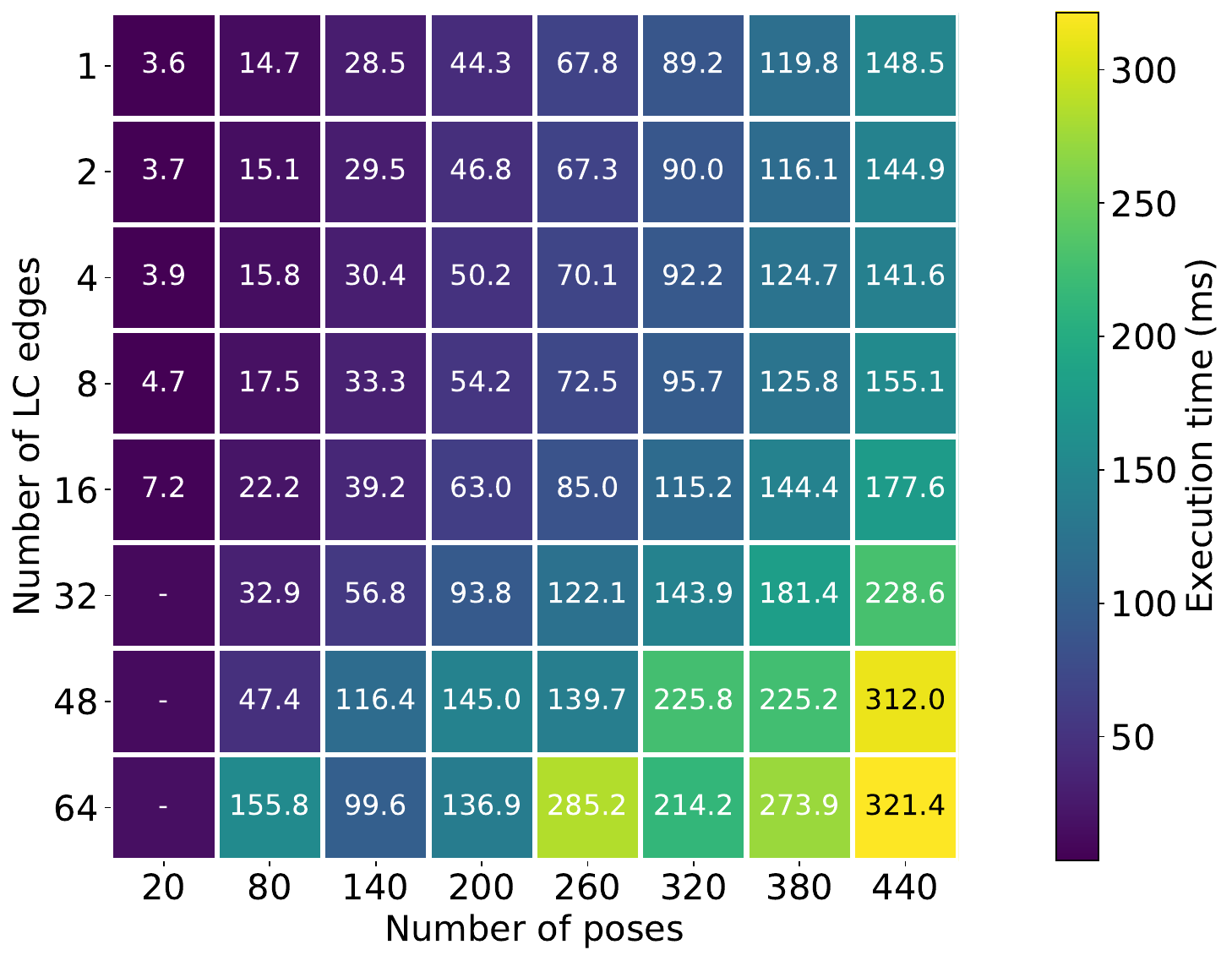}
\par\end{centering}
\centering{}
\caption{The execution time in \qty{}{\milli\second} of graph-based SLAM for multiple configurations of the number of poses and LC edges.}
\label{fig:slam-time}
\end{figure}
In the previous section, we have explained how every constitutive stage of graph-based SLAM is implemented.
In the following, we provide the execution time and complexity of every stage.
In this regard, we first analyze the Cholesky decomposition, which is the most complex and computationally intensive part of graph-based SLAM.
As explained in Section~\ref{sec:implementation-slam}, the decomposition is offloaded to the CL of GAP9 to accelerate its execution through parallelization over eight cores.
To highlight the advantages of parallelization, we also evaluate the execution time of the decomposition solely on the FC using a single core. 
Subsequently, we analyze the resulting measurements in comparison to those obtained on the CL.
Similarly to the example analyzed in Section~\ref{sec:implementation-slam}, we consider a graph with two LC edges, and we vary the number of poses in the range of 20 -- 440 with a step of 60.
Table~\ref{tab:cholesky-time} provides the results of this comparative analysis, showing the execution time as a function of the number of poses.
The \textit{Speedup} line gives the ratio between the execution time on the FC and the CL for each pose number.
The maximum speedup is achieved for 440 poses, reducing the execution time from \qty{232}{\milli\second} to \qty{34.96}{\milli\second} and resulting in a speedup of 6.64.
Overall, the table shows an increasing trend of the speedup with the number of poses.
This is because the overhead for moving the input matrix to L1 (accessible by the CL) becomes more and more negligible w.r.t. the computation time for larger graphs.
Furthermore, the \textit{Speedup SLAM} line shows how many times the whole graph-based SLAM algorithm is accelerated when the Cholesky decomposition is offloaded to the CL.
The maximum speed is 5.08, achieved with 440 poses.

\begin{table} [b]
\centering
\caption{The execution time in \qty{}{\milli\second} of graph-based SLAM.}
\label{tab:slam-breakdown}
\begin{tabular}{ >{\arraybackslash}m{2.0cm} >{\centering\arraybackslash}m{0.35cm} >{\centering\arraybackslash}m{0.35cm} >{\centering\arraybackslash}m{0.35cm} >{\centering\arraybackslash}m{0.35cm} >{\centering\arraybackslash}m{0.35cm} >{\centering\arraybackslash}m{0.4cm} >{\centering\arraybackslash}m{0.4cm} >{\centering\arraybackslash}m{0.4cm}} 
\hline\hline
\textbf{Nr. of poses}     & \textbf{20} & \textbf{80} & \textbf{140} & \textbf{200} & \textbf{260} & \textbf{320} & \textbf{380} & \textbf{440}  \\ 
\hline
H and b    & 0.3         & 1.1         & 2            & 2.9          & 3.9          & 4.6          & 5.4          & 6.3           \\
RCM              & 0.2         & 0.9         & 1.5          & 2.2          & 2.9          & 3.5          & 4.2          & 4.9           \\ 
Cholesky        & 0.5         & 2.5         & 5.5          & 9.4          & 14.7         & 20.2         & 26.9         & 35.0          \\
Fwd+Bwd & 0.2         & 0.8         & 1.3          & 1.8          & 2.3          & 2.8          & 3.3          & 3.8           \\
\textbf{Iter. time}                     & \textbf{1.3}         & \textbf{5.4}       & \textbf{10.4}         & \textbf{16.3}        & \textbf{23.7}        & \textbf{31.1}        & \textbf{39.9}        & \textbf{50} \\  
\textbf{Total (3 iter.)}                     & \textbf{3.7}         & \textbf{15.1}       & \textbf{29.5}         & \textbf{46.8}        & \textbf{67.3}        & \textbf{90}        & \textbf{116.1}        & \textbf{144.9} \\       
\hline\hline
\end{tabular}
\end{table}
Table~\ref{tab:slam-breakdown} presents the execution time analysis of the main stages of graph-based SLAM.
The stages are paired with step 3 of Listing~\ref{lst:graph-based-slam}.
The experiment involves a graph comprising two LC edges and a varying number of poses ranging from 20 to 440.
The initial four rows of the table provide detailed information about the individual execution times for each stage within a single iteration, while the subsequent row presents the total iteration time.
Notably, the Cholesky decomposition accounts for approximately 40\% to 70\% of the total iteration time.
Additionally, it is observed that all stages, except for the Cholesky decomposition, exhibit linear complexity.
Although the conventional decomposition has a complexity of $O(N^3)$, the numbers in Table~\ref{tab:slam-breakdown} demonstrate a purely quadratic relationship with the number of poses, showing a correlation of 99.9\% with a second-order polynomial fit.
This is due to our efficient implementation that exploits the sparsity properties.
The last row provides the total execution time of the three iterations.
This is approximately equal to the iteration time multiplied by three, but inter-iteration differences are possible due to different non-zero counts of the $\mat{H}$ and $\mat{L}$ matrices.

In the next experiment, we investigate the impact of varying both the number of poses and LC edges on the execution time of graph-based SLAM. 
The ranges considered for the number of poses and LC edges are 20 -- 440 and 1 -- 64, respectively, as depicted in Figure~\ref{fig:slam-time}.
The figure illustrates that increasing either parameter leads to an increase in the execution time, although the relationship is not strictly monotonic.
Interestingly, for instance, the scenario with 64 LC edges demonstrates a faster execution time with 320 poses compared to 260 poses.
This behavior can be attributed to the Cholesky decomposition's execution time, which is affected by the non-zero count of the matrix $\mat{L}$, determined by the permutation obtained through RCM.
Since RCM does not guarantee the same non-zero reduction for all matrices, some configurations could benefit from a higher non-zero reduction in $\mat{L}$ after applying the permutation to $\mat{H}$.
Given the \qty{128}{\kilo\byte} of L1 available to the CL, the graph-based SLAM algorithm can optimize at most 440 poses at a time, requiring \qty{321}{\milli\second} with 64 LC edges and \qty{148.5}{\milli\second} with one LC edge.

\begin{figure} [b]
\begin{centering}
\includegraphics[width=\columnwidth]{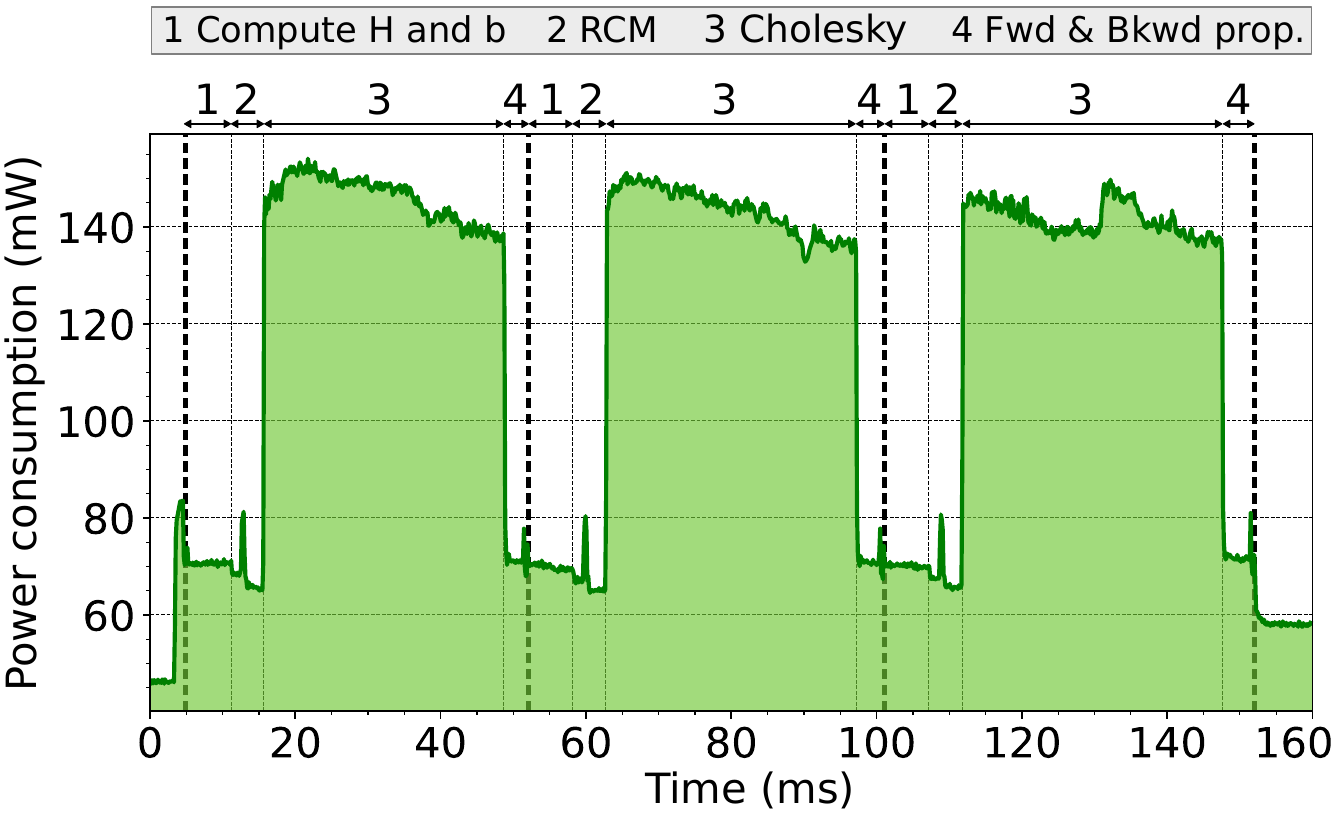}
\par\end{centering}
\centering{}
\caption{The power curve associated with the graph-based SLAM execution for 440 poses, showing the power consumption during every main stage of the algorithm.}
\label{fig:power-curve}
\end{figure}

The execution time of the hierarchical graph-based SLAM strongly depends on the structure of the sparse graph and subgraphs.
Using a large value for $d_{min}$ would result in a small, sparse graph and few large subgraphs.
On the other hand, using a small $d_{min}$ would result in a large sparse graph and many small subgraphs.
As a numerical example, for a graph of 2000 poses associated with a square loop corridor, a $d_{min}=\qty{0.3}{\meter}$ results in a total execution time of \qty{406}{\milli\second}, where the size of the sparse graph is 162 poses.
Assuming that the drone is flying with a constant velocity through the maze, the size of the subgraphs is about the same.
Under the assumption of a uniform subgraph size, the total execution time is $t_{sg} + (M-1)t_{subgraph}$, where $t_{sg}$ is the time required to optimize the sparse graph and $t_{subgraph}$ is constant.

\subsection{Power Analysis}
\begin{table} [t]
\centering
\caption{Power and energy consumption of ICP.}
\label{tab:icp-power}
\begin{tabular}{ >{\arraybackslash}m{2.2cm} >{\centering\arraybackslash}m{0.3cm} >{\centering\arraybackslash}m{0.3cm} >{\centering\arraybackslash}m{0.3cm} >{\centering\arraybackslash}m{0.3cm} >{\centering\arraybackslash}m{0.4cm} >{\centering\arraybackslash}m{0.4cm} >{\centering\arraybackslash}m{0.4cm} >{\centering\arraybackslash}m{0.4cm}} 
\hline\hline
\textbf{Scan Size}     & \textbf{128} & \textbf{256} & \textbf{384} & \textbf{512} & \textbf{640} & \textbf{768} & \textbf{896} & \textbf{1024}  \\ 
\hline
Avg. power (\qty{}{\milli\watt})    & 121.5        &149.5       & 165.0      & 169.9      & 172.5       & 175.2       & 176.3      & 177.6   \\
Energy (\qty{}{\milli\joule})       & 0.54         & 1.76        & 3.77       & 6.53      & 10.07       & 14.28       & 19.17      & 25.07     \\ 

\hline\hline
\end{tabular}
\end{table}
\begin{table} [t]
\centering
\caption{Power and energy consumption of graph SLAM.}
\label{tab:slam-power}
\begin{tabular}{ >{\arraybackslash}m{2.2cm} >{\centering\arraybackslash}m{0.3cm} >{\centering\arraybackslash}m{0.3cm} >{\centering\arraybackslash}m{0.3cm} >{\centering\arraybackslash}m{0.3cm} >{\centering\arraybackslash}m{0.4cm} >{\centering\arraybackslash}m{0.4cm} >{\centering\arraybackslash}m{0.4cm} >{\centering\arraybackslash}m{0.4cm}} 
\hline\hline
\textbf{Nr. of poses}     & \textbf{20} & \textbf{80} & \textbf{140} & \textbf{200} & \textbf{260} & \textbf{320} & \textbf{380} & \textbf{440}  \\ 
\hline
Avg. power (\qty{}{\milli\watt})    & 72.3         & 86.3       &94.7       & 88.9      & 108.7     & 111.6       & 114.9      & 119.3   \\
Energy (\qty{}{\milli\joule})       & 0.31         & 1.50        & 3.10       & 4.38      & 7.80       & 10.73       & 14.29      & 18.20     \\ 

\hline\hline
\end{tabular}
\end{table}

Table~\ref{tab:icp-power} shows the power and energy consumption of the ICP as a function of the scan size.
An observable rising pattern in the average power is observed, attributed to the correspondence calculation occupying a larger proportion of the overall execution time for larger scan sizes.
The maximum power consumption is \qty{177.6}{\milli\watt} for a scan size of 1024.
However, for the scan size that we use (i.e., 640), the average power and energy consumption are \qty{172.5}{\milli\watt} and \qty{10.07}{\milli\joule}, respectively.
In conclusion, for every LC, the system consumes about \qty{10}{\milli\joule} plus the energy consumed to optimize the graph.

In Figure~\ref{fig:power-curve}, the power trace of the GAP9 deck during the execution of graph-based SLAM is presented. The experiment involves 440 poses and 2 LC edges. The labels positioned above the plot represent the four primary stages of the algorithm: calculation of matrices $\mat{H}$ and $\vec{b}$, computation of the RCM permutation, Cholesky decomposition, and solution computed through forward and backward propagation.
An observation can be made that the power consumption is notably higher during the Cholesky decomposition phase, primarily due to the activity of the CL. The peak of the power curve reaches \qty{153}{\milli\watt}, while the average power value amounts to \qty{119.3}{\milli\watt}. When the CL is inactive, and the FC solely handles the computation, the power instant tends to remain below \qty{80}{\milli\watt}.
The total energy consumed for executing the graph-based SLAM is calculated to be \qty{18.2}{\milli\joule}. In Table~\ref{tab:slam-power}, the average power and energy are tabulated for the experiment conducted with various numbers of poses ranging from 20 to 440. The average power shows a monotonic trend, decreasing with a smaller number of poses due to reduced Cholesky decomposition execution time.

\section{In-Field Experiments} \label{sec:results}
In this section, we evaluate the algorithms introduced in Section~\ref{sec:algorithms} and recall that NanoSLAM is the framework that leverages hierarchical PGO to optimize the graph and correct the drone's trajectory while considering the LC edges provided by ICP.
We, therefore, present three main classes of results: \textit{(i)} an evaluation of the rotation and translation error achieved by the scan-matching algorithm; \textit{(ii)} an investigation on how NanoSLAM improves the trajectory estimation; \textit{(iii)} coherent maps generated out of the pose graph and the ToF measurements.
Our results are experimentally acquired and demonstrate the effectiveness of our closed-loop system that leverages NanoSLAM and carries the computation entirely onboard.
The Ground Truth (GT) used in our evaluation is provided by the Vicon Vero 2.2 motion capture system (mocap) installed in our testing arena.
To assess our system's localization and mapping capabilities, we build mazes of different complexities out of $\qty{1}{\meter} \times \qty{0.8}{\meter}$ chipboard panels.
\begin{figure} [b]
\begin{centering}
\includegraphics[width=\columnwidth]{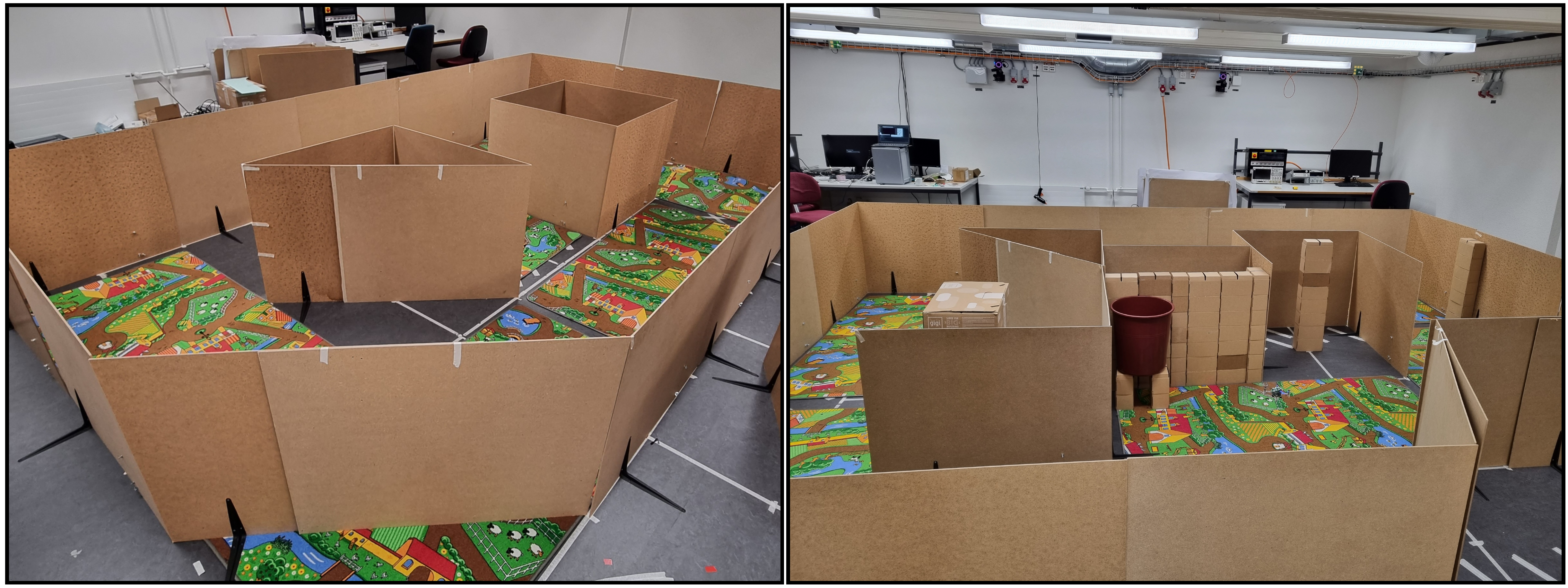}
\par\end{centering}
\centering{}
\caption{Maze 2 (left) and Maze 3 (right).}
\label{fig:mazes-real}
\end{figure}

\subsection{Scan-matching Evaluation}
\begin{figure*} [t]
\begin{centering}
\includegraphics[width=\linewidth]{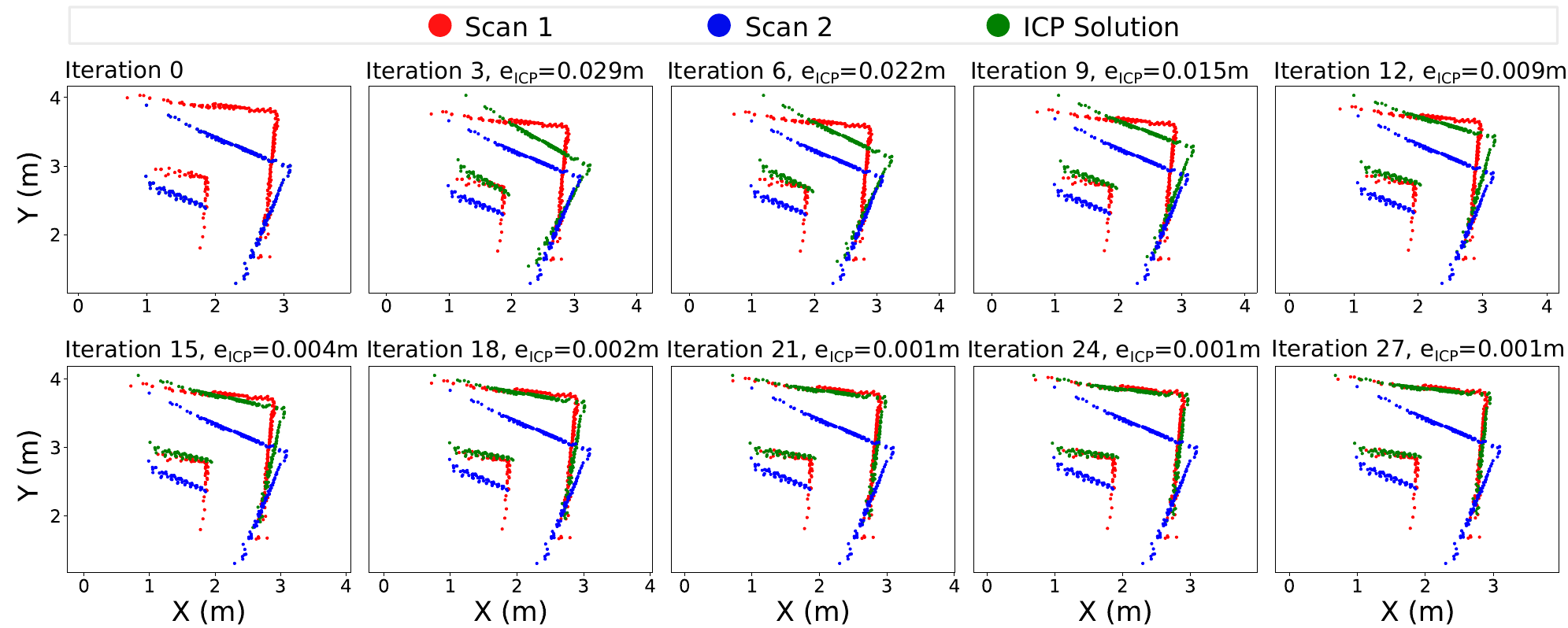}
\par\end{centering}
\centering{}
\caption{A breakdown of the ICP algorithm, indicating the solution found by the algorithm after every three iterations. Assuming a target overlapping error of $\approx\qty{0.001}{\meter}$, the algorithm converges in about 20 iterations.}
\label{fig:icp_results}
\end{figure*}
In the following, we analyze the scan-matching capabilities of the ICP algorithm. 
In this scope, we position the drone inside a \qty{90}{\degree} pipe of \qty{1}{\meter} width made out of chipboard panels. The drone is then commanded to take off and acquire a scan -- i.e., \textit{Scan 1}. Then, we manually change the position of the drone by about \qty{30}{\centi\meter} and \qty{30}{\degree} and repeat the same procedure to obtain \textit{Scan 2}.
The two scans are shown in \textit{Iteration 0} of Figure~\ref{fig:icp_results}.
The drone position is changed to simulate the odometry drift that the drone normally acquires when it revisits a location and evaluates the ICP performance in matching two non-overlapping scans.
Therefore, the resulting rotation and translation values by the ICP are compared with the ground truth -- obtained out of the ground truth of each individual pose. We obtain a translation error of $e_T = \qty{3.5}{\centi\meter}$ and a rotation error of $e_R = \qty{2.3}{\degree}$ -- the reported translation error $e_T$ represents the norm of the two components of the error -- i.e., on $x$ and $y$.
To ensure the validity of the results, we repeated the experiment multiple times, always obtaining a translation error $e_T < \qty{6}{\centi\meter}$ and a rotation error $e_R < \qty{5}{\degree}$.

Figure~\ref{fig:icp_results} shows the result of the ICP algorithm after every three iterations.
We recall that ICP aims to determine the rotation and translation that, once applied to \textit{Scan 2}, results in an optimal overlapping with \textit{Scan 1}.
This is represented by the green curve, which represents the scan obtained by applying the current ICP estimate to \textit{Scan 2}. Furthermore, $e_{ICP}$ represents the arithmetic mean of the Euclidean distances between each correspondence pair of the red and green curves. This metric evaluates the overlapping degree between \textit{Scan 1} and the ICP solution applied to \textit{Scan 2}, and it is a good indicator of when the algorithm should stop. We observe that in this case, as well as in other experiments we conducted, the ICP solution that leads to $e_{ICP}\approx \qty{0.001}{\meter}$ is found in about 20 iterations.
We note that the rotation and translation errors $(e_T, e_R)$ provide a quantitative indication of the precision of the solution found by ICP. Conversely, $e_{ICP}$ is only an intrinsic parameter indicating the convergence progress.
For example, if the input scans are affected by large amounts of noise or biases, $e_{ICP}$ could still indicate a small value, while the actual transformation found by ICP is inaccurate.

\begin{figure}[b]
    \centering
    \begin{subfigure}{0.505\columnwidth}
        \centering
        \includegraphics[width=\linewidth]{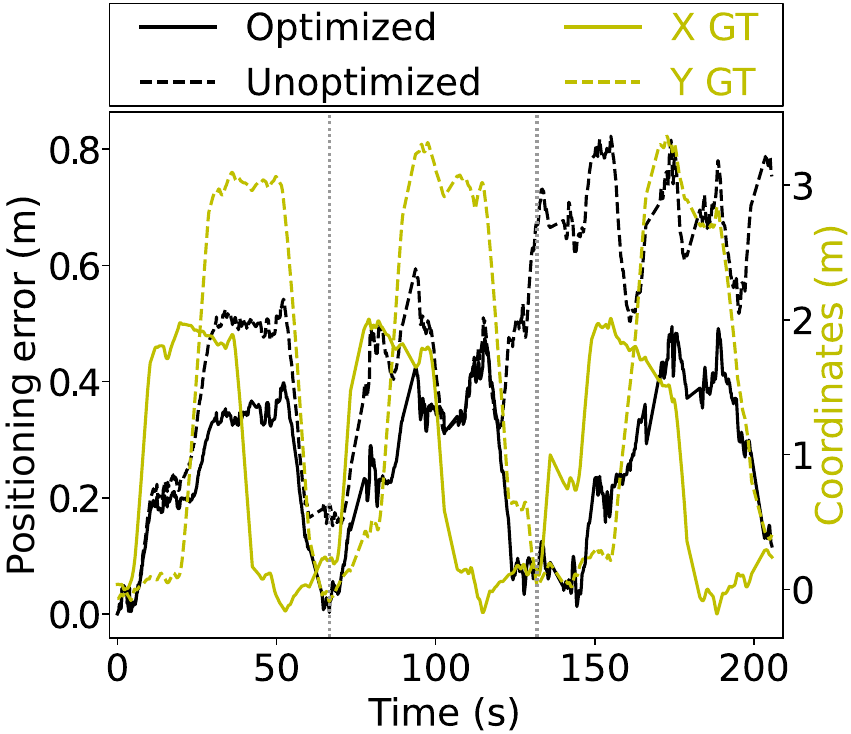}
        \caption{Tracking error over time.
        \label{fig:edo-err-a}}
    \end{subfigure}
    \begin{subfigure}{0.465\columnwidth}
        \centering
        \includegraphics[width=\linewidth]{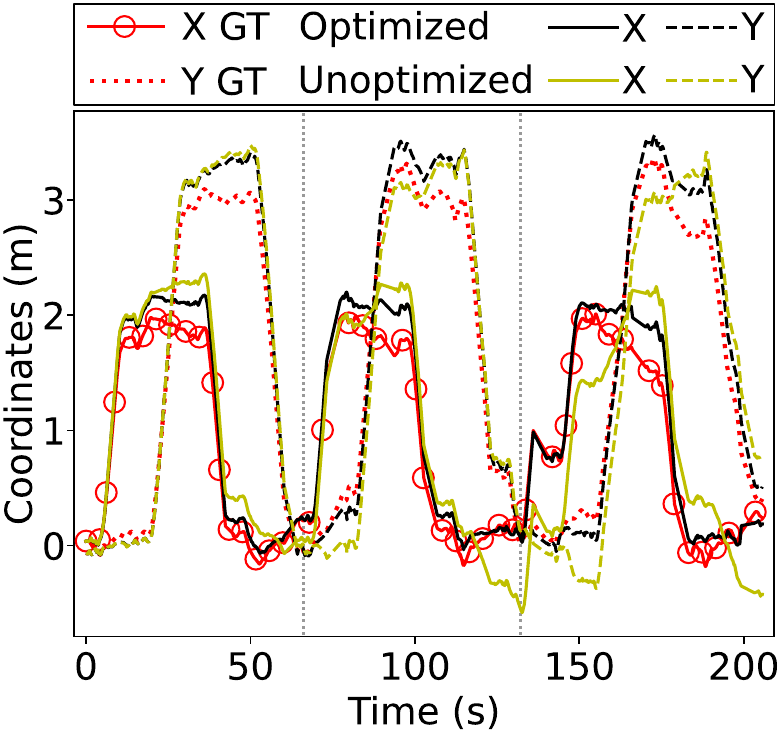}
        \caption{Trajectory over time.
        \label{fig:odo-err-b}}
    \end{subfigure}
    \caption{(a) The tracking error on both the $x$ and $y$ axis and (b) the evolution of the optimized (i.e., with NanoSLAM) and unoptimized trajectories represented against the ground truth.}
    \label{fig:edo-err}
\end{figure}
\subsection{SLAM Results}
In the following, we demonstrate our system's capabilities to correct trajectories and generate coherent maps in three different mazes of increasing complexity.
The first maze consists of a square loop corridor similar to the one used as an example in Section~\ref{sec:algorithms-slam-ex}, whereas the latter two mazes exhibit greater complexity and are illustrated in Figure~\ref{fig:mazes-real}.
\subsubsection{Maze 1}
\begin{figure} [b]
\begin{centering}
\includegraphics[width=\columnwidth]{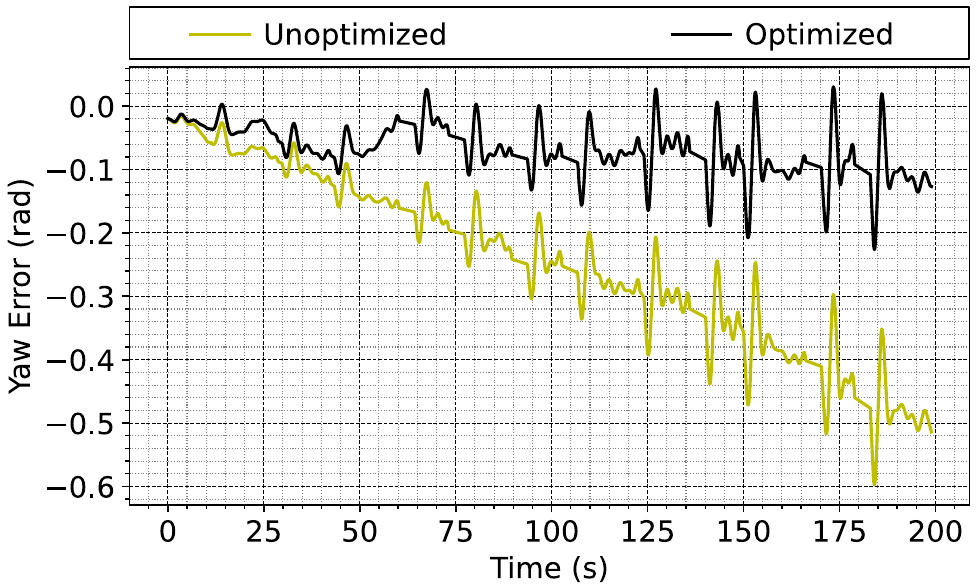}
\par\end{centering}
\centering{}
\caption{The heading error with and without NanoSLAM.}
\label{fig:yaw-error}
\end{figure}

\begin{figure*}[t]
    \centering
    \begin{subfigure}{0.32\linewidth}
    \centering
    \includegraphics[width=\linewidth]{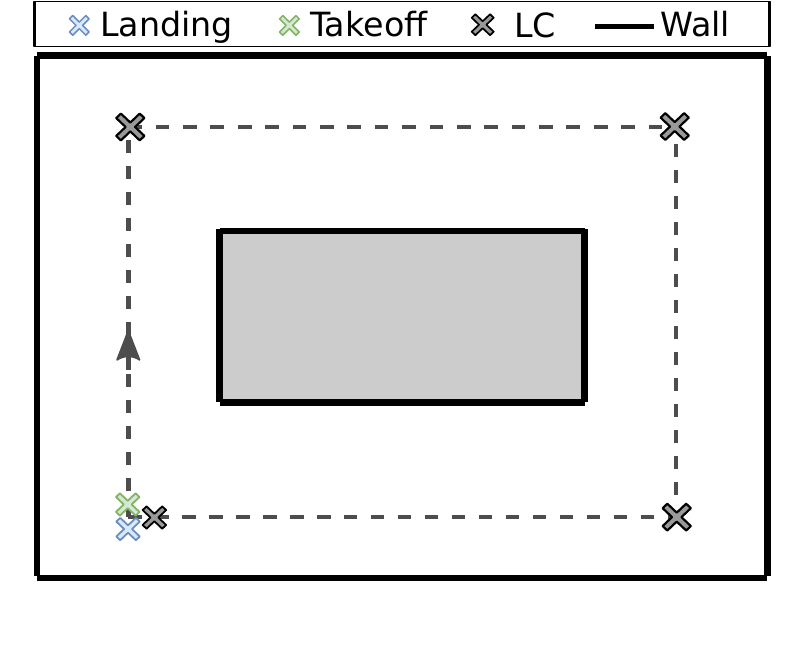}
    \captionsetup{skip=-0.1pt}
    \caption{Maze 1 layout.
    \label{fig:maze1-maze}}
    \end{subfigure}
    \begin{subfigure}{0.32\linewidth}
    \centering
    \includegraphics[width=\linewidth]{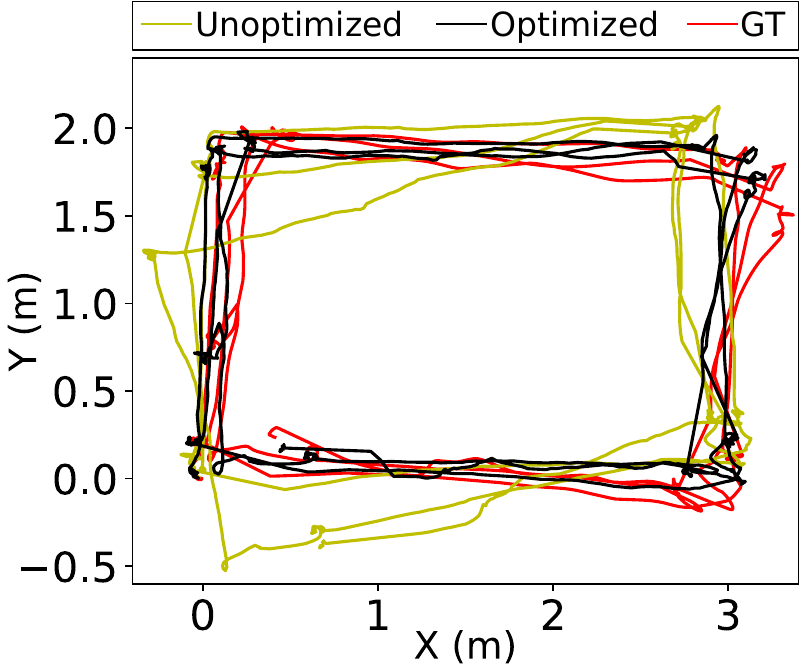}
    \captionsetup{skip=-0.1pt}
    \caption{Trajectory with / without NanoSLAM.
    \label{fig:maze1-traj0}}
    \end{subfigure}
    \begin{subfigure}{0.32\linewidth}
    \centering
    \includegraphics[width=\linewidth]{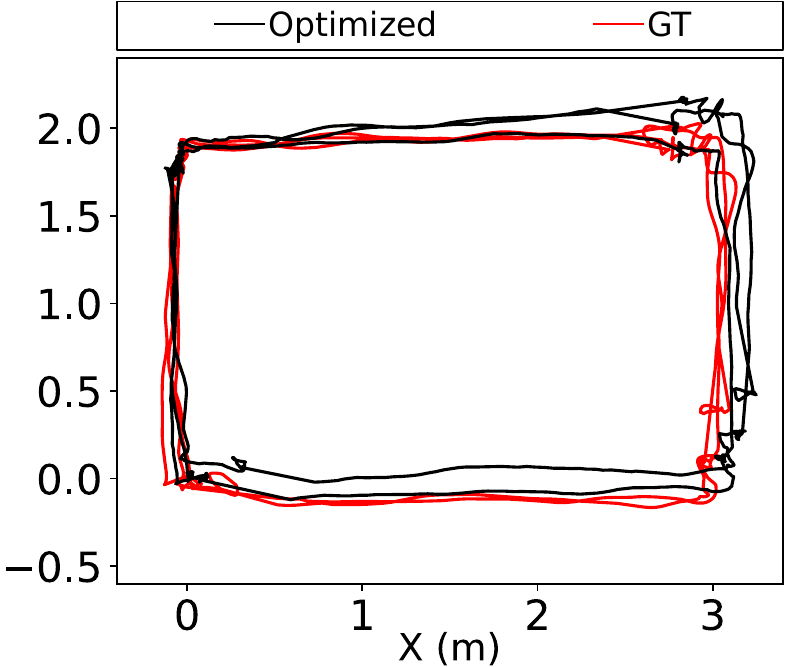}
    \captionsetup{skip=-0.1pt}
    \caption{Trajectory with 1-LC NanoSLAM.
    \label{fig:maze1-traj1}}
    \end{subfigure}
    \begin{subfigure}{0.32\linewidth}
    \vspace{3mm}
    \centering
    \includegraphics[width=\linewidth]{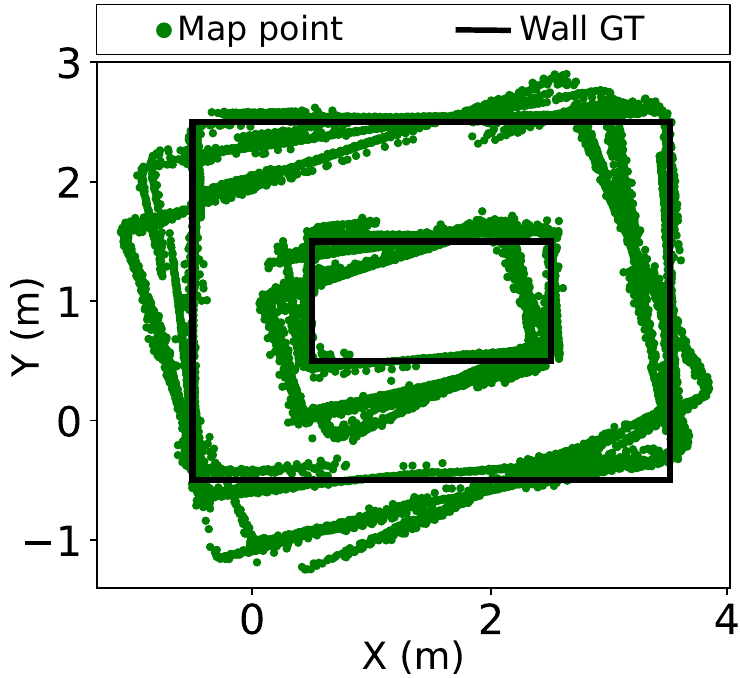}
    \captionsetup{skip=-0.1pt}
    \caption{Map without optimization.
    \label{fig:maze1-map0}}
    \end{subfigure}
    \begin{subfigure}{0.32\linewidth}
    \centering
    \includegraphics[width=\linewidth]{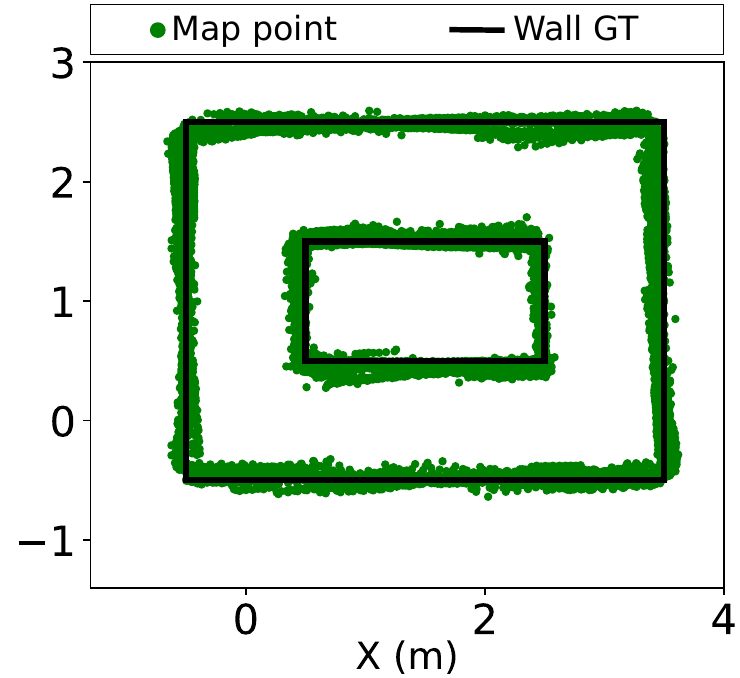}
    \captionsetup{skip=-0.1pt}
    \caption{Map with NanoSLAM.
    \label{fig:maze1-map1}}
    \end{subfigure}
    \begin{subfigure}{0.32\linewidth}
    \centering
    \includegraphics[width=\linewidth]{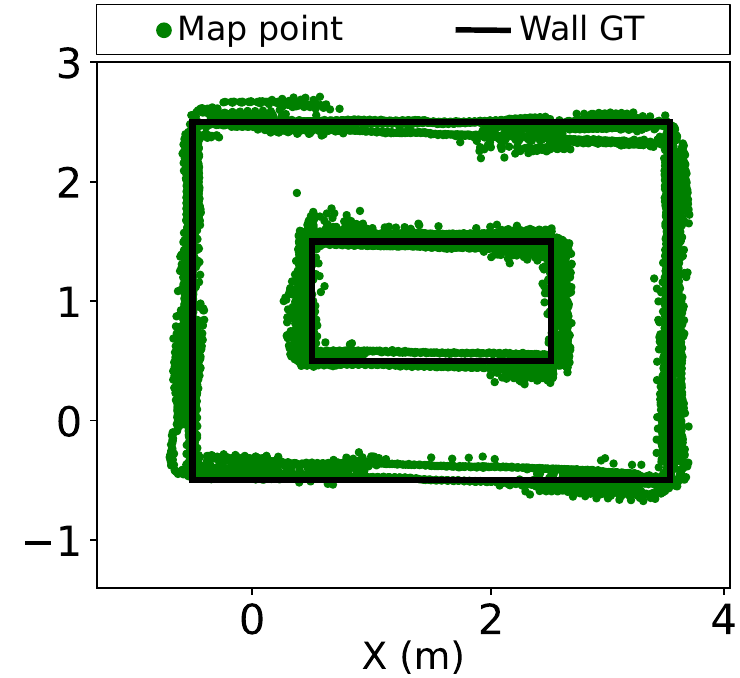}
    \captionsetup{skip=-0.1pt}
    \caption{Map with 1-LC NanoSLAM.
    \label{fig:maze1-map2}}
    \end{subfigure}
    \caption{Maze 1: (a) Illustration of the maze layout, showing the takeoff and landing locations and where an LC is performed. (b)-(c) The trajectories obtained with the two optimization approaches (in black) and the GT (in red). (d)-(f) The dense maps generated based on the unoptimized, NanoSLAM-corrected, and 1-LC NanoSLAM-corrected trajectories.}
    \label{fig:maze1}
\end{figure*}

We start with a simple circular maze (i.e., \textit{Maze 1}), shown in Figure~\ref{fig:maze1-maze}.
The drone's mission starts from the bottom left corner and flies three laps -- i.e., cycle through the maze three times, relying on the wall following strategy, introduced in Section~\ref{sec:implementation-exploration}.
During the first lap, the drone identifies every corner and acquires a reference scan in each of the four.
In the second and third laps, the drone acquires new scans in the revisited corners, creating LC constraints with the reference scans and adding new LC edges to the graph.
To allow for a consistent comparison between the unoptimized poses (i.e., no drift correction) and the optimized poses (i.e., with NanoSLAM), we only perform graph optimization at the end of the mission.
In this way, we show the benefits of using NanoSLAM on data from the same mission.

We define as \textit{trajectory} the set of all poses acquired during a mission and as positioning error the pose-wise Euclidean distance between the poses and their GT.
Figure~\ref{fig:edo-err-a} shows the positioning error (in black) of the optimized and unoptimized trajectories, calculated by subtracting the GT from the optimized and unoptimized poses, respectively.
The vertical grey lines indicate when the drone passes again through the starting point. 
Furthermore, in the same figure, we represent the $x$ and $y$ components of the trajectory over time (in yellow) to highlight a pattern between the positioning error and the trajectory.
While the error of the unoptimized trajectory drifts unbounded, the optimized trajectory shows a rather repetitive pattern, with the error being zero every time the drone crosses the starting point.
This is because the reference scan (of \textit{pose 0}) acquired right after take-off is error-free and, therefore, any LC constraint between a \textit{pose k} and \textit{pose 0} will correct the positioning error of \textit{pose k} almost completely -- the correction is only bounded by the accuracy of the ICP.

Despite using NanoSLAM, the positioning error takes considerable values of about \qty{0.4}{\meter} throughout one lap.
A fundamental assumption when using SLAM is that the odometry errors are increasing slowly~\cite{siegwart2011introduction}, and therefore the poses associated with the reference scans acquired at the beginning of the mission are accurate.
However, this assumption does not hold in our case, as the poses of the second and third reference scans already have errors of up to \qty{0.4}{\meter}, and these errors will represent a lower bound for a future LC correction.
Throughout a lap, the positioning error increases while the drone moves toward the top left corner of the maze and decreases during the last half of every lap.
In other words, forward movement on the $x$ and $y$ axis increases the positioning error, while a backward movement along one axis decreases the positioning error.
This translates into a direction-dependent odometry bias that is positive for forward movements and negative for backward movements.
Since the takeoff position is always $(0,0)$, the effect of the bias results in a scaled trajectory w.r.t. the GT.

The scaling effect can also be seen in Figure~\ref{fig:odo-err-b}, which shows information from the same mission but represents the $x$ and $y$ components of the optimized and unoptimized trajectories as well as the GT.
The optimized trajectory (black) is very similar to the ground truth (red), but scaled by a factor which we determined to be $\approx 11\%$.
On the other hand, we stress that the shape of the unoptimized trajectory (yellow) is often very different compared to the ground truth, which proves the effectiveness of NanoSLAM in correcting the trajectory and making it match the ground truth.
So far, we have proved that the errors uncorrectable by NanoSLAM (i.e., the direction-dependent drift) are deterministic, and simply scaling the poses obtained from the drone's state estimator by $0.9$ mitigates their effect.
To demonstrate that this scaling factor generalizes to any environment, we apply the same correction for all mazes considered for our experiments and presented later in this section.
The most likely cause of the direction-dependent errors is the down-pointing optical flow camera onboard the drone, which estimates the drone's velocity and enables the state estimator to determine the position by integrating the velocity.
However, since the drone tilts when moving in a particular direction, this also rotates the camera frame, which is no longer parallel to the ground and leads to errors.

Next, we also analyze the capability of NanoSLAM to correct the heading estimation error (i.e., yaw).
Figure~\ref{fig:yaw-error} shows the yaw estimation error for the optimized and unoptimized poses over time.
The error curves were again computed with the aid of the GT.
One can notice that the heading estimation tends to drift unbounded for the unoptimized trajectory.
Furthermore, the spikes in the error curves are associated with the scan acquisition when the drone rotates by \qty{45}{\degree} and subsequently returns to its initial heading. 
While these error spikes are also visible in the yaw corrected with NanoSLAM, the estimate's mean is stable, only exhibiting a steady state error mainly bounded by the precision of ICP. 

In the presented experiments within this section, we conducted optimization according to the approach detailed in Section~\ref{sec:algorithms}, wherein an LC edge is incorporated into the graph upon acquisition of an LC scan. 
With the NanoSLAM methodology that we introduced, the LC edges integrated into the graph are not subject to removal.
Consequently, the LC edges count can only increase throughout the mission, and as shown in the experiments conducted in Section~\ref{sec:performance}, this can even triple the execution time of PGO.
For this purpose, we also analyze the \textit{1-LC NanoSLAM}, which always discards the LC edge after optimization.
Therefore, within the 1-LC NanoSLAM, the graph will always have one LC edge.
Although this lightweight approach discards the prior constraints and consequently cannot guarantee future alignment of previously matched regions, we consider it a compelling exploration due to its superior scalability when many LCs are performed.
In the following, we analyze the trajectory correction and mapping performance of both optimization techniques: NanoSLAM and 1-LC NanoSLAM.
\begin{equation} \label{eq:rmse_poses}
    RMSE_{pos} = \sqrt{\frac{\sum_{i=0}^{N-1}{\lVert \vec{x}_i - \vec{x}_i^{GT} \rVert^2}}{N}}
\end{equation}

%
\begin{figure*}[ht!]
    \centering
    \begin{subfigure}{0.32\linewidth}
    \centering
    \includegraphics[width=\linewidth]{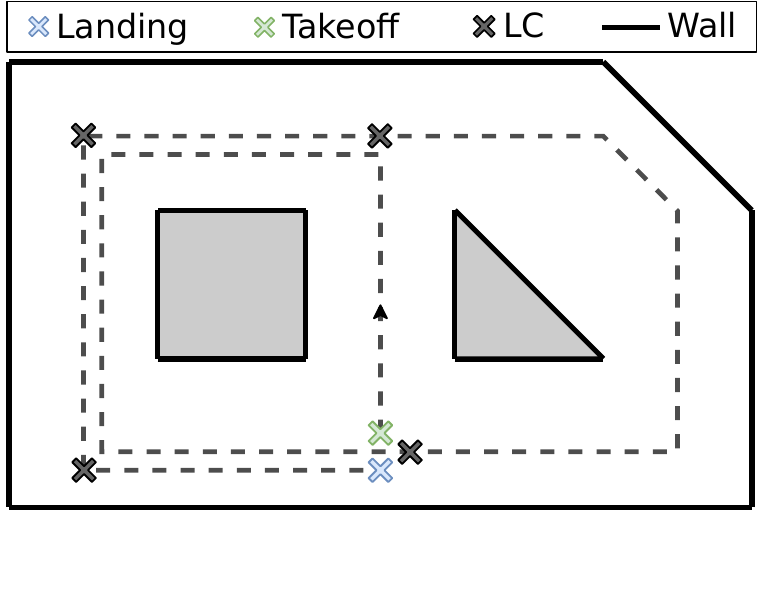}
    \captionsetup{skip=-0.1pt}
    \caption{Maze 2 layout.
    \label{fig:maze2-maze}}
    \end{subfigure}
    \begin{subfigure}{0.32\linewidth}
    \centering
    \includegraphics[width=\linewidth]{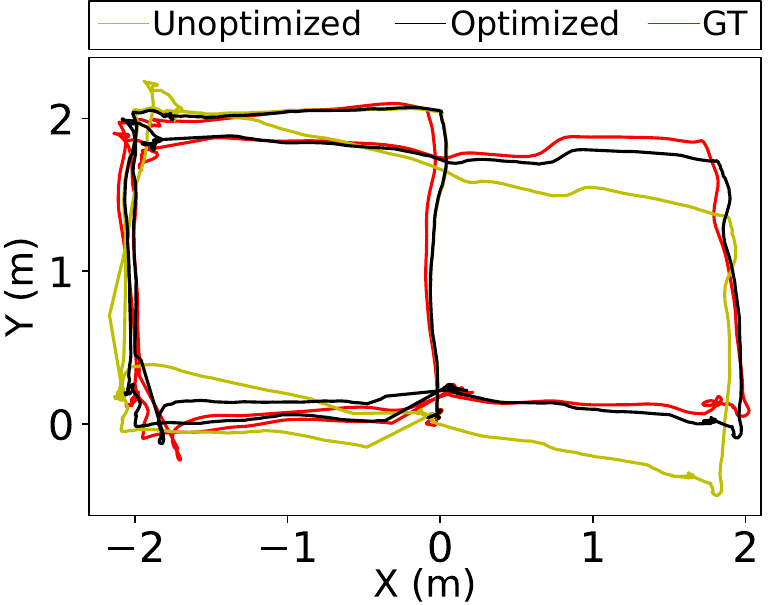}
    \captionsetup{skip=-0.1pt}
    \caption{Trajectory with / without NanoSLAM.
    \label{fig:maze2-traj0}}
    \end{subfigure}
    \begin{subfigure}{0.32\linewidth}
    \centering
    \includegraphics[width=\linewidth]{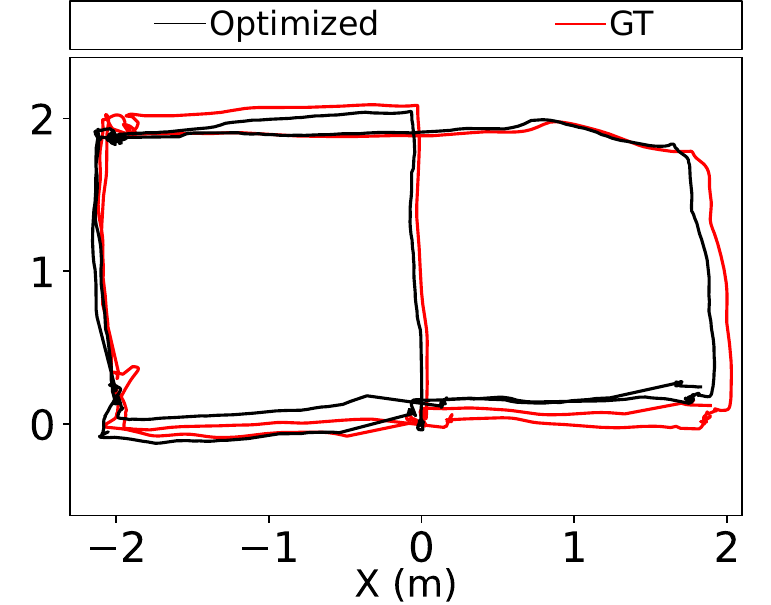}
    \captionsetup{skip=-0.1pt}
    \caption{Trajectory with 1-LC NanoSLAM.
    \label{fig:maze2-traj1}}
    \end{subfigure}
    \begin{subfigure}{0.32\linewidth}
    \vspace{3mm}
    \centering
    \includegraphics[width=\linewidth]{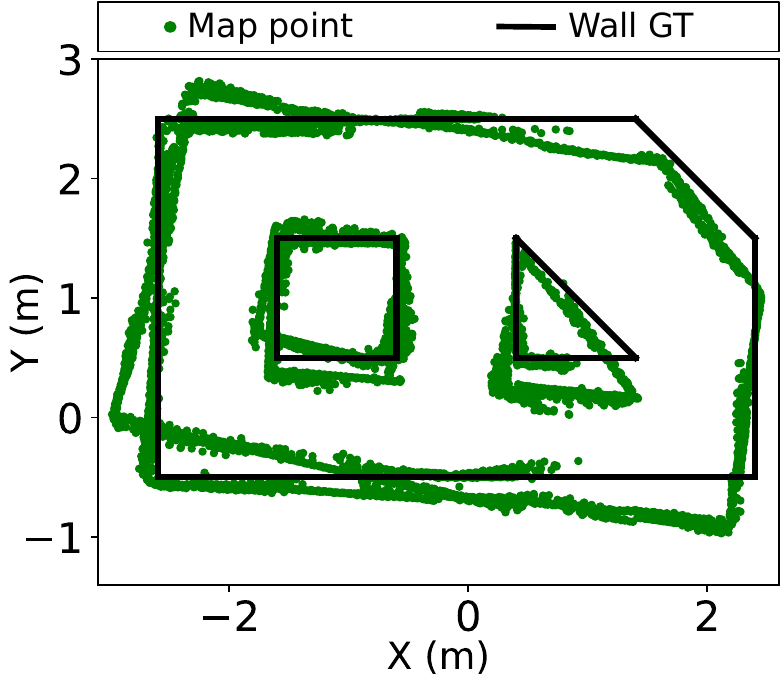}
    \captionsetup{skip=-0.1pt}
    \caption{Map without optimization.
    \label{fig:maze2-map0}}
    \end{subfigure}
    \begin{subfigure}{0.32\linewidth}
    \centering
    \includegraphics[width=\linewidth]{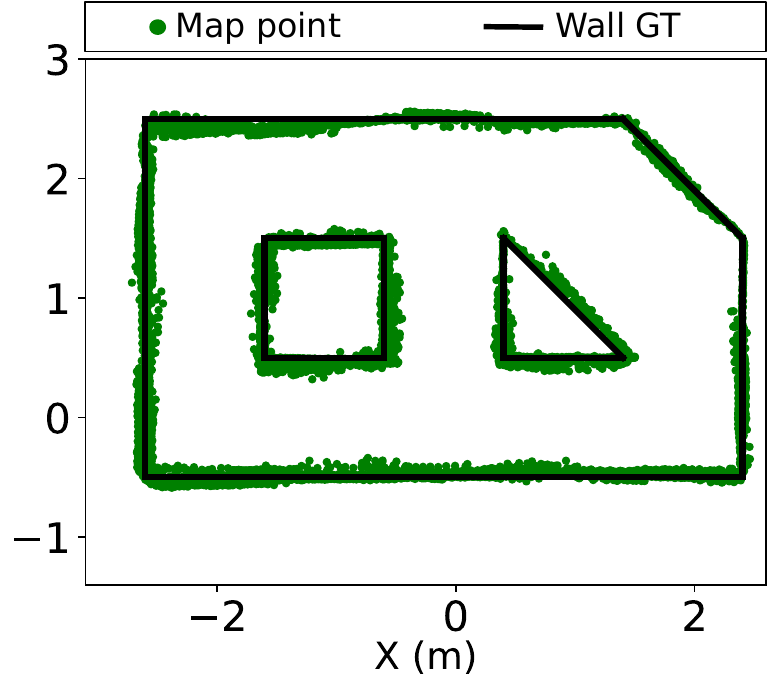}
    \captionsetup{skip=-0.1pt}
    \caption{Map with NanoSLAM.
    \label{fig:maze2-map1}}
    \end{subfigure}
    \begin{subfigure}{0.32\linewidth}
    \centering
    \includegraphics[width=\linewidth]{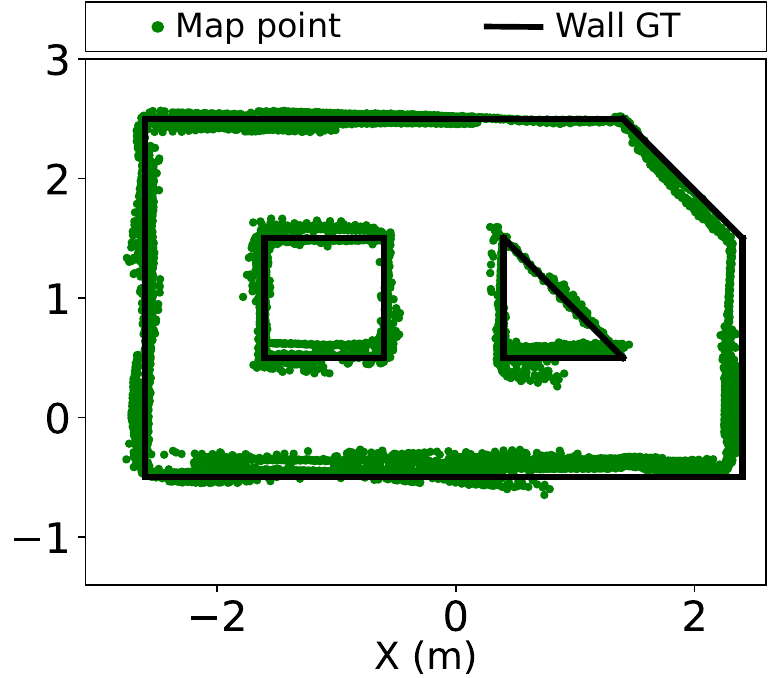}
    \captionsetup{skip=-0.1pt}
    \caption{Map with 1-LC NanoSLAM.
    \label{fig:maze2-map2}}
    \end{subfigure}
    \caption{Maze 2: (a) The maze layout. (b)-(c) The trajectories obtained with the two optimization approaches (in black) and the GT (in red). (d)-(f) The dense maps generated based on the unoptimized, NanoSLAM-based, and 1-LC NanoSLAM-based trajectories.}
    \label{fig:maze2}
\end{figure*}

Figure~\ref{fig:maze1-traj0} shows the unoptimized trajectory, the trajectory optimized with NanoSLAM and how they compare to the GT.
Although the unoptimized trajectory exhibits a noticeable deviation from the GT, we observe a substantial alignment between the optimized trajectory and the GT, providing compelling evidence of the efficacy of NanoSLAM.
While the 1-LC NanoSLAM trajectory shown in Figure~\ref{fig:maze1-traj1} also leads to satisfactory results, we notice a trajectory misalignment within the laps due to the LC constraint relaxation. 
We introduce a quantitative metric for evaluating how close each trajectory is to the ground truth, which we call positioning RMSE.
Given an arbitrary trajectory represented by the poses $\vec{x}_0 \ldots \vec{x}_{N-1}$, and the corresponding set of ground truth poses $\vec{x}_0^{GT} \ldots \vec{x}_{N-1}^{GT}$, the positioning RMSE is calculated as in Equation~\ref{eq:rmse_poses}.
The formula uses a reduced pose representation, considering only each pose's $x$ and $y$ components.
Applying Equation~\ref{eq:rmse_poses} for the trajectories in Figure~\ref{fig:maze1-traj0}, we obtain a positioning RMSE of \qty{0.46}{\meter} for the unoptimized trajectory and \qty{0.146}{\meter} for the trajectory optimized with NanoSLAM,
showing a reduction in the positioning RMSE of about three times.
Despite the inter-lap trajectory misalignment in Figure~\ref{fig:maze1-traj1}, it leads to a positioning RMSE of \qty{0.18}{\meter}, which is 23\% higher than the trajectory optimized with NanoSLAM.

%
\begin{figure*}[t]
    \centering
    \begin{subfigure}{0.32\linewidth}
    \centering
    \includegraphics[width=\linewidth]{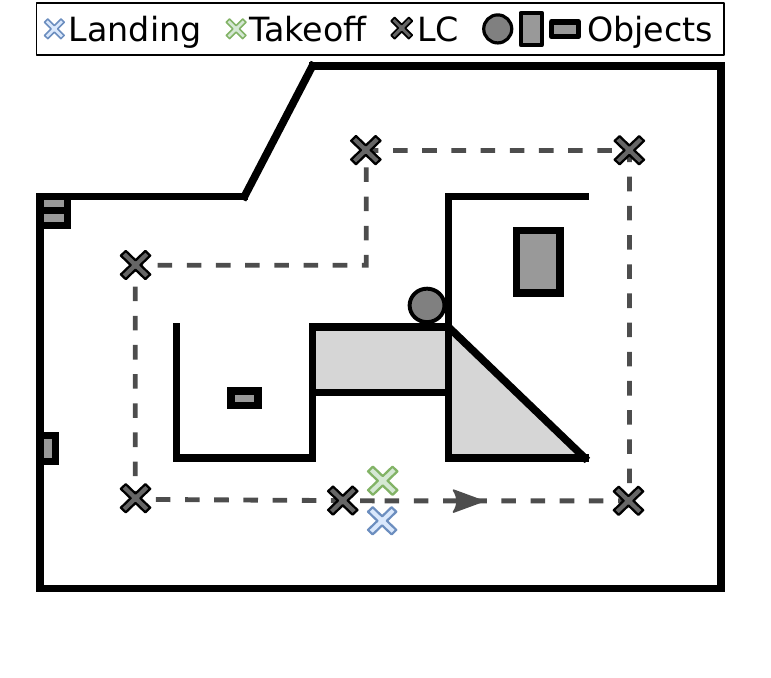}
    \captionsetup{skip=-0.1pt}
    \caption{Maze 3 layout.
    \label{fig:maze3-maze}}
    \end{subfigure}
    \begin{subfigure}{0.32\linewidth}
    \centering
    \includegraphics[width=\linewidth]{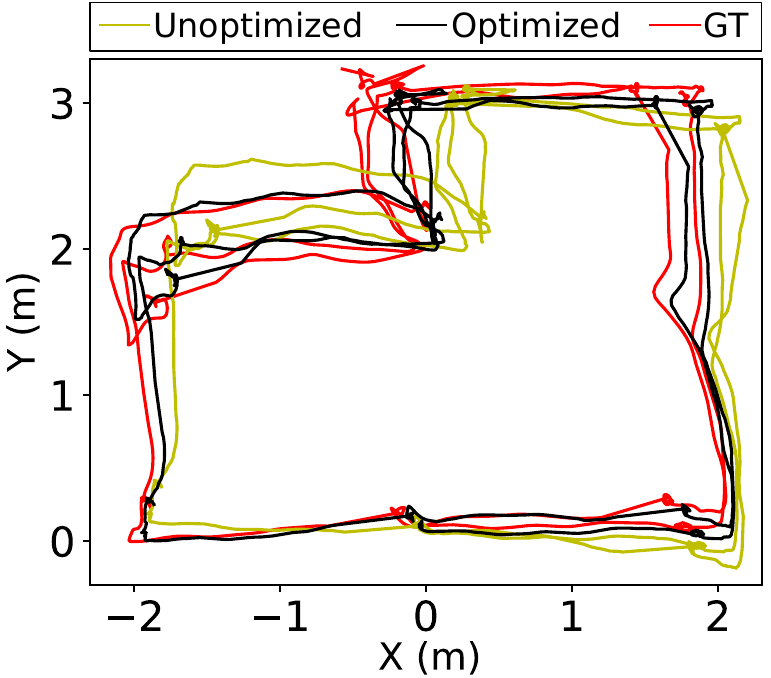}
    \captionsetup{skip=-0.1pt}
    \caption{Trajectory with / without NanoSLAM.
    \label{fig:maze3-traj0}}
    \end{subfigure}
    \begin{subfigure}{0.32\linewidth}
    \centering
    \includegraphics[width=\linewidth]{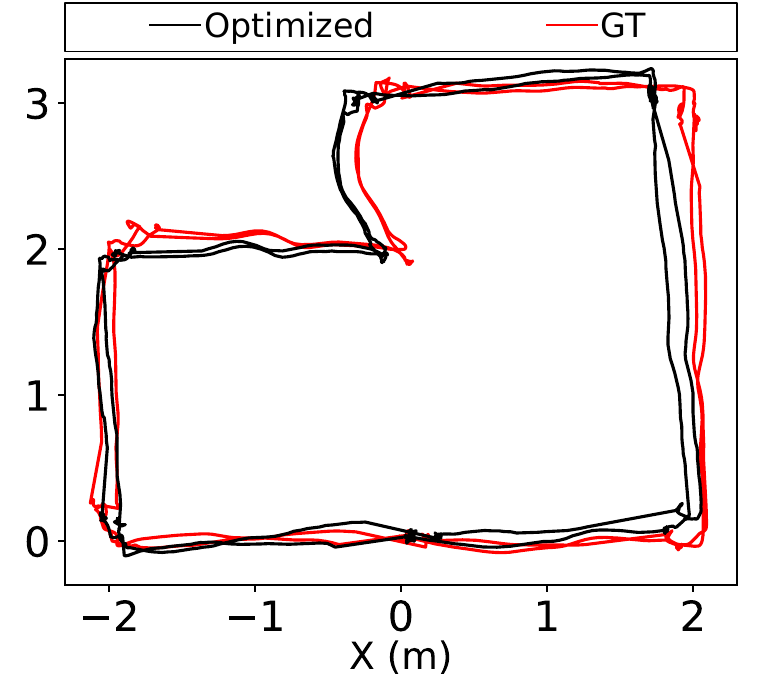}
    \captionsetup{skip=-0.1pt}
    \caption{Trajectory with 1-LC NanoSLAM.
    \label{fig:maze3-traj1}}
    \end{subfigure}
    \begin{subfigure}{0.32\linewidth}
    \vspace{3mm}
    \centering
    \includegraphics[width=\linewidth]{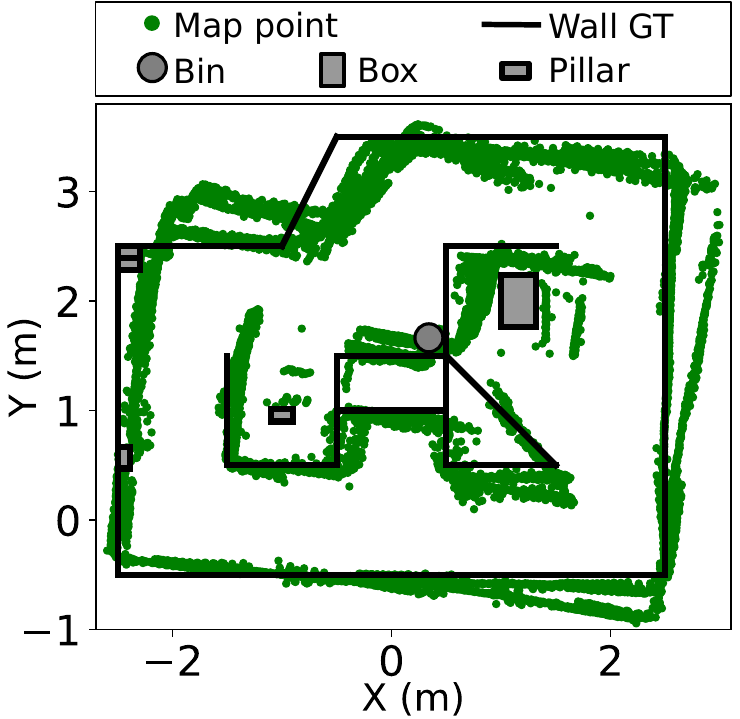}
    \captionsetup{skip=-0.1pt}
    \caption{Map without optimization.  
    \label{fig:maze3-map0}}
    \end{subfigure}
    \begin{subfigure}{0.32\linewidth}
    \centering
    \includegraphics[width=\linewidth]{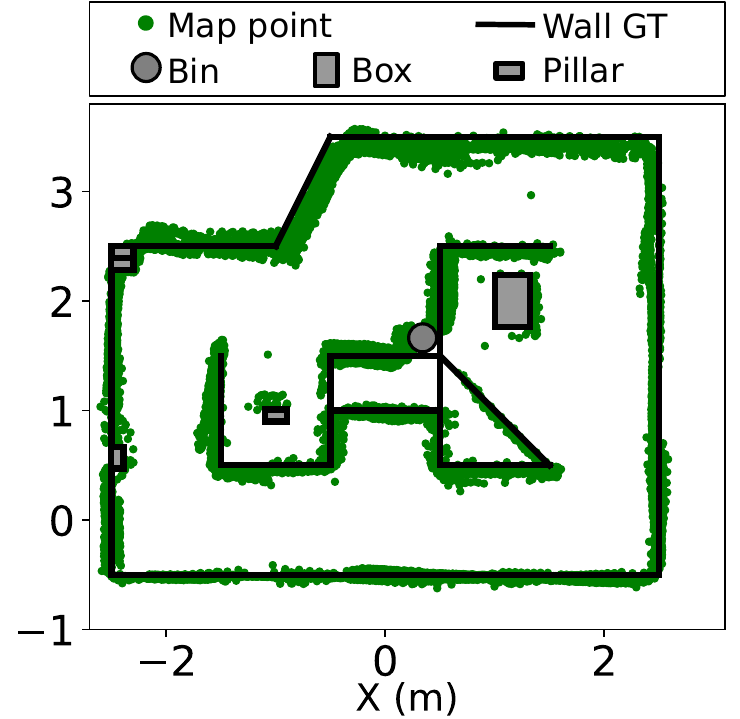}
    \captionsetup{skip=-0.1pt}
    \caption{Map with NanoSLAM.
    \label{fig:maze3-map1}}
    \end{subfigure}
    \begin{subfigure}{0.32\linewidth}
    \centering
    \includegraphics[width=\linewidth]{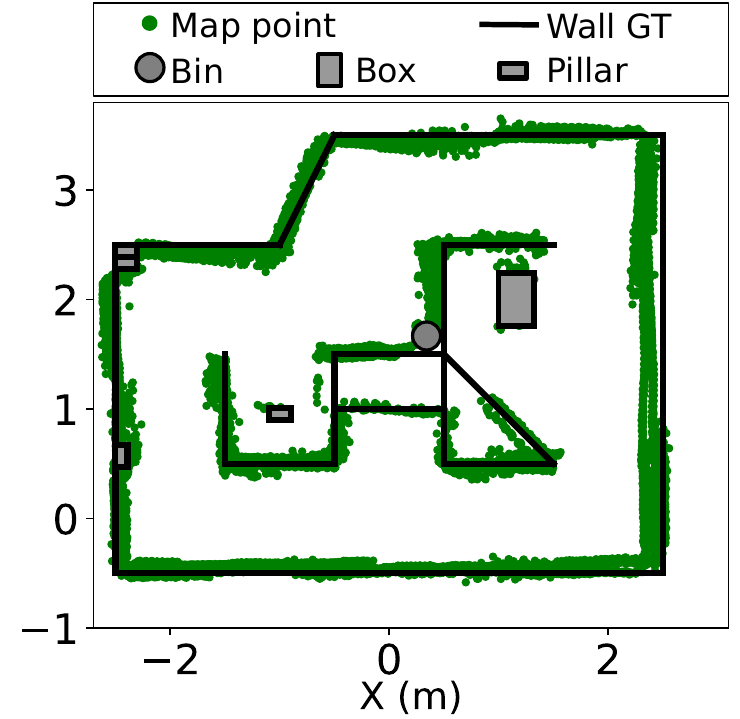}
    \captionsetup{skip=-0.1pt}
    \caption{Map with 1-LC NanoSLAM.
    \label{fig:maze3-map2}}
    \end{subfigure}
    \caption{Maze 3: (a) The maze layout that contains various objects. (b)-(c) The trajectories obtained with the two optimization approaches (in black) and the GT (in red). (d)-(f) The dense maps generated based on the unoptimized, NanoSLAM-corrected, and 1-LC NanoSLAM-corrected trajectories.}
    \label{fig:maze3}
\end{figure*}

After proving the capability of NanoSLAM to correct trajectories, we further to explore the mapping performance.
Applying Equation~\ref{eq:scan-proj} for each trajectories from Figures~\ref{fig:maze1-traj0} -- \ref{fig:maze1-traj1}, we obtain the maps shown in Figures~Figures~\ref{fig:maze1-map0} -- \ref{fig:maze1-map2}.
The maps in Figure~\ref{fig:maze1-map0} and Figure~\ref{fig:maze1-map1} are associated with the unoptimized and optimized trajectories in Figure~\ref{fig:maze1-traj0}, while the map in Figure~\ref{fig:maze1-map2} corresponds to the trajectory optimized with 1-LC NanoSLAM.
Visibly, Figure~\ref{fig:maze1-map0} shows the poorest accuracy, as no graph optimization is used, and therefore the position and heading drift heavily impact the alignment of the three laps.
The map corrected with NanoSLAM is the most accurate w.r.t. the GT.
Similarly, the map computed from the 1-LC NanoSLAM-optimized trajectory is definitely usable, but due to imposing one constraint at a time in the optimization process, not all corners corresponding to the three laps are perfectly aligned.
We also propose a quantitative metric to analyze the mapping accuracy, which we call \textit{the mapping RMSE}.
This metric first calculates the length of the projection to the closest straight line for every point in a given dense map -- the extensions of the maze lines are also considered.
The mapping RMSE represents the RMSE of all projection lengths, as shown in Equation~\ref{eq:rmse_map}.
This metric penalizes how far each map point is from a wall, and in a noise-free case, when all map points are on a maze line, the mapping RMSE is zero.
Applying Equation~\ref{eq:rmse_map} on the three maps from Figures~\ref{fig:maze1-map0} -- \ref{fig:maze1-map2} leads to \qty{21.5}{\centi\meter}, \qty{5.8}{\centi\meter}, and \qty{7.3}{\centi\meter} for the cases with no optimization, NanoSLAM and 1-LC NanoSLAM, respectively.
This proves that NanoSLAM reduces the mapping RMSE by about 3.7 times.
\begin{equation} \label{eq:rmse_map}
    RMSE_{map} = \sqrt{\frac{\sum_{i=0}^{N-1}{(\min_{\forall w \in W} dist(w, \vec{p}_i))^2}}{N}}
\end{equation}

\subsubsection{Maze 2}
\begin{table} [b]
\centering
\begin{tabular}{lccc} 
\hline\hline
\textbf{Metric}           & \textbf{No SLAM}      & \textbf{NanoSLAM}    & \textbf{1-LC NanoSLAM}    \\ \hline
Positioning RMSE & \qty{32.6}{\cm} & \qty{10.7}{\cm} & \qty{11.9}{\cm} \\
Mapping RMSE     & \qty{16.0}{\cm} & \qty{4.5}{\cm} & \qty{7.5}{\cm} \\ \hline\hline
\end{tabular}
\caption{Positioning and mapping RMSE for Maze 2 \label{tab:maze2}}
\end{table}

In the following, we present the trajectories and maps obtained by employing identical optimization methodologies on a marginally more intricate maze, as illustrated in Figure~\ref{fig:mazes-real}-(left).
A top-view layout of the maze depicted in Figure~\ref{fig:maze2-maze} reveals the presence of non-straight-angle walls, distinguishing it from Maze 1.
The trajectory followed by the drone is represented in Figure~\ref{fig:maze2-maze}, where the green cross marks the take-off point, and the arrow indicates the flying direction.
Since only the left half of the maze is revisited, the LC is only performed four times, as indicated by the grey crosses.
Table~\ref{tab:maze2} presents the positioning and mapping RMSE for the experiments in Maze 2.
We report a reduction of 67\% and 64\% in the positioning RMSE when applying NanoSLAM and 1-LC NanoSLAM, respectively, compared to the scenario without any optimization.
Looking at the maps from Figures~\ref{fig:maze2-map0}-Figures~\ref{fig:maze2-map2}, we notice a significant heading drift that rotates the map w.r.t. the GT when no optimization is performed.

Mapping with the 1-LC NanoSLAM approach leads to a mapping RMSE of \qty{7.5}{\centi\meter}, about 53\% smaller than the case without optimization.
Employing NanoSLAM reduces the mapping RMSE even further, to \qty{4.5}{\centi\meter}, representing a reduction of 72\% compared to the case without optimization.
While the positioning RMSE is somewhat similar for the NanoSLAM and 1-LC NanoSLAM approaches, the relative difference in the mapping RMSE is more significant -- about 40\%.
This observation is also evident in Figure~\ref{fig:maze2-map2}, where a comparison with the map depicted in Figure~\ref{fig:maze2-map1} reveals the presence of certain artifacts.
For instance, the bottom side of the triangle-shaped wall is distorted, or the bottom maze wall appears thicker.
This is again the effect of dropping previous LC edges and therefore failing to satisfy the constraints associated with all corners.

\subsubsection{Maze 3}
The last maze we propose is the most complex among the three because it also contains obstacles such as pillars, boxes, or a bin, and therefore it better generalizes a real-world indoor environment.
Figure~\ref{fig:mazes-real}-(right) shows an image of the maze, while the top-view layout is shown in Figure~\ref{fig:maze3-maze}.
The drone starts from the middle position marked with the green cross and performs two maze loops flying in a counterclockwise direction. 
Similar to the previous cases, throughout the first lap, the drone only acquires reference scans, while in the second lap, it closes the loop in every corner, as indicated by the grey crosses in Figure~\ref{fig:maze3-maze}.
Table~\ref{tab:maze3} shows the positioning and mapping RMSE obtained with Maze 3.
The optimized trajectories depicted in Figures~\ref{fig:maze3-traj0} -- \ref{fig:maze3-traj1} exhibit a positioning RMSE reduced by 65\% and 60\%, respectively, compared to the unoptimized trajectory.
Figures~\ref{fig:maze3-map0} -- \ref{fig:maze3-map2} show the three maps, and unlike the experiments in the first two mazes, the maps generated with NanoSLAM and 1-LC NanoSLAM are very similar.
One can still notice a misalignment on the right side of the map in Figure~\ref{fig:maze3-map2}, but it is not significant.
This is also visible in the mapping RMSE, where the error of the NanoSLAM-based map is only 14\% smaller than in the map obtained with the 1-LC NanoSLAM approach.
Both approaches bring a significant improvement w.r.t. the map in Figure~\ref{fig:maze3-map0} (i.e., without optimization), reducing the mapping RMSE by 63\% with NanoSLAM and 57\% with 1-LC NanoSLAM.
\begin{table} [b]
\centering
\begin{tabular}{lccc} 
\hline\hline
\textbf{Metric}           & \textbf{No SLAM}      & \textbf{NanoSLAM}    & \textbf{1-LC NanoSLAM}    \\ \hline
Positioning RMSE & \qty{44.1}{\cm} & \qty{15.4}{\cm} & \qty{17.3}{\cm} \\
Mapping RMSE     & \qty{20.3}{\cm} & \qty{7.5}{\cm} & \qty{8.7}{\cm} \\ \hline\hline
\end{tabular}
\caption{Positioning and mapping RMSE for Maze 3 \label{tab:maze3}}
\end{table}

In the NanoSLAM experiment, the optimization problem results in a graph with 1355 poses and five LC edges.
The subgraph has a size of 109 poses, and the total hierarchical optimization requires \qty{247}{\milli\second}.
For the experiment employing 1-LC NanoSLAM, the graph has 1293 poses, and the final optimization requires \qty{223}{\milli\second}.
The graphs associated with Maze 3 are larger than those of Maze 1 ($\approx$ 1200 poses) and Maze 2 ($\approx$ 850 poses).
For the experiment in Maze 3, running graph optimization results in an average power consumption of \qty{69.1}{\milli\watt} and a total energy consumption of \qty{17.06}{\milli\joule} -- note how optimizing a 1355 pose graph with the hierarchical graph-based SLAM results in the same energy consumption as optimizing a 440 pose graph with the direct graph-based SLAM.
Overall, for every LC, NanoSLAM implies an average power consumption of \qty{87.9}{\milli\watt}
and an energy consumption of \qty{27.13}{\milli\joule} -- accounting for both PGO and ICP.

\subsection{Discussion}

We discuss the limitations of our system and possible future improvements.
We first recall the importance of odometry calibration to maximize mapping accuracy. 
The calibration approach depends on the type of sensors available on the platform and, thus, the odometry calibration is a platform-specific tuning.
In the following, we estimate the maximum area that can be mapped with our system.
This limitation mainly comes from the maximum size (i.e., 440 poses) of the sparse graph used by the hierarchical PGO.
We attempt to perform such an estimation starting from Maze 3, the most complex environment we map in our work.
Note that when mapping Maze 3, we let the drone fly two laps to acquire more maze details.
On the other hand, this is not mandatory, and flying just one lap and closing the loop when the drone crosses again through the start would as well lead to good results.
Furthermore, in our experiments, we employed a $d_{min}=\qty{0.3}{\meter}$ when adding poses to the sparse graph to demonstrate that our system can work with large graphs.
However, as shown in Table~\ref{tab:hierarchical}, setting $d_{min}=\qty{0.8}{\meter}$ results in almost the same optimization accuracy.
Mapping one lap of Maze 3 implies the drone to travel about \qty{14}{\meter}, which results in 18 poses for the sparse graph when using $d_{min}=\qty{0.8}{\meter}$.
Therefore, it is possible to map an environment of about \qty{13}{\meter\squared} using 18 poses.
By extrapolation, 440 poses would allow our system to map an environment of about \qty{317}{\meter\squared}.

Another limitation of our system comes from the ToF sensors, which replace the conventional LiDARs.
This substitution introduces several distinctions between the two technologies. Firstly, LiDARs typically possess a greater operational range, enabling robots to map distant areas of the environment. In contrast, our ToF sensor is restricted by a maximum range of \qty{4}{\meter}. Additionally, LiDARs often exhibit a higher angular resolution, resulting in measurement accuracy that is less reliant on the distance magnitude. Conversely, our employed ToF sensor operates with distinct zones, assuming that any detected obstacle within a zone is located at the zone's center. 
Considering the angle of one zone of $\theta_{zone} = \qty{45}{\degree} / 8 = \qty{5.625}{\degree}$, the maximum distance error induced by this effect is approximately $e_{max} = d \cdot \tan(\theta_{zone}/2) \approx 0.05 \cdot d$. While this error is negligible for short distances, it exceeds \qty{5}{\centi\meter} for distances beyond \qty{1}{\meter}. Consequently, these errors can lead to map misalignments, regardless of the effectiveness of PGO.

\begin{figure}[t]
    \centering
    \begin{subfigure}{0.49\linewidth}
    \centering
    \includegraphics[width=\linewidth]{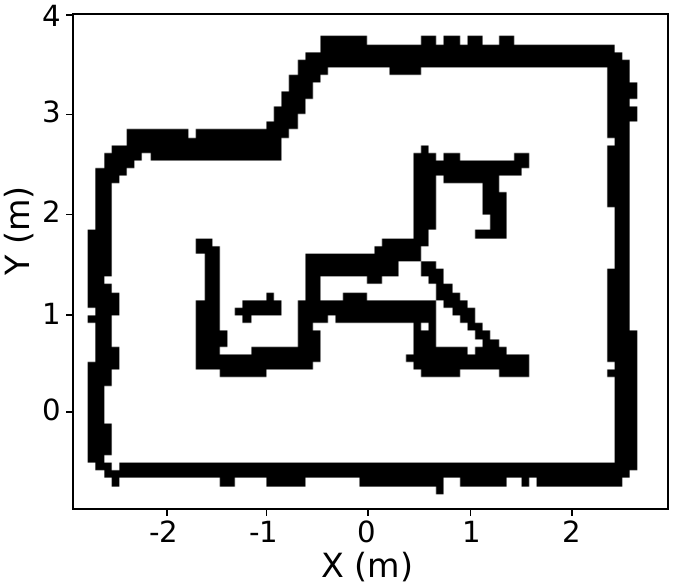}
    \caption{\qty{7.5}{\centi\meter} resolution.
    \label{fig:binary0}}
    \end{subfigure}
    \begin{subfigure}{0.49\linewidth}
    \centering
    \includegraphics[width=\linewidth]{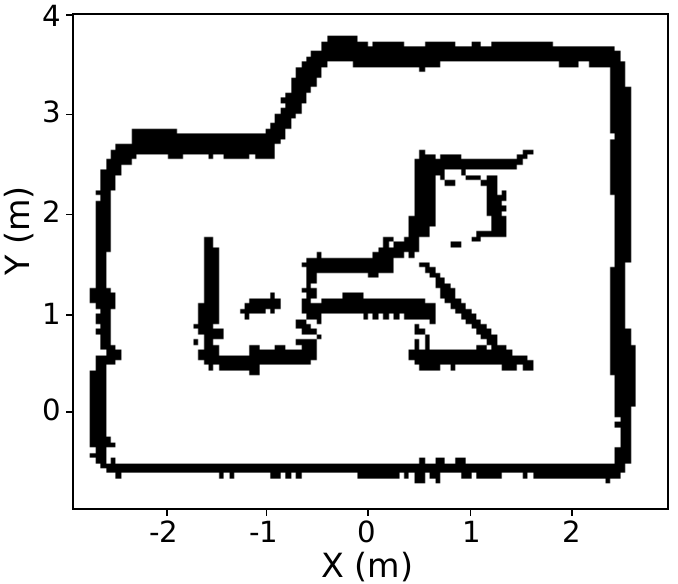}
    \caption{\qty{5}{\centi\meter} resolution.
    \label{fig:binary1}}
    \end{subfigure}
    \caption{Binary occupancy maps resulted from Maze 3.}
    \label{fig:gridmaps}
\end{figure}
A potential extension that we mention involves expanding our system's capabilities to accommodate a swarm of drones. Given that ICP can derive transformations between scans captured by different drones, it would be feasible to merge pose graphs from multiple drones and optimize them collectively to align their trajectories. 
This would enable faster mapping of an environment through parallel sensing of distinct areas by the swarm of drones, consequently reducing the overall mapping time.
Another area of future work is to perform mapping in 3D.
While our system assumes a flat environment, enabling absolute altitude estimation would allow to create a 3D map by acquiring depth measurements at various heights.
Furthermore, implementing a mechanism that discards the depth measurements associated with moving people or obstacles would enable our system to operate even in dynamic environments.

To enhance our system, NanoSLAM can be integrated with deep learning models, enabling the drone to identify objects in images and estimate their global coordinates based on the SLAM-corrected trajectory.
Previous works proved the feasibility of quantizing and deploying such models on low-power PULP platforms, with memory requirements of under \qty{500}{\kilo\byte}~\cite{lamberti2023bio, niculescu2021improving}. 
This substantiates the potential for running quantized learning models and NanoSLAM concurrently on the GAP9 SoC.
Lastly, another possible exploration entails leveraging the map generated by our approach to enable additional functionalities, such as optimal path planning. Converting the dense maps produced by our approach into binary maps -- using the approach from~\cite{polonelli2023towards} --  would further reduce the memory footprint of the maps. 
For example, Figure~\ref{fig:gridmaps} depicts the binary representation of the dense map from Figure~\ref{fig:maze3-map1} for resolutions of \qty{7.5}{\centi\meter} and \qty{5}{\centi\meter}.

As previously discussed, having the SLAM-related computation performed onboard brings significant advantages, such as reduced latency and increased security.
Nonetheless, considering the drone used in this work, we offer a rough estimate of the additional latency for offloading the computation to an external base station.
In this scenario, the drone transfers the pose graph to the base station for SLAM computations and then retrieves the optimized poses.
Considering a robust logging rate of \qty{5}{\kilo\byte / \second}~\cite{budaciu2019evaluation}, sending the pose graph would require about \qty{2.4}{\second} for a graph with 1000 poses.
Even neglecting the SLAM execution time, it results in a roundtrip time of \qty{4.8}{\second} every loop closure.
For several loop closures, this approach would translate into a mission flight time reduced by several tens of seconds, a significant value for a drone with a flight time in the order of a few minutes.
Furthermore, obstacles or walls could degrade the radio data rate or even make radio communication unviable.

\section{Conclusions}
\label{sec:conclusions}

The paper presented NanoSLAM, a lightweight SLAM for autonomous drones, and the methodology to enable fully onboard mapping for small robotic platforms, which before was only possible with larger and more power-intensive computational platforms.
NanoSLAM is the first system that enables SLAM for autonomous nano-UAVs and performs the computation entirely onboard exploring a novel RISC-V parallel low-power processor.
We demonstrated the effectiveness of NanoSLAM by mapping three different real-world mazes, achieving a mapping error down to \qty{4.5}{\centi\meter} and reducing the trajectory estimation error by up to  67\%.
The SLAM algorithm runs onboard in less than \qty{250}{\milli\second} and the whole mapping pipeline requires less than \qty{500}{\kilo\byte} of RAM.
In spite of its remarkably lightweight configuration (\qty{44}{\gram}), the system introduced in this study achieves mapping accuracy on par with SoA approaches developed for standard-size UAVs, but consumes only \qty{87.9}{\milli\watt}.
The system presented in this paper sets the foundation for increased autonomy in small form-factor robots with highly constrained hardware, thus introducing novel technology to the field of nano-UAVs. 
By enabling a comprehensive environmental map, this advancement opens up possibilities for advanced navigation solutions, including enhanced flight autonomy through optimal path planning.

\section*{Acknowledgments}
This work is partly supported by BRAINSEE project (\#8003528831) funded by armasuisse Science and Technology of the Swiss Confederation.

\bibliographystyle{IEEEtran}
\bibliography{IEEEabrv,bibliography}

\end{document}